%% file: gamearena2_nips25.tex
\documentclass{article}

\PassOptionsToPackage{numbers, compress}{natbib}

\usepackage[preprint]{neurips_2025}

\input{math_commands.tex}

\usepackage[dvipsnames]{xcolor}
\definecolor{MidnightBlue}{RGB}{25, 25, 112}
\definecolor{RoyalBlue}{RGB}{65, 105, 225} 
\usepackage{graphicx}
\usepackage{listings}
\usepackage{booktabs}
\usepackage{multirow}
\usepackage{enumitem}
\usepackage{subcaption}
\usepackage{caption}
\usepackage{framed}
\usepackage{fancybox}
\usepackage{soul}
\usepackage{tcolorbox}
\usepackage{tabularx}
\usepackage{amssymb}
\usepackage{pifont}
\colorlet{punct}{red!60!black}
\definecolor{background}{HTML}{EEEEEE}
\definecolor{delim}{RGB}{20,105,176}
\colorlet{numb}{magenta!60!black}

\usepackage[colorlinks=true, urlcolor=blue]{hyperref}

\newcommand{\cmark}{\ding{51}}  
\newcommand{\xmark}{\ding{55}}  

\lstdefinelanguage{json}{
    basicstyle=\scriptsize\ttfamily,
    showstringspaces=false,
    breaklines=true,
    frame=single,
    backgroundcolor=\color{white},
    literate=
     *{0}{{{\color{numb}0}}}{1}
      {1}{{{\color{numb}1}}}{1}
      {2}{{{\color{numb}2}}}{1}
      {3}{{{\color{numb}3}}}{1}
      {4}{{{\color{numb}4}}}{1}
      {5}{{{\color{numb}5}}}{1}
      {6}{{{\color{numb}6}}}{1}
      {7}{{{\color{numb}7}}}{1}
      {8}{{{\color{numb}8}}}{1}
      {9}{{{\color{numb}9}}}{1}
      {:}{{{\color{punct}{:}}}}{1}
      {,}{{{\color{punct}{,}}}}{1}
      {\{}{{{\color{delim}{\{}}}}{1}
      {\}}{{{\color{delim}{\}}}}}{1}
      {[}{{{\color{delim}{[}}}}{1}
      {]}{{{\color{delim}{]}}}}{1},
}

\usepackage[utf8]{inputenc} 
\usepackage[T1]{fontenc}    
\usepackage{hyperref}       
\usepackage{url}            
\usepackage{booktabs}       
\usepackage{amsfonts}       
\usepackage{nicefrac}       
\usepackage{microtype}      
\usepackage{xcolor}         

\usepackage{amsmath}
\usepackage{mathtools}
\usepackage{mdframed}
\usepackage{soul}
\usepackage{xcolor}
\usepackage{ragged2e}
\usepackage[ruled,vlined]{algorithm2e}

\sethlcolor{lightgray}
\definecolor{lightgreen}{rgb}{0.56, 0.93, 0.56}
\definecolor{deepblue}{HTML}{003D79}

\title{
\capsysname: How Good are LLMs at Playing Games?}

\author{
    Lanxiang Hu\thanks{Equal contributions.}\textsuperscript{\hspace{1.5mm}}$^{1}$ \hspace{1pt}
    Mingjia Huo\footnotemark[1]\textsuperscript{\hspace{1.5mm}}$^{1}$
    \hspace{1pt}
    Yuxuan Zhang\thanks{Significant contributions.}\textsuperscript{\hspace{1.5mm}}$^{1}$
    \hspace{1pt}
    Haoyang Yu\footnotemark[2]\textsuperscript{\hspace{1.5mm}}$^{1}$
    \And
    Eric P. Xing$^{2}$
    \hspace{1pt}
    Ion Stoica$^{3}$
    \hspace{1pt}
    Tajana Rosing$^{1}$
    \hspace{1pt}
    Haojian Jin$^{1}$
    \hspace{1pt}
    Hao Zhang$^{1}$\\
    $^1$UC San Diego
    \hspace{1pt}
    $^2$ MBZUAI
    \hspace{1pt}
    $^3$UC Berkeley
}

\newcommand{\capsysname}{\textsc{LMGame-Bench}}
\newcommand{\sysname}{\text{lmgame-Bench }}
\newcommand{\sysnamenospace}{\text{lmgame-Bench}}

\newcommand{\hlred}[1]{\sethlcolor{pink}\hl{#1}}
\newcommand{\hlgreen}[1]{\sethlcolor{lightgreen}\hl{#1}}
\newcommand{\hlyellow}[1]{\sethlcolor{yellow}\hl{#1}}

\newif\ifcomments
\commentstrue
\ifcomments
    \providecommand{\ion}[1]{{\color{blue}{[ion: #1]}}}
\else
    \providecommand{\ion}[1]{}
\fi

\begin{document}

\maketitle

\input{text/0_abstract.tex}
\input{text/1_intro.tex}
\input{text/3_methods.tex}
\input{text/4_experiments_1_rankings.tex}
\input{text/4_experiments_2_correlations.tex}
\input{text/4_experiments_3_train.tex}

\input{text/2_related.tex}

\input{text/5_conclusion.tex}


\newpage
\bibliography{gamearena2_nips25}
\bibliographystyle{splncs}

\appendix
\input{text/7_appendix_1_train}
\input{text/7_appendix_2_data_contamination}
\input{text/7_appendix_3_prompts_optimization}

\input{text/7_appendix_4_additional_correlation_analysis}

\input{text/7_appendix_5_harness_effectiveness}
\input{text/7_appendix_6_games}

\end{document}

%% file: math_commands.tex
\usepackage{amsmath,amsfonts,bm}

\def\eqref#1{equation~\ref{#1}}

\def\1{\bm{1}}

\DeclareMathAlphabet{\mathsfit}{\encodingdefault}{\sfdefault}{m}{sl}
\SetMathAlphabet{\mathsfit}{bold}{\encodingdefault}{\sfdefault}{bx}{n}



%% file: text/0_abstract.tex
\begin{abstract}




Playing video games requires perception, memory, and planning -- exactly the faculties modern large language model (LLM) agents are expected to master. 
We study the major challenges in using popular video games to evaluate modern LLMs and find that directly dropping LLMs into games cannot make an effective evaluation, for three reasons: brittle vision perception, prompt sensitivity, and potential data contamination. We introduce \sysname to turn games into reliable evaluations. \sysname features a suite of platformer, puzzle, and narrative games delivered through a unified Gym‑style API and paired with lightweight perception and memory scaffolds, and is designed to stabilize prompt variance and remove contamination. Across 13 leading models, we show \sysname is challenging while still separating models well. Correlation analysis shows that every game probes a unique blend of capabilities often tested in isolation elsewhere. More interestingly, performing reinforcement learning on a single game from \sysname transfers both to unseen games and to external planning tasks. Our evaluation code is available at \url{https://github.com/lmgame-org/GamingAgent/tree/main/lmgame-bench}.

\end{abstract}

%% file: text/1_intro.tex
\section{Introduction}











Games have long been a standard testbed for reinforcement learning (RL)~\cite{brockman2016openai, towers2024gymnasium}.
Recent advancements show LLMs can be trained as agents using RL to navigate interactive environments~\cite{wang2025ragen, zhou2025sweet}, including games~\cite{wang2025ragen, NEURIPS2024rl4vlm}. Gaming environments therefore naturally emerge as promising benchmarks~\cite{shi2025korgymdynamicgameplatform, ruoss2025lmactbenchmarkincontextimitation, wang2025enigmaeval, anthropic2025claude37, paglieri2024balrog, costarelli2024gamebench, wu2023smartplay}, because they stress the same skills we expect from useful LLM agents -- seeing, reasoning, and planning over many steps -- and because there are thousands of existing games that can be turned into benchmarks~\cite{perez2018general, laird2001human}.
Despite the advantages, existing work lacks quantitative understandings of how to interpret gaming performance and the effectiveness of using games to evaluate today's LLMs.
At first glance, evaluating LLM agents on games appears straightforward, 
by simply sending game screenshots to vision-language models (VLMs) to generate the next actions. However, directly placing a model in gaming environments can result in low performance, often close to that of random action-taking baselines. This is because even the most advanced reasoning models fall short in vision perception and long-horizon decision making~\cite{yang2024thinking, waytowich2024atari, mosquera2024can}. 
This leaves an open question: can we turn games into more effective benchmarks for evaluating LLMs?



We introduce \sysnamenospace: a benchmark that builds on well-established video games, including platformer, puzzle solving, narrative-driven detective games. 
%
%
\sysname introduces scaffolds in a principled way to overcome common challenges, including poor vision perception, potential data contamination, and prompt sensitivity. First, \sysname enriches evaluation settings by developing a gaming harness, including perception and memory modules to amortize vision perception limitations and facilitate long-horizon planning as shown in Fig.~\ref{fig:game_and_overview}. Second, \sysname adapts game settings to mitigate data contamination once it is detected. Finally, \sysname also employs a standardized prompt optimization technique to reduce prompt sensitivity. 

\input{figure_text/games_and_eval_overview}

We put leading models in \sysnamenospace.
Evaluation results from 13 models across 6 games demonstrate that \sysname presents a challenging benchmark far from being saturated due to the gap between model performance and human-level proficiency. The benchmark effectively differentiates models: o3~\cite{openai2025o3o4} and o1~\cite{openai2024o1} achieve top-2 best performance across all games, followed by other models with reasoning capabilities such as Gemini-2.5-pro-preview~\cite{google2025gemini25} and Claude-3.7~\cite{anthropic2025claude37}. Among non-reasoning models, GPT-4.1 leads the pack.

Given the lack of a quantitative framework for interpreting the performance of the game, we develop a set of techniques to understand LLM performance in \sysnamenospace. 
To study the correlation between \sysname and other widely used benchmarks, we perform low rank matrix factorization in the model benchmark performance matrix and use linear regression to predict game rankings. This allows us to uncover latent relationships between games and other domain-specific benchmarks. For instance, we show math- and coding-related benchmarks align more closely with long-horizon games like Sokoban and Tetris, while language-related benchmarks align closely with text-rich detective games like Ace Attorney.
More interestingly, RL training on Sokoban and Tetris leads to cross-game improvements and boosts performance on planning and agentic tasks like Blocksworld and WebShop. These results demonstrate that gaming environments are not only effective in evaluating core LLM abilities but also serve as valuable training environments to improve those abilities.



In summary, our paper makes the following contributions:
\begin{itemize}
    \item We introduce \sysnamenospace, the first benchmark that uses video games to evaluate state-of-the-art LLMs with or without scaffolds.
    \item We show that under different settings, \sysname can effectively discriminate between models by underscoring their key strengths and weaknesses.
    \item We provide quantitative analysis of model performance on \sysname and show games evaluate combinations of existing capabilities commonly assessed in isolation.
    \item We show training LLMs with RL on games improves their performance not only on games of various settings, but also on planning and agent-based tasks like Blocksworld and WebShop.
\end{itemize}


%% file: figure_text/games_and_eval_overview.tex
\begin{figure}
    \centering
    \includegraphics[width=\linewidth]{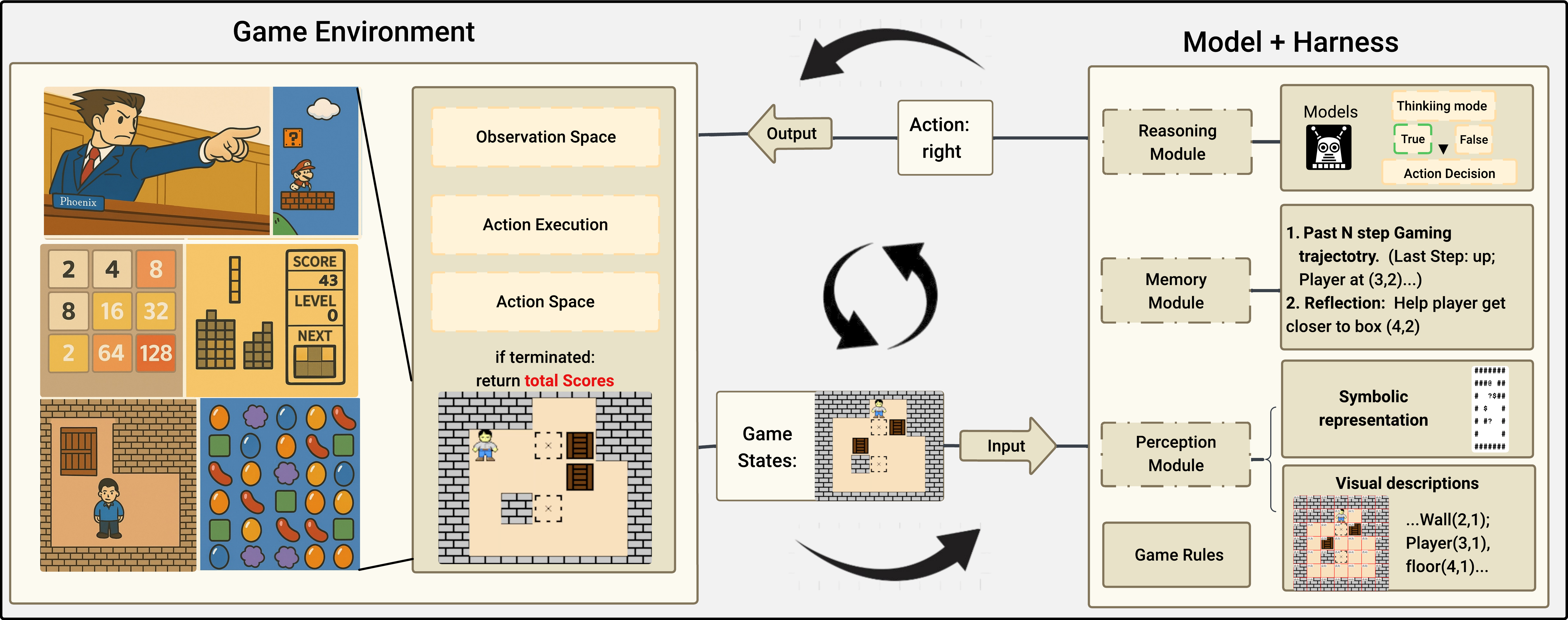}
    \caption{
    \sysname uses modular harnesses—such as perception, memory, and reasoning modules—to systematically extend a model’s game-playing capabilities, allowing the model to engage with a simulated game environment through iterative interaction loops.
    }
    \label{fig:game_and_overview}
    \vspace{-10pt}
\end{figure}

%% file: text/3_methods.tex
\section{\capsysname}
\label{sec:benchmark}



We build the backbone of \sysname on six well-known games (\S~\ref{sec:game_design}) to evaluate leading models' performance without providing them with scaffolds.
We choose the games to ensure diverse skill coverage—including spatial reasoning, pattern recognition, and long-horizon planning—for evaluating different aspects of model capabilities. 
However, directly evaluating models on games in their original forms poses three key challenges: (1) low discriminability – model scores often cluster near random‐play baselines, making it difficult to distinguish model's ability; (2) contamination risk - game assets and solutions may appear in training data, so high scores can reflect memorization rather than skill; (3) prompt variance – inconsistencies in prompt formatting can cause large performance swings, undermining comparability.  
To address these, we introduce mitigation techniques in \S~\ref{sec:system_design}.


\subsection{Benchmark Design}
\label{sec:game_design}


In designing \sysnamenospace, we intentionally recycle well-known games not only for their familiarity and popularity but also because they encapsulate a broad spectrum of reasoning and interaction skills. 
Our goal is to preserve the original game settings that are carefully designed to challenge human cognition.
In this section, we highlight the broad range of perception and generation abilities evaluated in \sysname's game settings.

\subsubsection{Games}
\label{sec:game_design_rules}

\textbf{Super Mario Bros.} Super Mario Bros is a side-scrolling platformer game where the player controls Mario to navigate through obstacles, defeat enemies, and reach the end of each level by navigating through the environment. Success requires precise timing and strategic movement, making it a classic benchmark for evaluating (1) visual perception, (2) spatial reasoning in 2D for character control, and (3) goal-directed planning with partial observability~\cite{rintanen2004complexity} in interactive environments.

\textbf{Tetris.} Tetris is a tile-matching puzzle game where players must strategically rotate and place falling Tetris tiles of 7 different geometric shapes to complete and clear horizontal lines. The game emphasizes (1) visual perception for pattern recognitions, (2) spatial reasoning for correct tile matching and geometric rotations~\cite{lau2017selective}, and (3) long-horizon planning with partial observability for decision-making on where and how to drop a tile~\cite{Demaine2003TetrisHard}. 

\textbf{Sokoban.} Sokoban is a grid-based puzzle game where the player pushes boxes to designated target locations within confined spaces. It emphasizes (1) visual perception, (2) spatial reasoning to navigate both the character and the box, and (3) long-horizon planning to avoid deadlocks~\cite{culberson1997sokoban}. The game’s low fault tolerance is especially pronounced. Many actions are irreversible, and a single wrong move can fail the puzzle.

\textbf{Candy Crush.} Candy Crush is a match-three puzzle game where players swap adjacent candies to form aligned sequences and trigger cascading effects to eliminate matched sequences. It requires (1) visual perception to identify different candies, (2) spatial reasoning to anticipate chain reactions at different locations, and (3) long-horizon planning to conserve moves to maximize total points. The gameplay features limited moves, making it crucial to plan moves carefully.

\textbf{2048.} 2048 is a sliding-tile puzzle game where players combine numbered tiles on a grid to reach the 2048 tile. It evaluates (1) visual perception for tracking tile values and movements, (2) spatial reasoning to manage merging paths, and (3) goal-directed planning to maximize merge potential~\cite{zaky2014minimax}. Errors compound quickly due to the game’s limited space and could lead to irreversible failure states.

\textbf{Ace Attorney.} Ace Attorney is an interactive courtroom-drama visual novel in which the player, acting as defense attorney Phoenix Wright, must investigate crime scenes, interview witnesses, and present evidence in court to reveal contradictions and secure a “Not Guilty” verdict. The game stresses (1) long-context language understanding, tracking hundreds of dialogue turns, testimonies, and evidentiary facts,
(2) causal \& deductive reasoning under partial observability—linking dispersed clues, inferring hidden motives, and spotting logical gaps,
and (3) long-horizon, low-fault-tolerance decision making, to decide when to press, object, or present evidence over multi-stage trials.



\subsubsection{Game Settings}
\label{sec:game_design_settings}

\sysname maintains the integrity of the original design choices, which ensure scalability. In this section, we focus on standardizing game settings including inputs and outputs of the gaming environments as part of our benchmark design as shown in Fig.~\ref{fig:game_and_overview} (right). We can formalize a gaming process as a partially or fully observable Markov Decision Process (MDP) with the following definitions as a generalizable formalism applicable to all games.

\textbf{Observation Space Representations.} Many existing games are inherently graphical, requiring human players to process multimodal information, including visual, textual, and spatial cues, from the user interface (UI) to interpret game states and make decisions~\cite{arxiv2024ufo, arxiv2025infiguiagent}. 
We denote symbolic and graphical representations of the game states as the set of all possible observations $S$. We don't make a distinction between game state space and observation space, which is not the main focus of this work.

\textbf{Action Space.} Let the set of all actions in action space be $\mathcal A$. To interface with the game, \sysname considers multi-turn interactions. \sysname streams game states $s_i \in S$ to model $M$, each time it generates action $a_i \in A$ in response to the current state to maximize rewards, which are meticulously crafted scores in classical games, with details specified in \S~\ref{sec:game_metrics}. 



\textbf{Difficulty.} Games in \sysname are designed with varying levels of difficulty, structured along two key dimensions: (1) fault tolerance and (2) state-action space complexity. We define three levels of fault tolerance -low (one wrong move fails), medium (errors accumulate but can be recovered), and high (many mistakes can be tolerated without significantly affecting future game states). We employ a memory module to curb search‑space explosion (\S~\ref{sec:system_design_modules}).













\subsubsection{Metrics Design}
\label{sec:game_metrics}

In line with Gymnasium~\cite{towers2024gymnasium}, we treat a reward as a function, $\mathcal{R}\colon \mathcal{S}\times\mathcal{A}\times\mathcal{S}\to\mathbb{R}$, which returns the payoff obtained when the agent executes action \(a\in\mathcal{A}\) in state \(s\in\mathcal{S}\) and transitions to state \(s'\). \sysname adopts this definition in \sysname to evaluate the performance of LLM-based agents. Depending on the nature of the game, we categorize evaluation metrics into two types: progression rewards and long-horizon rewards.



\textbf{Progression Rewards.} Progression rewards apply to games that are either designed to run indefinitely or follow a linear narative structure, where success is measured by completing a full progression sequence. These rewards capture how far an agent advances before failure or termination. For example, in Super Mario Bros, progression can be quantified linearly by the agent’s movement along the x-axis. In games like Tetris and 2048, progression is reflected in the cumulative score as the agent strategically places Tetrominoes or merges tiles to clear the board.

\textbf{Long-Horizon Rewards.} Long-horizon rewards are especially useful for games that span multiple levels and where intermediate progress is difficult to quantify. For example, in Sokoban, solving a level often requires a series of non-obvious steps to maneuver boxes into their target positions. Simply moving an individual box into place may not represent meaningful progress, as it could later need to be repositioned. Thus, linear metrics fail to capture the agent’s true advancement toward the final objective of completing the level.


\subsection{Benchmark Effectiveness Enhancement}
\label{sec:system_design}
While using games as evaluation presents challenging environments and breadth, we find that directly evaluating models on games exposes several challenges: low discriminability, contamination risk, and prompt variance. In this section, we address these issues by introducing gaming scaffolds for LLMs, contamination detection, and prompt standardization, enabling \sysname to function as a more robust benchmark that reliably differentiates LLMs.

\subsubsection{Gaming Harness}
\label{sec:system_design_modules}

Excluding text-only models, 40\% of game runs without the harness fail to outperform a random-play baseline. To raise the cap and bring higher contrast, \sysname provides a suite of harness modules that can be toggled on or off for any experiment (workflow in Fig. \ref{fig:game_and_overview}). Activating the harness boosts scores far beyond both random play and the unharnessed setting, creating clearer performance gaps between models. With harnessing, 86.7\% of game runs beat the random baseline, and paired-sample t-tests confirm that harnessed runs score significantly higher than their unharnessed counterparts on Candy Crush, 2048, Tetris, Ace Attorney, and Sokoban (details in Appx. \ref{appendix:harness_hypothesis_test} and \ref{appendix:memory_module_analysis}).





\textbf{Perception Modules.}
Since the video games are inherently multimodal, we build perception modules 
%
that convert UI inputs into symbolic representations or textual descriptions of game states to facilitate understanding. 
For grid-based games (Sokoban, Candy Crush, 2048, Tetris), the module converts the visual layout into a text-based table from game backends, listing object coordinates and their properties, e.g. "Box at (2,3)", "Wall at (4,5)". This allows models to directly understand spatial relationships in replacement of raw image inputs to minimize perception errors. For text-based games (e.g. Ace Attorney), the module extracts dialogues and describes visual elements in text format to provide narrative context and critical visual cues. Likewise, we use perception module to extract visual elements in platformer and action games (e.g. Super Mario) to facilitate decision making.

\textbf{Memory Modules.}
Some games, like Sokoban and Tetris, exhibit a rapidly growing decision space as gameplay advances and interactive elements scale (e.g., boxes, grid size, tetromino types). As a result, they come with higher difficulty levels than the other games. To better distinguish models, we integrate additional memory modules into \sysnamenospace. This setup allows selective activation of two components: (1) a transient memory module, which records the past $N$ game states and actions, and (2) a reflection module, which encodes explicit lessons learned to avoid failure, inducing actions in specific game states, thereby helping to narrow the action space.

\textbf{Reasoning Modules.} 
Reasoning models~\citep{guo2025deepseek, openai2025o3o4, google2025gemini25, anthropic2025claude37} have emerged as a new inference paradigm, where models explore multiple reasoning paths and synthesize a more accurate answer at the end. These models have shown strong performance in tasks such as mathematics, code generation, and planning. \sysname is designed to support such reasoning traces by allowing models to be evaluated with or without long chain-of-thought (long-CoT) reasoning.



\subsubsection{Data Contamination}
\label{sec:system_design_data_contamination}
Because \sysname reuses publicly available game assets, many images and scripts may already appear in model pre-training data. To ensure the model isn't merely recalling artifacts, we implement both vision-level and text-level contamination checks. Concretely, we test vision-level data contamination in \textit{Super Mario Bros} and text-level data contamination in \textit{Ace Attorney}, whose sprite and dialogue are widely distributed online (Appx.~\ref{appendix:data_contamination}). The other games, Tetris, 2048, Candy Crush, and Sokoban, feature combinatorial state spaces \cite{Dor1999Sokoban, Demaine2003TetrisHard, Guala2014BejeweledHard}, making overlap with training data negligible.


\textbf{Vision-level.} We assess whether models recall the visual structure of \textit{Super Mario Bros} level 1-1 by prompting them to reorder shuffled RGB frames. Only a few models exhibit a moderate positive alignment, yet these alignment scores do not significantly track with their performance rankings.  This suggests that they rely on local perception rather than memorized sequences. Since our metric evaluates how far models can play Super Mario Bros, we focus on vision-level contamination that may expose future frames, and disregard contamination within the current frame-such as prior knowledge that a "?" brick may contain a mushroom-as it does not affect sequence prediction.

\textbf{Text-level.} In \textit{Ace Attorney}, we test whether the models reproduce public fan transcripts. Using Sentence-BERT similarity, we find a strong correlation between output similarity and performance, especially in a 6-model subset. However, after applying structured prompt-based mitigation—including entity masking, paraphrasing, and enforced reasoning ~\cite{dong2024generalization, carlini2018secret} —the correlation disappears, and model rankings instead align with judged reasoning quality (Fig.~\ref{fig:text_comparison_side_by_side}). 

\input{figure_text/ace_attorney_data_contamination_visualization}

\subsubsection{Prompt Standardization}
\label{sec:system_design_prompt}

Evidence shows prompt engineering is very effective in enhancing LLM performance across various games~\cite{paglieri2024balrog, wang25lvlmgame}. However, we find performance variability may exceed the $\pm 1$ standard deviation range even among empirically optimized prompts as shown in Table~\ref{tab:prompt_optimization}. In \sysnamenospace, we develop a two-stage optimization technique to reduce prompt variance. 


\textbf{Empirical Prompt Engineering with Standardized Formats}. 
\sysname with harness can be regarded as an agentic workflow. We follow empirical practices in agent prompt design from recently released agents~\cite{agashe2024agent, tan2024cradle}. Our empirical prompt follows the format [$\{\mathcal{J}_{[\min\left(0, i-N\right):i-1]}\}$, $R_{i-1}$, $s_i$] to generate the next action $a_i$ for the current turn. $\{\mathcal{J}_{[\min\left(0, i-N\right):i-1]}\}$ is the gaming trajectory of past $N$ turns, each composes of a tuple of state, and action reward $[s_j, a_j, r_j]$, and $R_{i-1}$ is reflection made by the memory module.

\textbf{DSPy-based Optimization}. The second stage intends to standardize prompt optimization using DSPy~\cite{khattab2024dspy}. It leverages the chain-of-thought module and Instantiate Stochastic Introspective Mini-Batch Ascent (SIMBA) optimizer from DSPy (details are provided in Algorithm~\ref{alg:dspy}), using rewards from gaming environments as evaluation metrics. Given a set of game rules and an initial prompt, the optimizer iteratively refines the prompt, guided by evaluation metrics to maximize cumulative rewards. This approach results in a highly optimized system prompt with best average performance across all target models for each game. Results from Table~\ref{tab:prompt_optimization} show across 3 runs, standardized prompt optimization with DSPy can indeed reduce performance variance across different empirically optimized initializations in games such as 2048 by 33.8\% to 63.5\% (Appx.~\ref{appendix:prompt_optimization}).

\subsubsection{Other Limitations and Discussion}
\label{sec:other_limitations}


After implementing these improvements, \sysname significantly enhances model performance, achieves prompt convergence through bootstrapping in DSPy, and effectively detects and mitigates data contamination. However, a few limitations still persist, primarily in two areas. (1) Performance variance continues to be high in partially observable games like Super Mario Bros, where game randomness plays a significant role. (2) The computational cost remains substantial, as generating actions could result in long reasoning chains that are highly repetitive across multiple turns. We observe that the challenges faced by contemporary LLMs generalize across most existing games beyond those evaluated in \sysnamenospace, highlighting the need for improved model capabilities and more efficient inference to reduce operational costs.


%% file: figure_text/ace_attorney_data_contamination_visualization.tex
\begin{figure}[h]
    \centering
    \sethlcolor{yellow}
    \begin{minipage}{0.45\textwidth} 
        \begin{tcolorbox}[
            colback=gray!10,  
            colframe=MidnightBlue, 
            sharp corners,
            title={\textcolor{white}{\textbf{Text Comparison: Ground Truth vs. o3}}},
            coltitle=white, 
            fonttitle=\small,
            colbacktitle=black, 
            boxrule=0.6pt,
            width=\textwidth
        ]
        \scriptsize
        \begin{minipage}{0.49\textwidth} 
            \textbf{Ground Truth Text:} \\
            \vspace{0.2em}
            \raggedright
            \hl{Frank Sahwit, Round 1 Statement: }“I remember the time exactly: \hl{It was 1:00 PM.”} \hl{Present: “Cindy's Autopsy Report.”} \hl{Contradiction Exposed: Death occurred “between 4 PM and 5 PM,”} so the body could not be found at 1 PM.\\
        \end{minipage}%
        \hfill
        \begin{minipage}{0.49\textwidth} 
            \textbf{o3 Generated Text:} \\
            \vspace{0.2em}
            \raggedright
            \hl{Frank Sahwit, Round 1 - Statement: "It was exactly 1:00 PM} when I saw Larry Butz run out of the apartment." \hl{Present: Cindy's Autopsy Report} -\hl{Contradiction Exposed:} Autopsy shows \hl{death occurred between 4:00 and 5:00 PM}, making a 1:00 PM murder impossible.\\
        \end{minipage}
        \end{tcolorbox}
    \end{minipage}%
    \hfill
    \begin{minipage}{0.45\textwidth} 
        \centering
        \includegraphics[width=\linewidth]{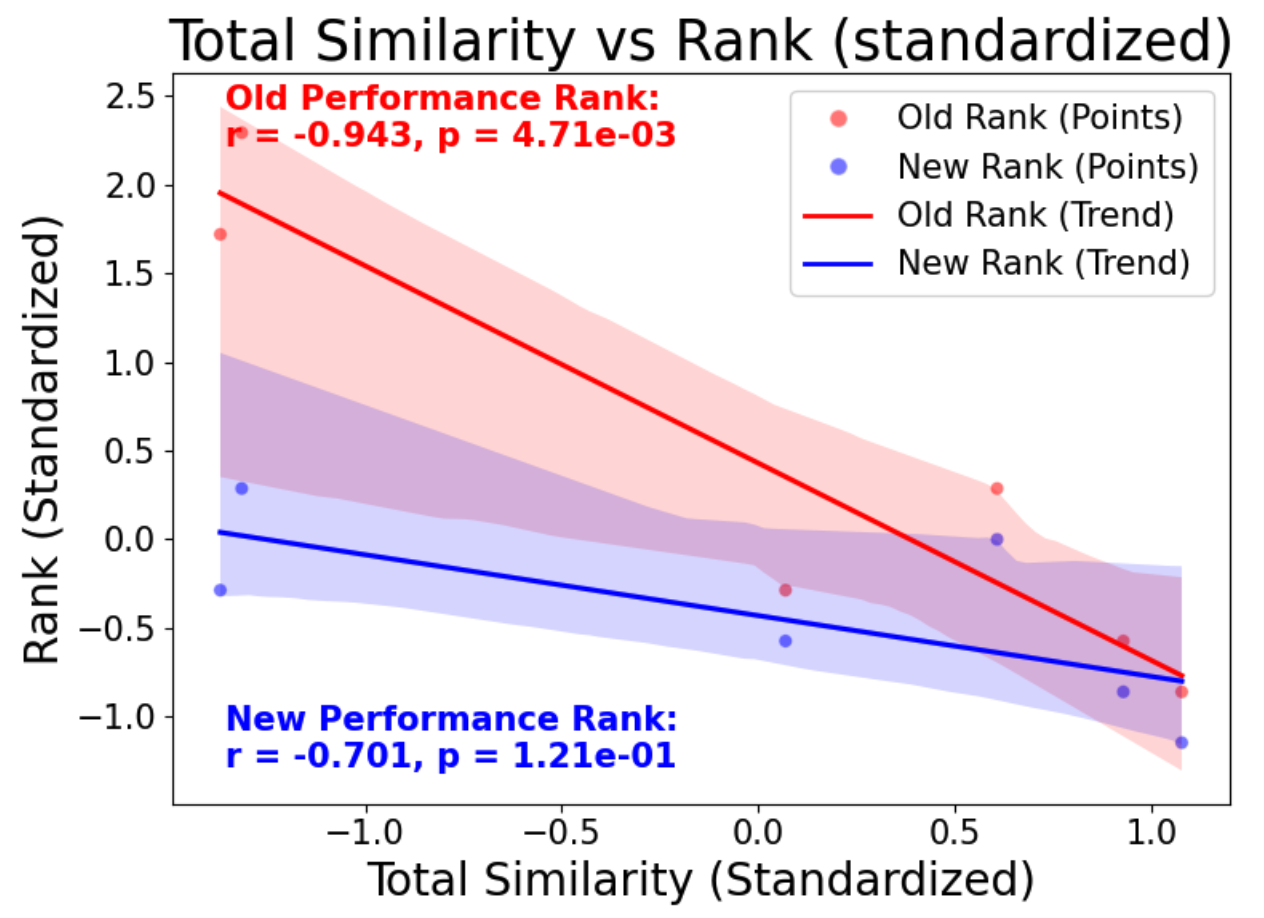} 
    \end{minipage}
    
    \caption{(Left) Example from Ace Attorney showing contradictions in O3-generated text vs. ground truth. (Right) Effect of mitigation on similarity-performance correlation; red and blue lines show correlations with old and new leaderboard ranks, respectively.}
    \label{fig:text_comparison_side_by_side}

\end{figure}

%% file: text/4_experiments_1_rankings.tex
\section{Experiments}


In this section, we present the rankings of 13 state-of-the-art models, both with and without the gaming harness, evaluated on a suite of 6 classical video games. We also analyze the effectiveness of each harness module (\S~\ref{sec:exp_rankings}), as well as the modules' combined effectiveness. Then we investigate how gaming environments reveal core capabilities commonly evaluated in LLMs through correlation analysis, low-rank factorization, linear modeling (\S~\ref{sec:exp_analysis}), and RL training (\S~\ref{sec:exp_train}). We quantify the issue of gaming data contamination and propose mitigation techniques, which are detailed in Appx.~\ref{appendix:data_contamination}.

\subsection{Model Performance}


\label{sec:exp_rankings}


When putting LLMs and VLMs in gaming environments, we first study if they can play the games well without gaming harness. 
Table~\ref{tab:model-game-performance-merged} shows most models perform poorly across classic video games.
Specifically, over three fourths of models often score no points on Sokoban and Ace Attorney without harness support. On Tetris and Candy Crush, their scores are close to random play, which suggests they succeed by chance rather than understanding.
As a result, it's numerically hard to distinguish models given poor model performances and randomness inherent to games. 

To address this issue, we design different levels of scaffolding, as described in \S~\ref{sec:game_design_settings}, to better differentiate model capabilities. Results in Table~\ref{tab:combined_module_scores} show that the harness leads to consistent and sometimes substantial gains across both games. Harness can also pull scores far away from random play, and reduce variance for more stable benchmark results. We provide detailed analysis for each module below. 

\textbf{Perception Modules.} In grid-based games like Sokoban, vision scaffolding helps models perform better by providing them with symbolic representation of game states read from the game backend. Models like Gemini and O4-mini show substantial improvements, revealing that structured spatial inputs can unlock planning capabilities not expressed under raw input conditions. 

In games with more complex graphical interface like Super Mario Bros and Ace Attorney, \sysname uses o3 in perception module to extract key iterative visual elements and status indicators as textual descriptions. We find in such settings perception module plays a less significant role and model performance still primarily relies on a model's vision understanding capability.

\textbf{Memory Modules.} In \sysnamenospace, we employ o3 to generate reflections in the module (Appx.~\ref{appendix:memory_module_analysis}). This module proves especially valuable in games like 2048, where incorporating memory significantly enhances performance. Models with weaker zero-shot capabilities benefit the most, underscoring the importance of leveraging sequential context in long-horizon games. By reflecting on past interactions, the memory module helps models make more informed decisions and avoid common pitfalls.

\textbf{Combined Support.} 
The strongest results are observed when both modules are enabled as shown in Table~\ref{tab:combined_module_scores}.
Direct comparison of model performance before and after using effect sizes in Appx.~\ref{appendix:harness_hypothesis_test} show 
models whose performance that were previously indistinguishable from random baselines now separate well, making comparisons more meaningful. We also employ paired-sample t-Test to demonstrate statistically significant improvement in model performance after applying gaming harness support in \sysnamenospace. Analysis of the coefficient of variation (CV, expressed as a percentage) across gaming settings shows that, for most games, CV is consistently lower when harness support is enabled compared to the unharnessed setting. This indicates that harness support tends to yield more stable game performance.

In summary, perception and memory modules boost performance across games. Perception is more useful in spatial reasoning tasks like Sokoban, while memory is essential for temporal planning in 2048. Together, they amplify model performance differences and make \sysname benchmarking results more informative.

\input{table/ranking_table}

\input{table/module_ablation_table}






%% file: table/ranking_table.tex
\begin{table}[t]
\centering
\small
\caption{Model performance raw scores, evaluated in both with and without harness settings. For games marked with $^{\dag}$, evaluation for text-only models is not supported, as vision understanding is essential for decision-making. 
 The reported results represent averages over three runs, except for models or games marked with *, which are based on a single run due to the high costs as of May 1, 2025. “N/A” indicates a non-applicable evaluation setting, as the specific model currently does not support image input.}
\vspace{0.5em}
\label{tab:model-game-performance-merged}
\resizebox{\linewidth}{!}{%
\begin{tabular}{l|c|c|c|c|c|c|c}
\toprule
\textbf{Model} & \textbf{Harness} & Sokoban & Super Mario Bros$^{\dag}$ & Tetris & 2048 & Candy Crush & Ace Attorney*  \\
\midrule
\multirow{2}{*}{claude-3-5-sonnet-20241022} 
& No  
& 0.0{\scriptsize$\pm$0.0} 
& 1540.0{\scriptsize$\pm$21.7} 
& 12.3{\scriptsize$\pm$2.5} 
& 57.8{\scriptsize$\pm$16.4} 
& 17.0{\scriptsize$\pm$18.1}
& 1.0{\scriptsize$\pm$0.0}
\\
& Yes 
& 0.0{\scriptsize$\pm$0.0} 
& 1267.7{\scriptsize$\pm$484.1}  
& 14.7{\scriptsize$\pm$1.2} 
& 108.2{\scriptsize$\pm$5.8} 
& 106.0{\scriptsize$\pm$53.4} 
& 2.0{\scriptsize$\pm$0.0} 
\\
\midrule
\multirow{2}{*}{claude-3-7-sonnet-20250219 (thinking)} 
& No  
& 0.0{\scriptsize$\pm$0.0} 
& 1430.0{\scriptsize$\pm$162.2}  
& 13.0{\scriptsize$\pm$0.0} 
& 114.2{\scriptsize$\pm$7.2} 
&126.3{\scriptsize$\pm$69.1} 
& 3.0{\scriptsize$\pm$0.0} 
\\
& Yes 
& 2.3{\scriptsize$\pm$1.5} 
& 1418.7{\scriptsize$\pm$660.3} 
& 16.3{\scriptsize$\pm$2.3} 
& 113.3{\scriptsize$\pm$3.1} 
& 484.0{\scriptsize$\pm$53.7} 
& 7.0{\scriptsize$\pm$0.0} 
\\
\midrule
\multirow{2}{*}{deepseek-r1} 
& No  
& N/A 
& N/A 
& N/A 
& N/A 
& N/A 
& N/A
\\
& Yes 
& 1.3{\scriptsize$\pm$1.2} 
& N/A 
& 14.3{\scriptsize$\pm$0.6} 
&105.2{\scriptsize$\pm$12.2}
&  447.3{\scriptsize$\pm$45.1} 
&  0.0{\scriptsize$\pm$0.0}  \\
\midrule
\multirow{2}{*}{gemini-2.5-flash-preview-04-17 (thinking)} 
& No  
& 0.0{\scriptsize$\pm$0.0} 
& 1540.7{\scriptsize$\pm$262.4} 
& 19.0{\scriptsize$\pm$4.6} 
& 107.4{\scriptsize$\pm$3.4} 
& 97.7{\scriptsize$\pm$36.1} 
& 1.0{\scriptsize$\pm$0.0} 
\\
& Yes 
& 1.7{\scriptsize$\pm$1.5} 
&1395.0{\scriptsize$\pm$240.1} 
& 16.3{\scriptsize$\pm$3.2} 
& 106.6{\scriptsize$\pm$5.3} 
&  334.7{\scriptsize$\pm$65.5} 
& 4.0{\scriptsize$\pm$0.0} 
\\
\midrule
\multirow{2}{*}{gemini-2.5-pro-preview-05-06 (thinking)} & No  & 1.0{\scriptsize$\pm$0.0} & 1025.3{\scriptsize$\pm$443.2} &  12.3{\scriptsize$\pm$3.1} & 120.5{\scriptsize$\pm$3.9} & 177.3{\scriptsize$\pm$64.9} & 8.0{\scriptsize$\pm$0.0} \\
                                         & Yes & 4.3{\scriptsize$\pm$0.6} & 1498.3{\scriptsize$\pm$203.4} & 23.3{\scriptsize$\pm$0.6} & 117.3{\scriptsize$\pm$5.9} & 416.3{\scriptsize$\pm$6.8} & 7.0{\scriptsize$\pm$0.0} \\
\midrule
\multirow{2}{*}{grok-3-mini-beta (thinking)} 
  & No  
  & N/A 
  & N/A 
  & N/A 
  & N/A 
  & N/A 
  & N/A 
\\
  & Yes 
  & 5.7{\scriptsize$\pm$0.6} 
  & N/A 
  & 21.3{\scriptsize$\pm$7.1} 
  & 118.6{\scriptsize$\pm$7.1} 
  & 254.0{\scriptsize$\pm$107.8} 
  & 0.0{\scriptsize$\pm$0.0} 
\\

\midrule
\multirow{2}{*}{llama-4-maverick-17b-128e-instruct-fp8} 
  & No  
    & 0.0{\scriptsize$\pm$0.0} 
    & 786.0{\scriptsize$\pm$462.6} 
    & 11.7{\scriptsize$\pm$1.2} 
    & 44.6{\scriptsize$\pm$11.8} 
    & 32.3{\scriptsize$\pm$41.4} 
    & 0.0{\scriptsize$\pm$0.0} 
\\
  & Yes 
    & 0.0{\scriptsize$\pm$0.0} 
    & 1468.7{\scriptsize$\pm$555.7} 
    & 10.3{\scriptsize$\pm$1.5} 
    & 106.0{\scriptsize$\pm$3.8} 
    & 128.7{\scriptsize$\pm$57.2} 
    & 0.0{\scriptsize$\pm$0.0} 
\\

\midrule
\multirow{2}{*}{gpt-4.1-2025-04-14} 
  & No  
    & 0.0{\scriptsize$\pm$0.0} 
    & 1991.3{\scriptsize$\pm$1018.5} 
    & 13.0{\scriptsize$\pm$1.7} 
    & 94.5{\scriptsize$\pm$17.0} 
    & 101.0{\scriptsize$\pm$120.2} 
    & 0.0{\scriptsize$\pm$0.0} 
\\
  & Yes 
    & 0.0{\scriptsize$\pm$0.0} 
    & 2126.3{\scriptsize$\pm$1778.4} 
    & 13.7{\scriptsize$\pm$0.6} 
    & 105.7{\scriptsize$\pm$7.0} 
    & 182.0{\scriptsize$\pm$28.7} 
    & 2.0{\scriptsize$\pm$0.0} 
\\

\midrule
\multirow{2}{*}{gpt-4o-2024-11-20} 
  & No  
  & 0.0{\scriptsize$\pm$0.0} 
  & 1028.3{\scriptsize$\pm$656.0} 
  & 14.7{\scriptsize$\pm$2.1} 
  & 70.4{\scriptsize$\pm$15.2} 
  & 59.0{\scriptsize$\pm$54.6} 
  & 0.0{\scriptsize$\pm$0.0} 
\\
  & Yes 
  & 0.0{\scriptsize$\pm$0.0} 
  & 2047.3{\scriptsize$\pm$528.2} 
  & 14.0{\scriptsize$\pm$3.6} 
  & 106.7{\scriptsize$\pm$3.5} 
  & 147.3{\scriptsize$\pm$53.4} 
  & 0.0{\scriptsize$\pm$0.0} 
\\

\midrule
\multirow{2}{*}{o1-2024-12-17 *} 
  & No  
    & 0.0{\scriptsize$\pm$0.0} 
    & 1434.0{\scriptsize$\pm$0.0} 
    & 13.0{\scriptsize$\pm$0.0} 
    & 128.1{\scriptsize$\pm$0.0} 
    & 90.0{\scriptsize$\pm$0.0} 
    & 3.0{\scriptsize$\pm$0.0} 
\\
  & Yes 
    & 2.3{\scriptsize$\pm$0.6} 
    & 855.0{\scriptsize$\pm$0.0} 
    & 35.0{\scriptsize$\pm$0.0} 
    & \textbf{128.9{\scriptsize$\pm$0.0}} 
    & 159.0{\scriptsize$\pm$0.0} 
    & \textbf{16.0{\scriptsize$\pm$0.0} }
\\

\midrule
\multirow{2}{*}{o1-mini-2024-09-12} 
  & No  
    & N/A 
    & N/A 
    & N/A 
    & N/A 
    & N/A 
    & N/A 
\\
  & Yes 
    & 1.3{\scriptsize$\pm$0.6} 
    & N/A 
    & 11.7{\scriptsize$\pm$1.2} 
    & 114.0{\scriptsize$\pm$3.7} 
    & 48.0{\scriptsize$\pm$33.9} 
    & 0.0{\scriptsize$\pm$0.0} 
\\

\midrule
\multirow{2}{*}{o3-2025-04-16 *} 
  & No  
    & 2.0{\scriptsize$\pm$0.0} 
    & 1955.0{\scriptsize$\pm$0.0} 
    & 31.0{\scriptsize$\pm$0.0} 
    & 128.2{\scriptsize$\pm$0.0} 
    & 106.0{\scriptsize$\pm$0.0} 
    & 8.0{\scriptsize$\pm$0.0} 
\\
  & Yes 
    & \textbf{8.0{\scriptsize$\pm$2.8} }
    & \textbf{3445.0{\scriptsize$\pm$0.0}}
    & \textbf{42.0{\scriptsize$\pm$0.0}} 
    &128.0{\scriptsize$\pm$0.0} 
    & \textbf{647.0{\scriptsize$\pm$0.0} }
    &\textbf{16.0{\scriptsize$\pm$0.0}} 
\\

\midrule
\multirow{2}{*}{o4-mini-2025-04-16}
  & No  
    & 1.3{\scriptsize$\pm$0.6}
    & 1348.3{\scriptsize$\pm$178.1}
    & 15.0{\scriptsize$\pm$3.6}
    & 97.6{\scriptsize$\pm$29.2}
    & 110.7{\scriptsize$\pm$49.7}
    & 2.0{\scriptsize$\pm$0.0}
\\
  & Yes 
    & 5.3{\scriptsize$\pm$1.2}
    & 1448.0{\scriptsize$\pm$161.0}
    & 25.3{\scriptsize$\pm$8.5}
    & 120.6{\scriptsize$\pm$4.9}
    & 487.3{\scriptsize$\pm$198.0}
    & 4.0{\scriptsize$\pm$0.0}
\\

\midrule
Random & --
  & 0.0{\scriptsize$\pm$0.0}
  & 987.0{\scriptsize$\pm$414.5}
  & 10.2{\scriptsize$\pm$1.8}
  & 100.4{\scriptsize$\pm$7.8}
  & 116.5{\scriptsize$\pm$51.5}
  & 0.0{\scriptsize$\pm$0.0} \\
\bottomrule
\end{tabular}%
}
\end{table}

%% file: table/module_ablation_table.tex
\begin{table}[t]
\centering
\small
\caption{Model performance (averaged across 3 runs) in \textit{Sokoban} and \textit{2048} under various conditions. ZS indicates zero-shot without any module support or memory prompt.}
 \vspace{0.5em}
\label{tab:combined_module_scores}
\setlength{\tabcolsep}{6pt} 
\renewcommand{\arraystretch}{1.1} 
\footnotesize
\resizebox{\linewidth}{!}{%
\begin{tabular}{@{}l
    *{4}{c}
    *{4}{c}@{}}
\toprule
\multirow{2}{*}{\textbf{Model}}
  & \multicolumn{4}{c}{\textbf{Sokoban}}
  & \multicolumn{4}{c}{\textbf{2048}} \\
\cmidrule(lr){2-5} \cmidrule(lr){6-9}
  & \textbf{ZS}
  & \textbf{+Memory}
  & \textbf{+Vision}
  & \textbf{+Both}
  & \textbf{ZS}
  & \textbf{+Memory}
  & \textbf{+Vision}
  & \textbf{+Both} \\
\midrule
o4‑mini-2025‑04‑16
  & 1.3{\scriptsize$\pm$0.6} & 1.3{\scriptsize$\pm$0.6} & 5.3{\scriptsize$\pm$2.1} & 5.3{\scriptsize$\pm$1.2}
  & 97.6{\scriptsize$\pm$29.2} & 115.1{\scriptsize$\pm$9.7} & 117.0{\scriptsize$\pm$6.4} & 120.6{\scriptsize$\pm$4.9} \\
\addlinespace
gemini‑2.5-Pro-03‑25(thinking)
  & 1.0{\scriptsize$\pm$0.0} & 1.0{\scriptsize$\pm$0.0} & 6.0{\scriptsize$\pm$2.0} & 4.3{\scriptsize$\pm$0.6}
  & 120.5{\scriptsize$\pm$3.9} & 118.0{\scriptsize$\pm$8.5} & 117.4{\scriptsize$\pm$5.8} & 117.3{\scriptsize$\pm$5.9} \\
\addlinespace
claude‑3-7-Sonnet-2025-0219 (thinking)
  & 0.0{\scriptsize$\pm$0.0} & 0.3{\scriptsize$\pm$0.6} & 0.7{\scriptsize$\pm$0.6} & 2.3{\scriptsize$\pm$1.5}
  & 114.2{\scriptsize$\pm$7.2} & 107.1{\scriptsize$\pm$5.1} & 115.3{\scriptsize$\pm$2.3} & 113.3{\scriptsize$\pm$3.1} \\
\addlinespace
llama-4-maverick-17b-128e-instruct-fp8
  & 0.0{\scriptsize$\pm$0.0} & 0.0{\scriptsize$\pm$0.0} & 0.0{\scriptsize$\pm$0.0} & 0.0{\scriptsize$\pm$0.0}
  & 44.6{\scriptsize$\pm$11.8} & 98.1{\scriptsize$\pm$3.8} &  73.7{\scriptsize$\pm$15.6} & 106.0{\scriptsize$\pm$3.8}\\
\addlinespace
claude-3-5-sonnet-20241022
  & 0.0{\scriptsize$\pm$0.0} & 0.0{\scriptsize$\pm$0.0} & 0.0{\scriptsize$\pm$0.0} & 0.0{\scriptsize$\pm$0.0}
  & 57.8{\scriptsize$\pm$16.4} & 102.5{\scriptsize$\pm$1.6} &  66.3{\scriptsize$\pm$9.6} & 108.2{\scriptsize$\pm$5.8} \\
\addlinespace
gpt-4o-2024-11-20
  & 0.0{\scriptsize$\pm$0.0} & 0.0{\scriptsize$\pm$0.0} & 0.0{\scriptsize$\pm$0.0} & 0.0{\scriptsize$\pm$0.0}
  & 70.4{\scriptsize$\pm$15.2} & 107.0{\scriptsize$\pm$6.3} &  73.3{\scriptsize$\pm$5.4} & 106.7{\scriptsize$\pm$3.5} \\
\bottomrule
\end{tabular}
}
\end{table}

%% file: text/4_experiments_2_correlations.tex
\subsection{Understanding Gaming Performance}
\label{sec:exp_analysis}

Games are designed to challenge human reasoning, perception, and planning abilities. Similarly, they require a combination of core LLM capabilities for strong performance. To identify which capabilities are involved, we evaluate the following 8 models: Claude-3.5-Sonnet, Claude-3.7-Sonnet-Thinking, Gemini-2.5-Pro-Preview, Llama-4-Maverick, GPT-4o, o1, o3, and o4-mini in 20 established benchmarks that span factual knowledge~\cite{vals2025mmlu, MMLU_paper, phan2024humanity, hle_leaderboard, hle_textonly_leaderboard, rein2023gpqa, gpqa_leaderboard}, physics~\cite{qiu2025phybench, EMMA_benchmark}, mathematics~\cite{EMMA_benchmark, math500_vals_leaderboard, patel2024aime, aime25_vals_leaderboard, livebench, livebench_leaderboard}, code generation~\cite{zhuo2024bigcodebench, aider_leaderboards, bigcodebench_leaderboard}, visual reasoning~\cite{vista_leaderboard, chatbot_arena_leaderboard, yue2023mmmu, mmmu_vals_leaderboard}, language understanding~\cite{sirdeshmukh2025multichallenge, multichallenge_leaderboard}, and puzzle solving~\cite{wang2025enigmaeval, enigmaeval_leaderboard, mazur2025nytconnections} to analyze their correlations with gaming performance. See Appx.~\ref{appendix:correlation} for the complete benchmark list. 

\textbf{Correlation Analysis}. 
We calculate Spearman’s rank correlation coefficient to assess alignment between model performance on games from \sysname and widely-used benchmarks. Results from Fig.~\ref{fig:corr_and_feature} (left) reveal positive correlations between several games and a few commonly used benchmarks.
In particular, Sokoban shows strong correlations with math and coding benchmarks. Tetris and 2048 closely align with pattern recognition tasks like EnigmaEval and NYT-connections. Candy Crush is notably related to coding, hinting at algorithmic reasoning. Ace Attorney strongly correlates with LiveBench-Language, suggesting a focus on narrative understanding distinct from other games. High variance games like \textit{Super Mario Bros.} (Table~\ref{tab:model-game-performance-merged}), is excluded from this analysis. More details are provided in Appx.~\ref{appendix:Mario_inlclude}.

\input{figure_text/correlations_and_feature}

\textbf{Latent Ability Decomposition.} 
To uncover relationships between the benchmarks and the capabilities of the model, we apply a low-rank matrix factorization to the model–benchmark performance matrix. This decomposes each LLM as a vector in latent ability space and each benchmark, including our games, as a sparse, weighted combination of these abilities. 

Due to the non-uniqueness of low-rank matrix factorization and entanglement of features in vector spaces, the resulting components are not necessarily human-interpretable. However, meaningful patterns can still be observed. As shown in Fig.~\ref{fig:low_rank}, Feature 1 aligns with language and multi-task knowledge (e.g., MMLU-Pro, MultiChallenge). Feature 2 captures coding capabilities (e.g., BigCodeBench, Aider). Feature 3 corresponds to symbolic and puzzle-solving skills (e.g., EnigmaEval, NYT Connections). Feature 4 reflects physical reasoning (e.g., PHYBench, EMMA).

Each game in \sysname loads on different subsets of these latent abilities. \textit{Sokoban} emphasizes symbolic and physical reasoning (Features 3 and 4), while \textit{Ace Attorney} strongly engages long-context language reasoning (Feature 1). \textit{Tetris} and \textit{2048} mainly represents mathematical and spatial reasoning (Features 3), and \textit{Candy Crush} reflects visual pattern recognition with moderate ties to coding (Features 2 and 3). This decomposition aligns closely with the correlation study above, while also suggests that games cover compositional capabilities rather than isolated skills.

\paragraph{Linear Modeling.}

\input{table/polynomial_model_linear}

We also employ linear modeling to predict game ranking based on model capabilities across different categories (benchmark choices are provided in Appx.~\ref{appendix:correlation}; technical details and more results on polynomial models fitted with different categorical combinations are presented in Appx.~\ref{appendix:polynomial_modeling}). When using the following five benchmark categories: language, physics understanding, visual understanding, mathematics, and coding as explanatory variables, Table~\ref{tab:linear_model} reveals how each game decomposes into familiar skill domains.

In linear modeling, our results show long-horizon games like Sokoban, Tetris and 2048's rankings are driven primarily by math and coding performance. Games requiring spatial reasoning, like Sokoban, Candy Crush and Super Mario Bros, align closely with the physics‑understanding and visual understanding benchmark. Text-heavy narrative games like Ace Attorney are dominated by language-related benchmarks. 
Beyond linear modeling, we find in most cases (an example is provided in Table~\ref{tab:polynomial_model}), polynomial models lack interpretability because the feature categories are not strictly orthogonal, leading to the emergence of cross terms that complicate attribution. Similarly, we do not consider higher-order polynomial models.

%% file: figure_text/correlations_and_feature.tex
\begin{figure}[t]
    \centering
    \begin{minipage}[c]{0.49\linewidth}
        \centering
        \includegraphics[height=5cm]{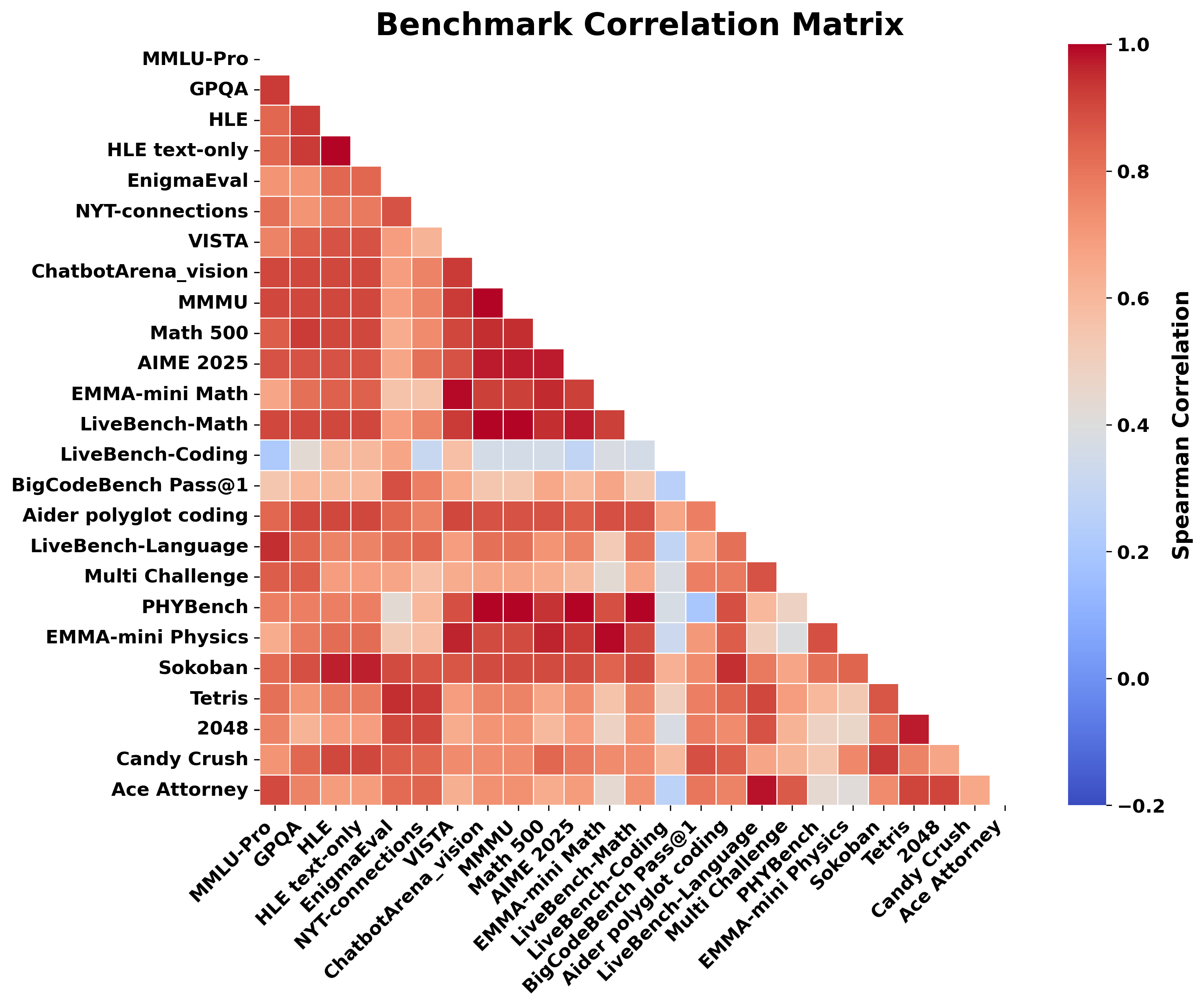}
        \subcaption{Spearman Correlation among \sysname{} and other benchmarks.}
        \label{fig:corr_only}
    \end{minipage}
    \hfill
    \begin{minipage}[c]{0.49\linewidth}
        \centering
        \includegraphics[height=5cm]{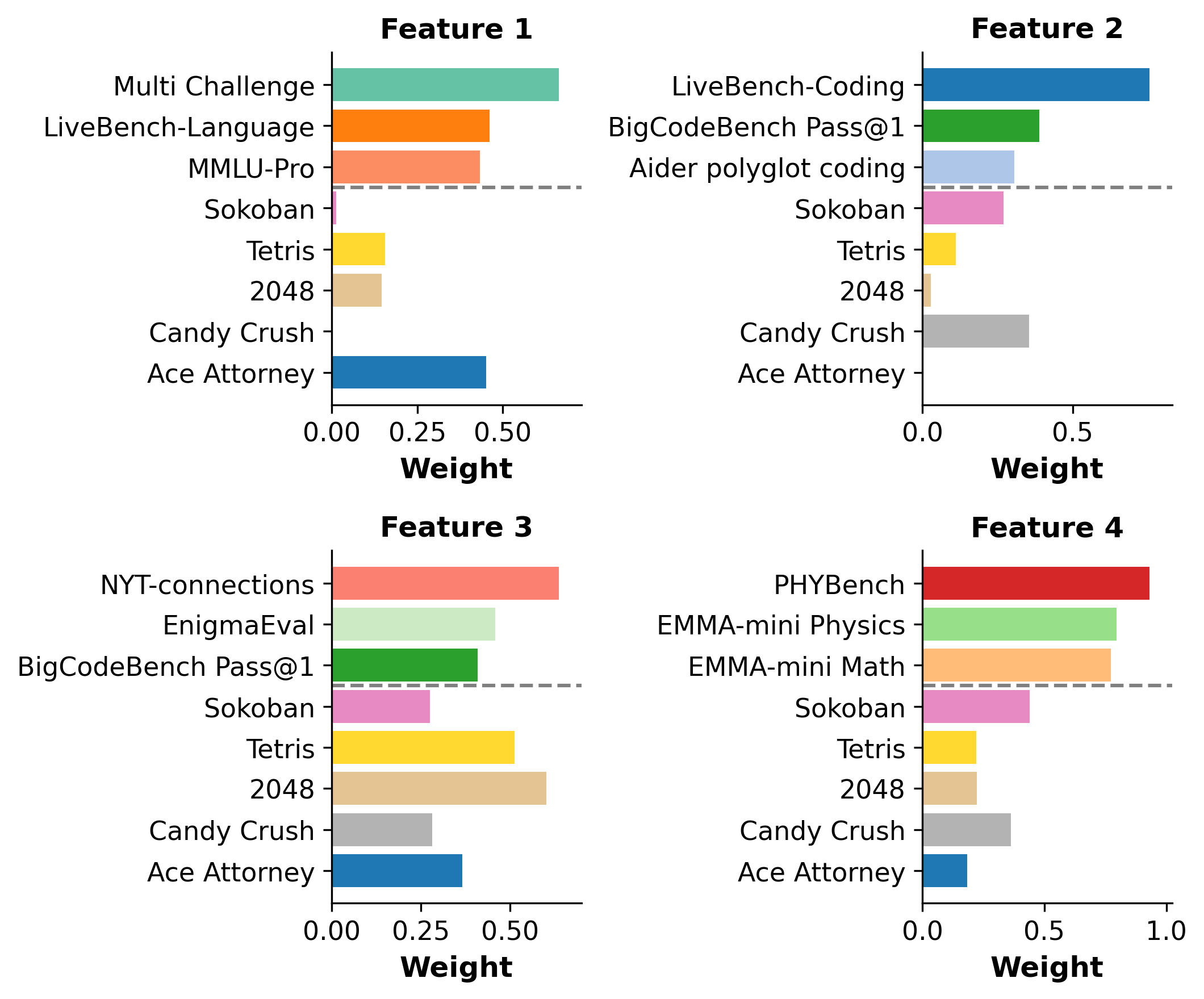}
        \subcaption{Top‑weight benchmarks under each feature after low‑rank decomposition.}
        \label{fig:low_rank}
    \end{minipage}
    \caption{Correlation Analysis and Latent Feature Decomposition Among Benchmarks.}
    \label{fig:corr_and_feature}
\end{figure}

%% file: table/polynomial_model_linear.tex
\begin{table}[h]
  \centering
  \scriptsize
  \caption{Learned weights for game ranking prediction using a linear model, where $r$ and RE denote for Pearson’s r and mean-normalized residual errors respectively. For the chosen set of categories, the linear models can hardly predict SMB and 2048 rankings correctly.}
  \vspace{5pt}
  \resizebox{0.9\textwidth}{!}{%
  \begin{tabular}{l|ccccc|ccc}
    \toprule
    \textbf{Game} & \textbf{Language} & \textbf{Physics} & \textbf{Visual} &
    \textbf{Math} & \textbf{Coding} & Offset & $r$ & RE \\
    \midrule
    Sokoban          & 0.408 & 1.011 & 0.810 & \textbf{2.160} & \textbf{2.206} & 0.297 & 0.930 & 0.4758 \\
    Tetris           & 1.759 & 0.001 & 1.356 & \textbf{1.979} & \textbf{2.222} & 0.825 & 0.825 & 0.814 \\
    Ace~Attorney     & \textbf{3.392} & 0.000 & 0.962 &  2.430 & 0.004 & 0.853 & 0.853 & 0.800 \\
    Super~Mario~Bros & 0.275 & \textbf{1.905}      & 0.000      & 0.597 & 0.000 & \textbf{2.940} & \underline{0.295} & 1.377 \\
    2048             & 0.008  & 0.332 & 0.000   & \textbf{1.880}      & 0.000      & \textbf{3.130} & \underline{0.248} & 1.467\\
    Candy~Crush      & 0.678 & \textbf{3.444} & 0.002      & 1.275 & 2.456 & 0.088      & 0.864 & 0.730 \\
    \bottomrule
  \end{tabular}%
  \label{tab:linear_model}
  }
\end{table}

%% file: text/4_experiments_3_train.tex
\subsection{Training Generalizability Study}
\label{sec:exp_train}
In this section, we investigate the generalization effects of game-based training, providing empirical evidence that optimizing on game benchmarks can lead to broader improvements in LLM capabilities.

\textbf{Experiment Settings.} Our training experiments involve two games, simplified Sokoban and Tetris. For Sokoban, we trained on gym-sokoban~\citep{SchraderSokoban2018} environment on $6\times 6$ board, and for Tetris, we build on TetrisRL~\citep{tetrisrl} with a $4\times 4$ board and 2 simple tetrominoes types (1x2 and 2x1 blocks). We finetuned Qwen2.5-7B-Instruct~\citep{qwen2.5} under RAGEN~\citep{wang2025ragen}, a multi-turn RL framework for LLMs, with prompts that include explicit thinking tokens.
We evaluate the performance on various downstream tasks, including same-game with different settings, cross-game, planning, math/code, and agentic tasks, using the same prompting formats as training. Note that math and coding are not directly applicable to multi-turn settings, so we simulate multi-turn behavior by allowing self-correction attempts using feedback ``Incorrect. Please think again''.
Detailed training settings, hyperparameters, and additional experiments on task and prompt variations are provided in Appx.~\ref{appendix:training}. 

\textbf{Cross-Game and Planning Generalization.} 
As shown in Table~\ref{tab:train_sokoban}, training on Sokoban yields strong gains in cross-game and spatial reasoning performance—improving results on larger Sokoban boards, boosting Blocksworld 1D/2D by at least 10\%, and achieving up to 8\% zero-shot improvement on Tetris. Similarly, training on Tetris also enhances performance on cross-game and planning tasks. These results suggest that the spatial reasoning and planning heuristics learned during training transfer effectively across board rules and settings considered in our study.





\textbf{Math, Coding, and Agentic-Task Generalization.}
Despite gains in board games and planning, neither Sokoban nor Tetris training transfers to tasks like GSM8K, and BIRD, indicating that spatial reasoning and search heuristics alone are insufficient for the reasoning skills required in math and coding. On the multi‐turn WebShop benchmark, game‐trained models achieve at least 6\% improvement, demonstrating that grid‐game–derived skills can benefit real‐world decision‐making.

\input{table/train}

%% file: table/train.tex
\begin{table}
\footnotesize
    \centering
    \vspace{-10pt}
    \caption{Model performance on diverse tasks after training on simplified Sokoban and Tetris.}
    \vspace{0.25em}
    \label{tab:train_sokoban}
    \resizebox{\linewidth}{!}{%
    \begin{tabular}{c|c|c|c|c|c|c|c|c|c|c|c}
    \toprule
        &\multicolumn{4}{c|}{\textbf{Games}} & \multicolumn{3}{c|}{\textbf{Planning}} & \multicolumn{3}{c|}{\textbf{Math/Coding}} & \multicolumn{1}{c}{\textbf{Agentic}} \\
        \multirow{1}{*}{\textbf{Model}} & \multicolumn{2}{c|}{\textbf{Sokoban}} & \multicolumn{2}{c|}{\textbf{Tetris}} & \multicolumn{3}{c|}{\textbf{Blocksworld}} & \multicolumn{2}{c|}{\textbf{GSM8K}} & \multicolumn{1}{c|}{\textbf{BIRD}} & \multicolumn{1}{c}{\textbf{WebShop}} \\
         & $6\times 6$ & $8\times 8$ & 1 type & 2 types & Text & 1D & 2D & 1 turn & 5 turns & 1 turn  & \\\midrule
        Qwen-7B-Instruct & 11.3 & 5.9 & 9.0 & 4.7 & 64.7 & 17.9 & 9.0  & \textbf{89.5} & \textbf{95.3} & \textbf{25.0} & 7.0 \\
        Ours (Sokoban) &  \textbf{24.2} & \textbf{9.0} & 17.6 & 5.1  & 64.1 & \textbf{32.7} & \textbf{29.5} & 89.0 & 94.1 & 17.5 & \textbf{19.1} \\
        Ours (Tetris) &  13.3& 6.7& \textbf{49.5} & \textbf{14.5} & \textbf{66.2} &21.5& 15.2& 89.0& 93.4 & 19.8& 13.4\\
         \bottomrule
    \end{tabular}%
    }
    \vspace{-5pt}
\end{table}

%% file: text/2_related.tex
\section{Related Work}

\paragraph{Games as AI Testbeds.}
Games have long served as foundational benchmarks in AI research, particularly in reinforcement learning. From TD-Gammon~\cite{tesauro1995temporal} to AlphaGo~\cite{silver2017mastering}, they have offered controlled environments for studying planning and sequential decision-making. OpenAI Gym~\cite{brockman2016openai} further standardized this paradigm by providing a unified interface for interacting with diverse game environments.
More recently, games have been adopted to evaluate LLM agents on specific domains, such as grid-based games~\cite{nasir2024gametraversalbenchmark}, open-ended strategy games~\cite{hopkins2025factorio}, or murder-mystery games~\cite{xie2024whodunitbench}. Others evaluate natural language reasoning through text-based or conversational games~\cite{costarelli2024gamebench,hudi2025textgames,qiao2023gameeval,hu2024gamearena}, lacking visual understanding. Multimodal gaming benchmark like BALROG~\cite{paglieri2024balrog} assess planning in rich game environments, but focus primarily on qualitative observations on game-related LLM abilities.
Instead, we link games to LLM abilities through correlation analysis and multi-turn training.
%

\paragraph{LLM Agentic Benchmarks.}
Current agentic benchmarks tend to focus on domain-specific tasks—such as code editing~\cite{jimenez2023swe}, web browsing~\cite{zhou2023webarena,he2024webvoyager}, API control flows~\cite{trivedi2024appworld}, GUI control~\cite{agashe2024agent} or system operations~\cite{xie2024osworld}. Recent efforts also explored broader evaluation across multiple aspects~\cite{liu2024agentbench,mialon2023gaia}.
While these benchmarks provide valuable insights into specialized domains, they often require significant engineering effort to construct, limiting their scalability. In contrast, games offer a more scalable and skill-diverse environment for evaluating general agentic behavior.

\paragraph{Fine-tuning LLMs via Games.}
Recent work~\cite{guo2025deepseek} has shown that post-training LLMs with verifiable rewards in domains like math and code can produce strong reasoning models. Games similarly offer structured, verifiable rewards in sequential decision makings, making them a natural testbed for training decision-capable agents. Recent efforts such as RAGEN~\cite{wang2025ragen} and RL4VLM~\cite{zhai2024fine} fine-tune LLMs in game-based environments, but they do not systematically examine generalization to broader planning or agentic benchmarks. In contrast, we trained on well-known games and evaluated whether this can transfer to both in-domain and out-of-domain tasks.




%% file: text/5_conclusion.tex
\section{Conclusion}


We introduce \sysnamenospace, the first agentic benchmark for evaluating LLMs on games with and without gaming harness support. \sysname leverages a gaming harness composed of agentic modules to better distinguish state-of-the-art models. Our benchmark identifies and addresses data contamination through a series of mitigation, and it reduces prompt variance by integrating a two-stage prompt optimization using DSPy. Additionally, we present \sysname can be regarded as compositions of existing core LLMs capabilities often evaluated in isolation, supported by a comprehensive correlational study. Lastly, we show performance improvement on \sysname through RL training can transfer to both in-domain tasks (e.g. variations of the same game, similar game types) and some out-of-distribution tasks (e.g. BlocksWorld and WebShop).

%% file: text/7_appendix_1_train.tex
\appendix
\onecolumn

\section{Training Details}
\label{appendix:training}

\subsection{Training Framework}
Sokoban and Tetris training is based on the StarPO-S (State-Thinking-Actions-Reward Policy Optimization - Stable) framework introduced in the RAGEN system~\cite{wang2025ragen}, which extends reinforcement learning to multi-turn agent settings by optimizing full interaction trajectories. 
LLM responses are structured in two formats: with <think> tokens, where the LLM generates intermediate reasoning traces followed by actions enclosed in <answer> tokens; and without <think> tokens, where the LLM directly outputs actions within <answer> tokens. The only exception is GSM8K, where both thinking and answer tokens are omitted to maintain consistency with standard evaluation prompts on math datasets.
Each task is implemented as an environment that provides rule-based cumulative rewards over entire rollouts. 
Optimization is performed using Proximal Policy Optimization (PPO)~\cite{schulman2017ppo}, applied at the token level with clipped probability ratios. Training proceeds by generating multiple trajectories from a batch of prompts. The LLM (policy model) is updated using estimated advantages computed by a separately trained critic model.
\begin{tcolorbox}[title=Example Response Formats, colback=gray!3, colframe=gray!80, fonttitle=\bfseries, left=1mm, right=1mm, top=1mm, bottom=1mm, boxrule=0.5pt]
With thinking tokens:
\texttt{<think>...</think><answer>action || action</answer>}

Without thinking tokens:
\texttt{<answer>action || action</answer>}
\end{tcolorbox}

\paragraph{Hyperparameter Settings.}
The maximum model response length per turn is 400 tokens during training, and 1024 tokens during evaluation. During training, the temperature is set to 1.0 to encourage exploration with 16 trajectories generated per sample; at inference, we use greedy sampling to ensure reproducibility. We train Qwen-2.5-7B-Instruct for 200 steps, with a batch size of 32 samples. We then report the performance on a separate test set on the checkpoint saved at step 100 or 200, whichever leads to a better performance on the validation set on the trained task.
The test set size is 256 samples for all tasks, except for Blocksworld, where the complete set of 156 possible configurations is evaluated exhaustively.
To improve training stability in long-horizon tasks, we follow most of the default training settings from StarPO-S.
We filter out low-variance rollouts and retain only the top 25\% most informative trajectories~\cite{settles2009active}, selected based on the standard deviation of rewards across trajectories generated from the same prompt.
We remove the KL penalty term from PPO, and apply asymmetric clipping (with lower and upper bounds set to 0.2 and 0.28 respectively) to encourage learning from high-reward rollouts~\cite{yu2025dapo}.
Experiments were conducted using 8 NVIDIA H100 GPUs (80 GB each) with model parallelism. Batch size and response length were adjusted to fit within hardware memory constraints, as reported above. Each training instance takes approximately 2–3 hours for 100 steps.

\subsection{Tasks}
We consider the following tasks in this analysis. We trained Qwen2.5-7B-Instruct~\cite{qwen2.5} on $6\times 6$ Sokoban and 2-block-type Tetris separately, and evaluated on all the below tasks. LLM is allowed to output up to 5 actions per response. The number of maximum response turns varies across tasks, as specified below. Detailed prompts are listed in \ref{appendix:training_prompts}.
\begin{itemize}
    \item SimpleSokoban~\cite{SchraderSokoban2018,wang2025ragen}: We consider simplified Sokoban with $6\times6$ or $8\times 8$ grids, containing only a single box. The LLM is allowed up to 5 response turns to complete the task and in total 10 actions to complete the task.
    \item SimpleTetris~\cite{tetrisrl}: 
    This task involves a simplified version of Tetris on a $4 \times 4$ board with only one or two types of blocks. In the 1-type setting, all blocks are $1 \times 1$. In the 2-type setting, we use $1 \times 2$ and $2 \times 1$ blocks. We disable rotation operations for both settings. The LLM is allowed up to 10 response turns and up to 40 actions in total. 
    \item Blocksworld~\cite{valmeekam2023planbench}: Blocksworld is a classical planning task that stacks blocks on a table, where a block can only be moved if no other block is on top of it. We focus on the 3-block setting and evaluate three different formats to represent states: natural language, 1D list, and 2D table representations. We allow up to 10 response turns and up to 20 actions in total. 
    \begin{itemize}
        \item \textit{Natural Language:} ``Block 1 is on top of block 3, block 3 is on top of block 2, and block 2 is on the table.''
        \item \textit{1D List:} ``[3, 0, 2]'' – indicating block 1 is on top of block 3, block 2 is on the table, and block 3 is on top of block 2.
        \item \textit{2D Table:} See the prompt example for Blocksworld in Appendix~\ref{appendix:training_prompts}.
    \end{itemize}
    \item GSM8K~\cite{cobbe2021gsm8k}: 
    This dataset consists of grade school math word problems. We use chain-of-thought prompting (e.g., “Let’s think step by step”) to elicit reasoning from the LLM. Unlike other tasks that require multiple turns, LLMs will give the answer in a single response. To adapt it to our multi-turn setting, if the LLM provides a correct answer, the session ends. If the answer is incorrect, we append a user message—“Incorrect. Please think again.”—to trigger a new response. We report both single-turn and 5-turn success rates.
    \item BIRD~\cite{bird}: This task evaluates text-to-SQL translation. We report only the single-turn success rate for this task.
    \item WebShop~\cite{yao2022webshop}: WebShop simulates an online shopping environment where the model can search for items and complete purchases. The LLM is allowed up to 10 response turns and up to 30 actions in total to complete the order.
\end{itemize}

\subsection{Thinking Tokens}

\input{table/train_prompt}

Table~\ref{tab:train_sokoban} shows that training on Sokoban and Tetris improves cross-game, planning, and agentic task performance when models are explicitly prompted to "think" using thinking tokens. To test whether this improvement persists under different prompting formats, Table~\ref{tab:train_prompt} compares model generalization with and without thinking tokens at training and inference.
The impact varies across tasks: for same-game and WebShop, performance is highest when the prompting format is consistent between training and inference. For Tetris and planning, the effects are mixed. Notably, in planning tasks, models trained with thinking tokens mostly outperform others regardless of test-time prompting, highlighting the importance of training models to ``think'' to perform well in planning.

\subsection{Effectiveness of Game-Only, Math-Only, and Mixed Training}
As shown in Section~\ref{sec:exp_train}, training solely on games such as Sokoban or Tetris improves planning and spatial reasoning tasks, but yields limited benefits for math and coding. To further explore cross-domain generalization, we evaluate two additional training settings: (1) training solely on GSM8K, and (2) training on an equal mixture of Sokoban and GSM8K. The performance on Sokoban, Tetris and Blocksworld are evaluated using thinking tokens, while GSM8K is prompted using a chain-of-thought (CoT) style. The resulting generalization performance is compared in Figure~\ref{fig:success_curves}. Success rates are evaluated for every 10 steps during Sokoban training, and a smoothed trend line is shown using a running average with a centered window.

Training solely on Sokoban leads to the strongest gains in Sokoban itself, as well as significant improvements in other spatial-reasoning environments such as Blocksworld-2D and Tetris (1-type). However, it provides no improvement on GSM8K, which focuses on math reasoning. In contrast, models trained only on GSM8K achieve the best performance on GSM8K but show minimal or no generalization to Sokoban, Blocksworld, or Tetris. The mixture-trained agent exhibits performance in between the two extremes. Notably, it starts out matching the GSM8K-only model on GSM8K, but plateaus earlier and underperforms slightly by the end of training. This suggests that while mixture training promotes moderate generalization across domains, it may compromise peak performance on domain-specific tasks. Interestingly, the mixture-trained model achieves comparable gains to GSM8K-only training within the first 100 steps on GSM8K, indicating mixture training has potential for reducing the number of math-specific samples required during the early training phase.

\input{figure_text/success_curves}

\subsection{Prompts.}
\label{appendix:training_prompts}
We use the following list of prompts for training and evaluation. Specifically, each prompt explicitly informs the LLM of the expected reasoning format, the maximum number of actions allowed per response, and the response length limit, and reward. Notably, the length limit stated in the prompt is intentionally shorter than the actual enforcement threshold to discourage overly long responses. 
\begin{tcolorbox}[colback=gray!10, colframe=deepblue, title=SimpleSokoban, fonttitle=\bfseries, sharp corners=south]
\textbf{\textcolor{blue}{Prompt:}}\\
You are solving the Sokoban puzzle. You are the player and you need to push all boxes to targets. When you are right next to a box, you can push it by moving in the same direction. You cannot push a box through a wall, and you cannot pull a box. The answer should be a sequence of actions, like <answer>Right || Right || Up</answer>.\\
The meaning of each symbol in the state is:\\
$\#$: wall, $\_$: empty, O: target, $\sqrt{}$: box on target, X: box, P: player, S: player on target\\
Your available actions are:\\
Up, Down, Left, Right\\
You can make up to 10 actions, separated by the action separator `` || ''\\

Turn 1:\\
State:\\
\#\#\#\#\#\#\\
\#\#\#\#\#\#\\
\#O\#\#\#\#\\
\#XP\#\#\#\\
\#\_\_\#\#\#\\
\#\#\#\#\#\#\\
You have 10 actions left. Always output: <think> [Your thoughts] </think> <answer> [your answer] </answer> with no extra text. Strictly follow this format. Max response length: 100 words (tokens).

\vspace{1em}
\textbf{\textcolor{blue}{Prompt for next turn:}}\\
Reward:\\
-0.1\\

Turn 2:\\
State:\\
\textcolor{blue}{\{Current State\}}\\
You have 9 actions left. Always output: <think> [Your thoughts] </think> <answer> [your answer] </answer> with no extra text. Strictly follow this format. Max response length: 100 words (tokens).

\end{tcolorbox}

\begin{tcolorbox}[colback=gray!10, colframe=deepblue, title=Blocksworld, fonttitle=\bfseries, sharp corners=south]
\textbf{\textcolor{blue}{Prompt:}}\\
You are solving Blocksworld problems. Each state is represented by a graphical stack of blocks. Each state is represented by a graphical stack of blocks. There are 3 blocks: block 1, block 2, and block 3. You can only move a block if there is no other block on top of it. Each action should be formatted as:"move X to Y", where X is the block number you're moving, and Y is either 0, meaning move the block to the table; or another block number, meaning move it on top of that block.\\
A state is like this:\\
\ \_\ \\
|2|\\
\ ¯\ \\
\ \_\ \ \_\ \\
|1||3|\\
\ ¯\ \ ¯\ \\
This means: Block 2 is on top of block 1, and Block 1, 3 is on the table. Another example:\\
\ \_\ \\
|1|\\
\ ¯\ \\
\ \_\ \\
|3|\\
\ ¯\ \\
\ \_\ \\
|2|\\
\ ¯\ \\
This means: Block 1 is on top of block 3, Block 3 is on top of block 2, and Block 2 is on the table.\\ 
You are given the current state and the goal state, and you need to find a sequence of actions to move the blocks to the goal state. The answer should be a sequence of actions, like <answer>(move 2 to 0) || (move 1 to 2)</answer>\\
Turn 1:\\
State:\\
Graphical representation (Current):\\
\textcolor{blue}{\{Current State\}}\\
Graphical representation (Goal):\\
\textcolor{blue}{\{Goal State\}}\\
You have 20 actions left. Always output: <think> [Your thoughts] </think> <answer> [your answer] </answer> with no extra text. Strictly follow this format. Max response length: 100 words (tokens).

\vspace{1em}
\textbf{\textcolor{blue}{Prompt for next turn:}}\\
Reward:\\
-1\\

Turn 2:\\
State:\\
Graphical representation (Current):\\
\textcolor{blue}{\{Current State\}}\\
Graphical representation (Goal):\\
\textcolor{blue}{\{Goal State\}}\\
You have 19 actions left. Always output: <think> [Your thoughts] </think> <answer> [your answer] </answer> with no extra text. Strictly follow this format. Max response length: 100 words (tokens).
\end{tcolorbox}

\begin{tcolorbox}[colback=gray!10, colframe=deepblue, title=SimpleTetris, fonttitle=\bfseries, sharp corners=south]
\textbf{\textcolor{blue}{Prompt:}}\\
You are solving a simplified version of Tetris puzzle. The game board is a 4*4 grid, and blocks fall one at a time from the top. Your goal is to fill the last row completely with blocks to get points, with the state such as\\
\_\_\_\_\\
\_\_\_\_\\
\_\_\_\_\\
XXXX\\
You can only move the current block left, right, or down. The answer should be a sequence of actions, like <answer>Left || Right || Down</answer> Hint: If we see empty space (\_) in the last row, try to left/right move the current block to let it drop down to the empty space.\\
The meaning of each symbol in the state is:\\
\#: filled spaces, \_: empty, X: current block
Your available actions are:\\
Left, Right, Down\\
You can make up to 40 actions, separated by the action separator `` || ''\\

Turn 1:\\
State:\\
XX\_\_\\
\_\_\_\_\\
\_\_\_\_\\
\_\_\_\_\\
You have 20 actions left. Always output: <think> [Your thoughts] </think> <answer> [your answer] </answer> with no extra text. Strictly follow this format. Max response length: 100 words (tokens).

\vspace{1em}
\textbf{\textcolor{blue}{Prompt for next turn:}}\\
Reward:\\
-0.1\\

Turn 2:\\
State:\\
\_\_\_\_\\
XX\_\_\\
\_\_\_\_\\
\_\_\_\_\\
You have 19 actions left. Always output: <think> [Your thoughts] </think> <answer> [your answer] </answer> with no extra text. Strictly follow this format. Max response length: 100 words (tokens).

\end{tcolorbox}





\begin{tcolorbox}[colback=gray!10, colframe=deepblue, title=GSM8K, fonttitle=\bfseries, sharp corners=south]
\textbf{\textcolor{blue}{Prompt:}}\\
You are solving Math problems. Let's think step by step. Always put the answer in integer at the end of your response. Question:
Frankie watches TV after he finishes his homework every night. On Monday and Tuesday, he watched a 1-hour episode of his favorite show each night. On Wednesday, he watched a few episodes of a 30-minute show. On Thursday, he finished homework early and watched a 1-hour episode and a 30-minute show. On Friday, he got to stay up late for the weekend, so he watched two 1-hour episodes. If he watched 7 hours of TV in all, how many 30-minute episodes did he watch on Wednesday?

\vspace{1em}
\textbf{\textcolor{blue}{Prompt for next turn:}}\\
Reward: \\
-0.1\\
Question: \\
Incorrect. Please think again.
\end{tcolorbox}

\begin{tcolorbox}[colback=gray!10, colframe=deepblue, title=BIRD, fonttitle=\bfseries, sharp corners=south]
\textbf{\textcolor{blue}{Prompt:}}\\

Translate the question into SQL given database information. Replay with one fenced block:
\begin{verbatim}
```sql
{SQL query here}
```
\end{verbatim}
Turn 1:
State:\\
DB schema:\\
\textcolor{blue}{\{Table Schema 1\}}\\
\textcolor{blue}{\{Table Schema 2\}}\\
\textcolor{blue}{\{Table Schema 3\}}\\
...\\

Question: How many customers share the most common last name?\\
You have 3 actions left. Always output: <think> [Your thoughts] </think> <answer> [your answer] </answer> with no extra text. Strictly follow this format. Max response length: 1024 words (tokens).

\vspace{1em}
\textbf{\textcolor{blue}{Prompt for next turn:}}\\
Reward:\\
-0.1\\

Turn 2:\\
State:
DB schema:\\
\textcolor{blue}{\{Table Schema 1\}}\\
\textcolor{blue}{\{Table Schema 2\}}\\
\textcolor{blue}{\{Table Schema 3\}}\\
...\\
DB query output from last turn:\\
\textcolor{blue}{\{Query output/error\}}\\

You have 2 actions left. Always output: <think> [Your thoughts] </think> <answer> [your answer] </answer> with no extra text. Strictly follow this format. Max response length: 100 words (tokens).

\end{tcolorbox}








\begin{tcolorbox}[colback=gray!10, colframe=deepblue, title=Webshop, fonttitle=\bfseries, sharp corners=south]
\textbf{\textcolor{blue}{Prompt:}}\\
You're a helpful assistant. You are browsing an online shop. Based on the instruction, find the product that matches the production description. You need to iteratively take action, read the website and decide what action to take next until finding the product. Available actions depends on the page: in the search page you can search keywords, in the search result page you can click an item url or click[next >] to navigate to next page, in the product page you can click[description] or click[features] to see the details, click[blue] or click[x-large] to choose size and colors, click[buy now] when you decided to buy the product, click[back to search] to return to search page. Example process: I need a gingko light and 20x20 pillow cover that is hand painted. First search[gingko light 20x20 pillow cover hand painted], answer format: <answer>search[blanket with fleece throw]</answer>. Valid answer is search[<keywords>] or click[<clickable>].\\

Turn 1:\\
State:\\
Instruction:\\
\textcolor{blue}{\{target task with constraints\}}\\
Available actions: search[<content>]\\
You have 10 actions left. Always output: <think> [Your thoughts] </think> <answer> [your answer] </answer> with no extra text. Strictly follow this format. Max response length: 100 words (tokens).\\

\vspace{1em}
\textbf{\textcolor{blue}{Prompt for next turn:}}\\
Reward:\\
0.0\\
Turn 2:\\
State:\\
Instruction:\\
\textcolor{blue}{\{target task with constraints\}}\\
\textcolor{blue}{\{web page description\}}\\
Available actions:\\
\textcolor{blue}{\{all available actions\}}\\
You have 9 actions left. Always output: <think> [Your thoughts] </think> <answer> [your answer] </answer> with no extra text. Strictly follow this format. Max response length: 100 words (tokens).\\

\end{tcolorbox}













%% file: table/train_prompt.tex
\newcolumntype{P}[1]{>{\centering\arraybackslash}p{#1}}

\begin{table}
    \centering
    \caption{Impact of thinking‐token prompting on model generalization across games, planning, and agentic tasks.}
    \label{tab:train_prompt}
    \footnotesize
    \resizebox{0.9\linewidth}{!}{%
    \begin{tabular}{c|P{0.9cm}|P{0.9cm}|c|c|c|c|c|c|c|c}
    \toprule
        & \multicolumn{2}{c|}{\multirow{2}{*}{\textbf{Thinking Token}}} & \multicolumn{4}{c|}{\textbf{Games}} & \multicolumn{3}{c|}{\textbf{Planning}} & \multicolumn{1}{c}{\textbf{Agentic}} \\
        \multirow{1}{*}{\textbf{Model}} &\multicolumn{2}{c|}{} & \multicolumn{2}{c|}{\textbf{Sokoban}} & \multicolumn{2}{c|}{\textbf{Tetris}} & \multicolumn{3}{c|}{\textbf{Blocksworld}} & \multicolumn{1}{c}{\textbf{WebShop}} \\
        & Train & Test &  $6\times 6$ & $8\times 8$ & 1 type & 2 types & Text & 1D & 2D & \\\midrule
        Qwen-7B-Instruct & - & \cmark & 
        11.3 & 5.9 & 9.0 & 4.7 & 64.7 & 17.9 & 9.0 & 7.0\\
        Ours (Sokoban) & \cmark & \cmark &  \textbf{24.2} & \textbf{9.0} & 17.6 & 5.1  & 64.1 & \textbf{32.7} & \textbf{29.5} & \textbf{19.1}\\
        Ours (Sokoban)  & \xmark & \cmark & 10.9 & 8.6 & \textbf{26.9} & \textbf{7.0} & \textbf{68.6} &  21.2 & 15.4 & 10.5  \\\midrule
        Qwen-7B-Instruct & - & \xmark & 21.1 & 7.8 & 1.6 & 11.3 &34.0 & 17.3 & 12.8 & 37.9 \\
        Ours (Sokoban) & \xmark & \xmark & \textbf{24.6} & \textbf{12.9} & \textbf{34.4} & \textbf{19.5} & \textbf{44.9} & 23.1 & 13.5 & \textbf{43.8}\\
        Ours (Sokoban)  &  \cmark & \xmark & 19.5 & 5.1 & 32.8 & 12.1 & \textbf{44.9} & \textbf{30.1} & \textbf{23.1} & 41.8 \\
    \bottomrule
    \end{tabular}%
    }
\end{table}


%% file: figure_text/success_curves.tex
\begin{figure}
    \centering
    \includegraphics[width=\linewidth]{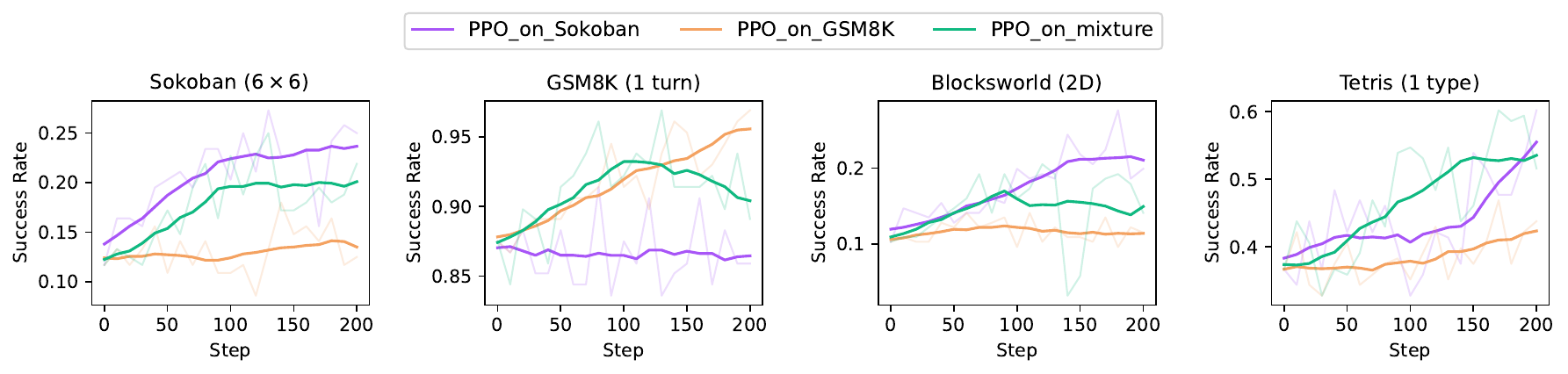}
    \caption{Success rates during training across different evaluation tasks, with models trained on Sokoban, GSM8K, or a half-half mixture of both.}
    \label{fig:success_curves}
\end{figure}

%% file: text/7_appendix_2_data_contamination.tex
\section{Data contamination Study}
\label{appendix:data_contamination}
We evaluate two types of potential data contamination in \sysname: \textbf{vision-level} and \textbf{text-level}. Our goal is to determine whether pretrained LLMs rely on memorized assets instead of real-time reasoning.

\subsection{Vision-Level Contamination: Super Mario Bros}

\textbf{Setup.} We extracted the first ten RGB frames from SMB Level 1‑1 and randomly shuffled their order. Each model was then prompted to reconstruct the original temporal sequence 15 times. Reconstruction quality was measured by pairwise frame‐order accuracy, Kendall’s $\tau$ rank coefficient~\cite{kendall1938new}, and Rank-Biased Overlap (RBO)~\cite{webber2010similarity}. Finally, we computed the Pearson's and Spearman's correaltion coefficients between these alignment metrics and the overall performance rank of each model.

\textbf{Models tested:} Claude-3.5-Sonnet, Claude-3.7-Sonnet-Thinking, Gemini-2.5-pro-Preview, o4-mini, o3, LLaMA-4-Maverick.

\textbf{Results.} The pairwise accuracy remains relatively low overall, with the highest accuracy reaching only around 30\%, as shown in Table~\ref{tab:vision_contamination}. Notably, both the beginning and final positions exhibit relatively high accuracy in terms of ordering, whereas the middle positions perform significantly worse, as illustrated in Fig.~\ref{fig:position-wise_heatmap}. Beyond pairwise accuracy, we compute Kendall’s rank correlation coefficient and Rank‐Biased Overlap (RBO) to evaluate how well each model’s predicted frame order aligns with the ground‐truth temporal sequence. As shown in Fig.~\ref{fig:Kendall_RBO_Bar_Plots}, all models exhibit positive correlation, although only Gemini-2.5-pro-preview, o3, and o4-mini achieve moderate agreement with ground truth, while the remaining models show weak alignment. To further quantify alignment strength, we normalize each metric by computing the percentage of the perfect score, using the formula $(\text{value} - \text{random}) / (\text{perfect} - \text{random})$, which reveals similar ranking patterns for both Kendall’s $\tau$ and RBO. To test whether alignment quality is predictive of general model performance, we compute both Pearson and Spearman correlations between the alignment metrics and the performance ranks of the models. Kendall’s $\tau$ shows a moderate negative Pearson correlation with performance rank ($r = -0.7089$, $p = 0.1148$, testing the null hypothesis of no linear association) and a moderate negative Spearman correlation ($\rho = -0.5429$, $p = 0.2657$, testing the null hypothesis of no monotonic association). RBO shows a weaker Pearson correlation ($r = -0.3847$, $p = 0.4515$, testing the null hypothesis of no linear association) and a moderate negative Spearman correlation ($\rho = -0.6571$, $p = 0.1562$, testing the null hypothesis of no monotonic association). Kendall’s $\tau$ and RBO remain highly correlated (Pearson $r = 0.8772$, $p = 0.0217$, testing the null hypothesis of no linear association). Despite this internal consistency, the lack of statistically significant correlation with model performance rank suggests that visual sequence alignment—used here as a proxy for vision‐based data contamination—does not appear to be a major factor in determining current model performance rankings.

\input{table/mario_data_contamination}
\input{figure_text/mario_data_corr_heatmap}

\input{figure_text/mario_data_contamination_correlation}

\subsection{Text-Level Contamination: Ace Attorney}

\textbf{Setup.} We test whether models reproduce scripted lines from the first two publicly available cases of \textit{Ace Attorney}. Each case is split into an evidence list and a cross-examination script. We then prompt models to generate these sections and compute cosine similarity to the ground truth using Sentence-BERT embeddings.

\textbf{Models tested:} Claude-3.5-Sonnet, Claude-3.7-Sonnet-Thinking, Gemini-2.5-pro-Preview, o4-mini, o3, LLaMA-4-Maverick.

\input{table/ace_data_contain}
\input{figure_text/ace_data}

\textbf{Results.} Table~\ref{tab:ace_attorney_combined} presents similarity scores for 11 models in both cases, alongside their in-game performance and cross-case comparisons. In particular:
\begin{itemize}
    \item Models with higher similarity to the script tend to perform better in the game (e.g., o3, Gemini-2.5-Pro-Preview), suggesting possible memorization effects.
    \item Cross-case similarities (e.g., Gen[First] vs GT[Sister]) are consistently lower, demonstrating that the metric is sensitive to true alignment rather than generic language similarity.
    \item Self-similarity between generated case outputs (Gen[First] vs Gen[Sister]) is relatively high for some models, suggesting stylistic or template reuse.
    \item The Sentence-BERT cosine similarity between the ground-truth scripts for the two cases (1st Turnabout vs Sister Turnabout) is 0.599, which serves as a baseline for evaluating cross-case similarities.
\end{itemize}

To quantify the relationship between textual similarity and performance, we compute linear correlations between total similarity scores and both game score and leaderboard rank. As shown in Figure~\ref{fig:similarity_correlations_combined}, the results hold consistently across both the full model set and the 6-model subset. In all cases, similarity is significantly correlated with better performance, confirming that models may benefit from memorized content.
\input{figure_text/ace_data_linear_before}

\textbf{Mitigation.} To suppress memorized recall and enforce reasoning-based responses, we apply structured prompt interventions. These include: (1) explicitly instructing the model to forget prior knowledge of the game; (2) requiring detailed causal reasoning (cause, evidence, effect) for each action; (3) asking the model to self-evaluate whether its behavior was memory- or reasoning-driven; and (4) modifying the input by replacing all character and item names with neutral tokens (e.g., \textit{"Lawyer 1"}, \textit{"Evidence A"}), and paraphrasing both background context and key contradictions.

\textbf{Results.}
\begin{itemize}
    \item Before mitigation (full model set), total similarity scores strongly correlate with leaderboard rank ($r = -0.773$, $p = 0.005$), indicating that models with higher overlap to the original script tend to perform better.
    \item After applying our prompt-based mitigation (name masking, paraphrasing, and reasoning enforcement), this correlation becomes statistically insignificant ($r = -0.700$, $p = 0.120$), suggesting a reduced reliance on memorization.
    \item Similarly, the correlation between similarity and game score drops to $r = 0.617$ ($p = 0.192$), further supporting the effectiveness of our intervention (see Figure~\ref{fig:similarity_post_mitigation}).
    \item Independent reasoning-based evaluations using o3 as an LLM judge remain predictive of post-mitigation performance, with a strong negative correlation to rank ($r = -0.850$, $p = 0.031$). (see Figure~\ref{fig:evaluator_reasoning_correlation}).

\end{itemize}

\input{figure_text/ace_data_linear_newrank}

\textbf{Conclusion.}  
Initial high performance in the \textit{Ace Attorney} task is partially attributable to memorization of publicly available scripts. Prior to mitigation, script similarity was significantly predictive of performance. After structured prompt-based mitigation—including name masking, paraphrasing, and enforced reasoning—this correlation disappears. Post-intervention rankings instead align with reasoning quality, as verified by an independent evaluator (o3).

\subsection{Extra Text-Level Contamination: Super Mario Bros}
\textbf{Setup.} To assess text-level contamination for \textit{Super Mario Bros.}, we extract a detailed level layout description of World 1-1 from MarioWiki. This description is segmented into a sequence of temporally ordered segments, denoted by $[T1]$ through $[T18]$, representing key points in gameplay. We prompt models to generate layout descriptions aligned with these time points and compare them to the ground truth.

\textbf{Models tested:} o4-mini, o3, Claude-3.7-Sonnet, Claude-3.5-Sonnet, LLaMA-4-Maverick.

\textbf{Result.} We use Sentence-BERT embeddings to calculate pairwise cosine similarity between generated texts and the 1-1 ground truth. Notably, the similarity between the generated descriptions and the true 1-1 layout is consistently \textit{lower} than the similarity between two distinct ground truth descriptions (World 1-1 vs. 1-2). This suggests that models do not rely on memorized textual layout from training corpora and instead generalize loosely or hallucinate. See Table~\ref{tab:smb_text_sim} for quantitative comparisons. A representative side-by-side sample comparison between O3 and ground truth is shown in Figure~\ref{fig:smb_text_comparison}.

\input{table/mario_text_datacontain_table}
\input{figure_text/mario_text_datacontain_figure}

\subsection{Summary}

Our contamination analysis reveals:

\begin{itemize}
    \item \textbf{Vision-level contamination} remains negligible: models reliably reconstruct only the first and last frames, and their alignment coefficients (Kendall’s $\tau$ , RBO) show at best moderate correlation that is not statistically linked to overall performance rank, indicating no reliance on memorized visual sequences.```

    \item \textbf{Text-level contamination} is initially significant in \textit{Ace Attorney}, where models' performance strongly correlates with script similarity. After prompt-based mitigation, this correlation disappears, and performance aligns instead with independently judged reasoning quality.
    
    \item \textbf{LLM-as-Judge (o3)} evaluations confirm that post-mitigation success stems from causal reasoning rather than rote recall. This reinforces the importance of controlled prompting for disentangling memorization from genuine inference.
\end{itemize}

%% file: table/mario_data_contamination.tex
\begin{table}[htbp]
\centering
\caption{Evaluation metrics (Average Pairwise Accuracy, Kendall's $\tau$, and RBO) on the frame ordering task for Super Mario Bros level 1-1. Higher values indicate better reconstruction of the correct temporal sequence from shuffled RGB inputs.}
\begin{tabular}{lccc}
\toprule
\textbf{Model} & \textbf{Accuracy} & \textbf{Kendall's $\tau$} & \textbf{RBO} \\
\midrule
o3           & 0.307 & 0.449 & 0.498 \\
gemini-2.5-pro-preview (thinking)   & 0.300 & 0.458 & 0.489 \\
o4-mini      & 0.233 & 0.324 & 0.463 \\
claude-3-7-sonnet  & 0.147 & 0.044 & 0.422 \\
claude-3-5-sonnet  & 0.180 & 0.099 & 0.418 \\
llama-4-maverick     & 0.100 & 0.019 & 0.324 \\
Perfect      & 1.000 & 1.000 & 0.651 \\
Random       & 0.090 & -0.037 & 0.299 \\
\bottomrule
\end{tabular}
\vspace{0.25em}
\label{tab:vision_contamination}
\end{table}

%% file: figure_text/mario_data_corr_heatmap.tex
\begin{figure}[h]
    \centering
    \includegraphics[width=0.75\linewidth]{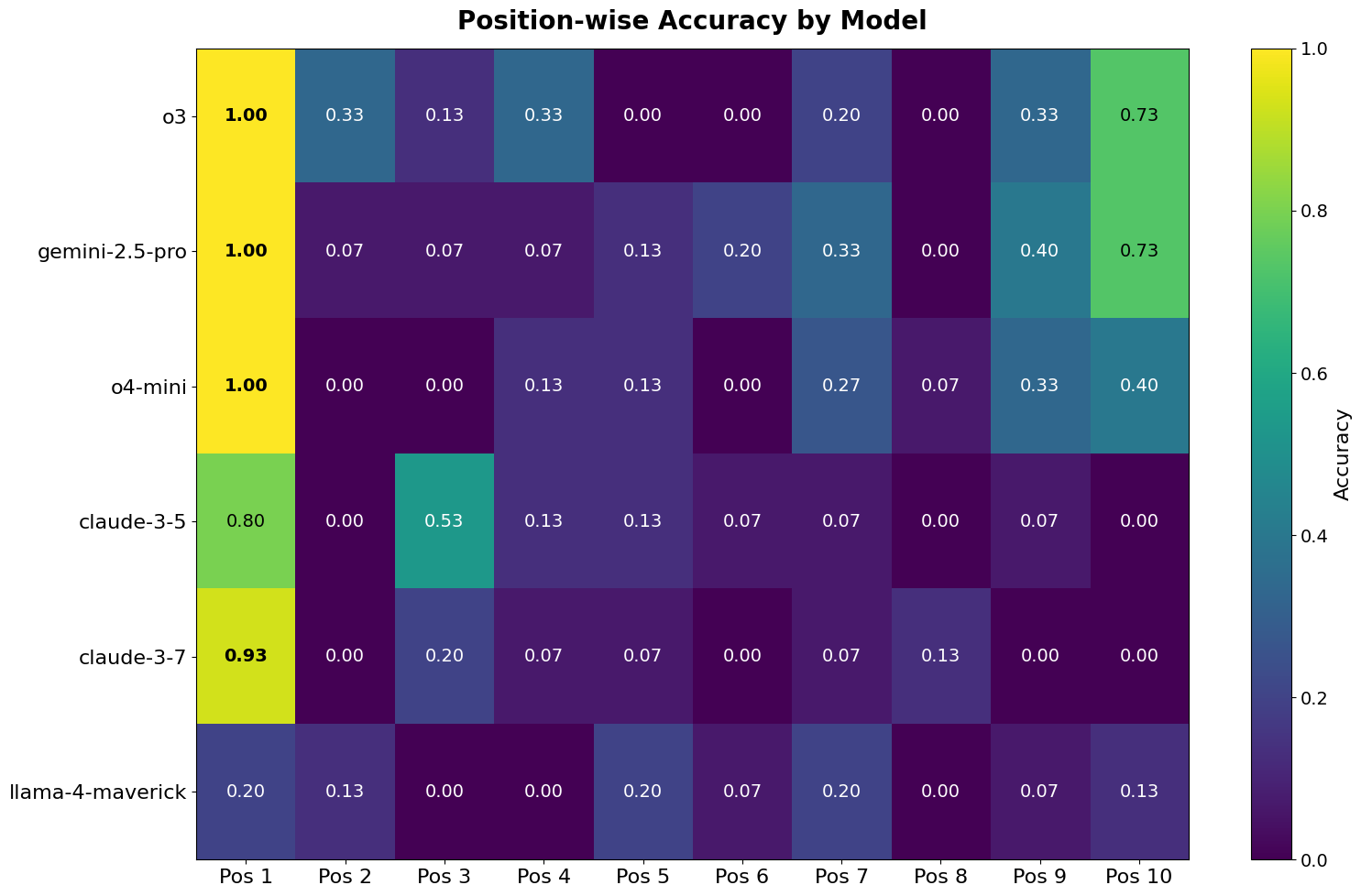}
    \caption{Position‑wise reconstruction accuracy for the shuffled Super Mario Bros level 1‑1 frames. Lighter cells denote higher accuracy; only the first and last positions remain relatively high accuracy}
    \label{fig:position-wise_heatmap}
\end{figure}

\begin{figure}[h]
    \centering
    \includegraphics[width=0.8\linewidth]{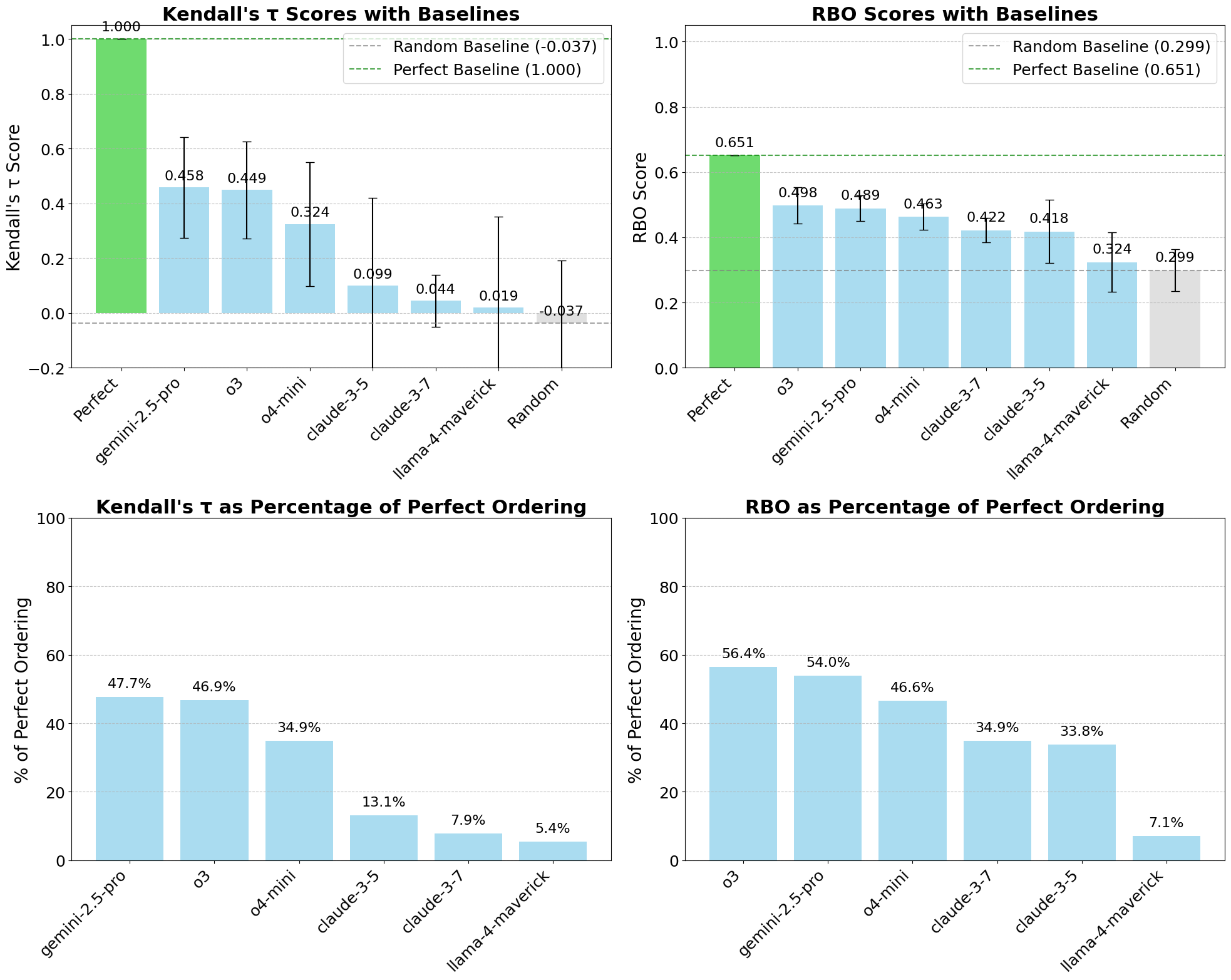}
    \caption{Kendall’s $\tau$ and RBO scores for each model on the frame ordering task. Higher values indicate stronger alignment between predicted and ground-truth frame sequences. Only Gemini-2.5-pro-preview, o3, and o4-mini achieve moderate agreement.}
    \label{fig:Kendall_RBO_Bar_Plots}
\end{figure}

%% file: figure_text/mario_data_contamination_correlation.tex
\begin{figure}[h]
    \centering
    \includegraphics[width=0.9\linewidth]{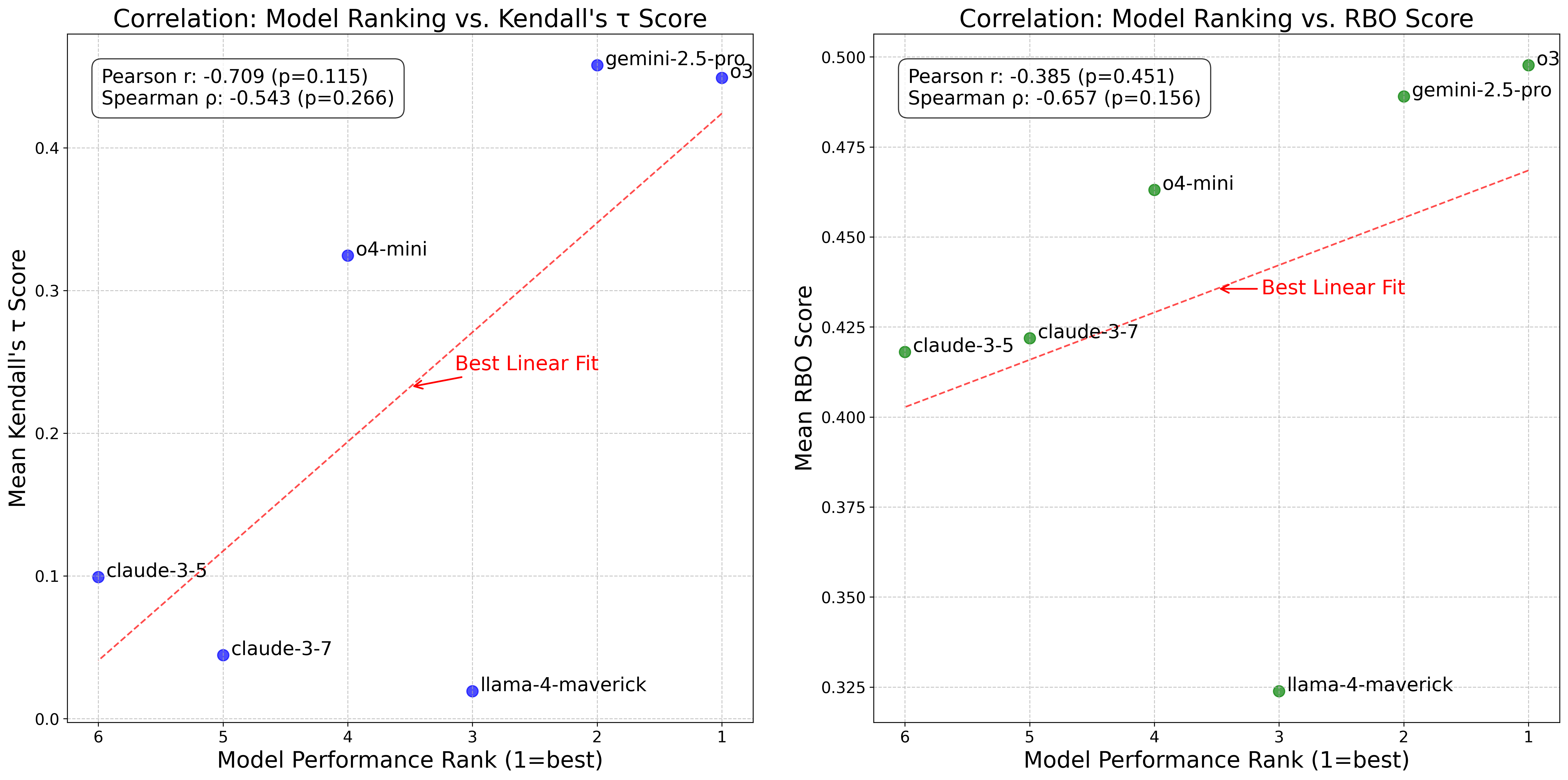}
    \caption{Pearson and spearman correlation between model performance ranks and their alignment scores (Kendall’s $\tau$ and RBO). Although a negative trend is observed, the correlation is not statistically significant.}
    \label{fig:alignment_rank_correlation}
\end{figure}

%% file: table/ace_data_contain.tex
\begin{table}[h]
\centering
\caption{Text-level similarity metrics and performance on \textit{Ace Attorney}. Similarity is measured via Sentence-BERT cosine scores. Cross-case comparisons verify metric reliability.}
\resizebox{\textwidth}{!}{ 
\begin{tabular}{lcccccccc}
\toprule
\textbf{Model} & \textbf{1st Turnabout} & \textbf{Sister Turnabout} & \textbf{Total Sim.} & \textbf{Rank} & \textbf{Game Score} & \textbf{Gen[F] vs GT[S]} & \textbf{Gen[S] vs GT[F]} & \textbf{Gen[F] vs Gen[S]} \\
\midrule
o3-2025-04-16 & 0.863 & 0.904 & 1.767 & 2 & 23 & 0.607 & 0.593 & 0.657 \\
gemini-2.5-pro-preview-05-06 (thinking) & 0.867 & 0.867 & 1.734 & 3 & 20 & 0.561 & 0.544 & 0.532 \\
claude-3-5-sonnet-20241022 & 0.845 & 0.816 & 1.660 & 6 & 6 & 0.609 & 0.538 & 0.672 \\
o1-2024-12-17 & 0.809 & 0.842 & 1.651 & 1 & 26 & 0.634 & 0.589 & 0.675 \\
grok-3-mini-beta & 0.782 & 0.833 & 1.614 & 5 & 7 & 0.646 & 0.550 & 0.707 \\
gpt-4.1-2025-04-14 & 0.755 & 0.812 & 1.567 & 7 & 6 & 0.697 & 0.591 & 0.787 \\
claude-3-7-sonnet-20250219(thinking) & 0.708 & 0.830 & 1.538 & 4 & 8 & 0.684 & 0.535 & 0.588 \\
gemini-2.5-flash-preview-04-17 (thinking) & 0.731 & 0.728 & 1.460 & 8 & 4 & 0.617 & 0.538 & 0.729 \\
gemini-2.0-flash-thinking-exp-1219 & 0.659 & 0.780 & 1.438 & 9 & 4 & 0.719 & 0.562 & 0.810 \\
LLaMA-4 Maverick & 0.491 & 0.734 & 1.224 & 13 & 0 & 0.594 & 0.535 & 0.772 \\
O4 Mini & 0.586 & 0.625 & 1.212 & 11 & 1 & 0.574 & 0.531 & 0.764 \\
\bottomrule
\end{tabular}
}
\vspace{5pt}
\label{tab:ace_attorney_combined}
\end{table}

%% file: figure_text/ace_data.tex
\begin{figure}[h]
    \centering
    \includegraphics[width=0.6\textwidth]{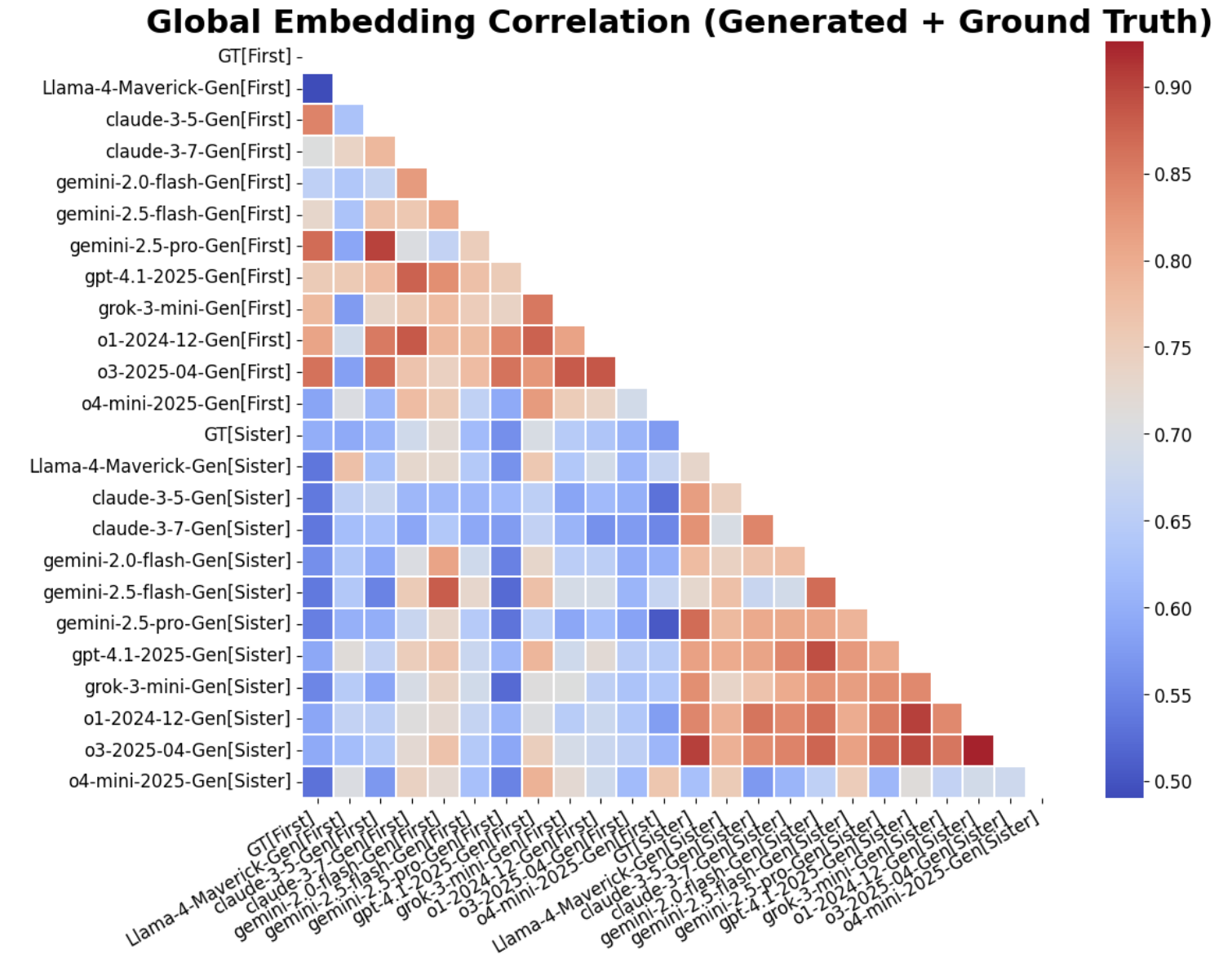}
    \captionof{figure}{Global Embedding Correlation (Generated + Ground Truth) between model-generated texts across "First" and "Sister" turnabouts.}
    \label{fig:embedding_correlation}
\end{figure}

%% file: figure_text/ace_data_linear_before.tex
\begin{figure}[h]
    \centering
    \begin{subfigure}[t]{0.49\linewidth}
        \includegraphics[width=\linewidth]{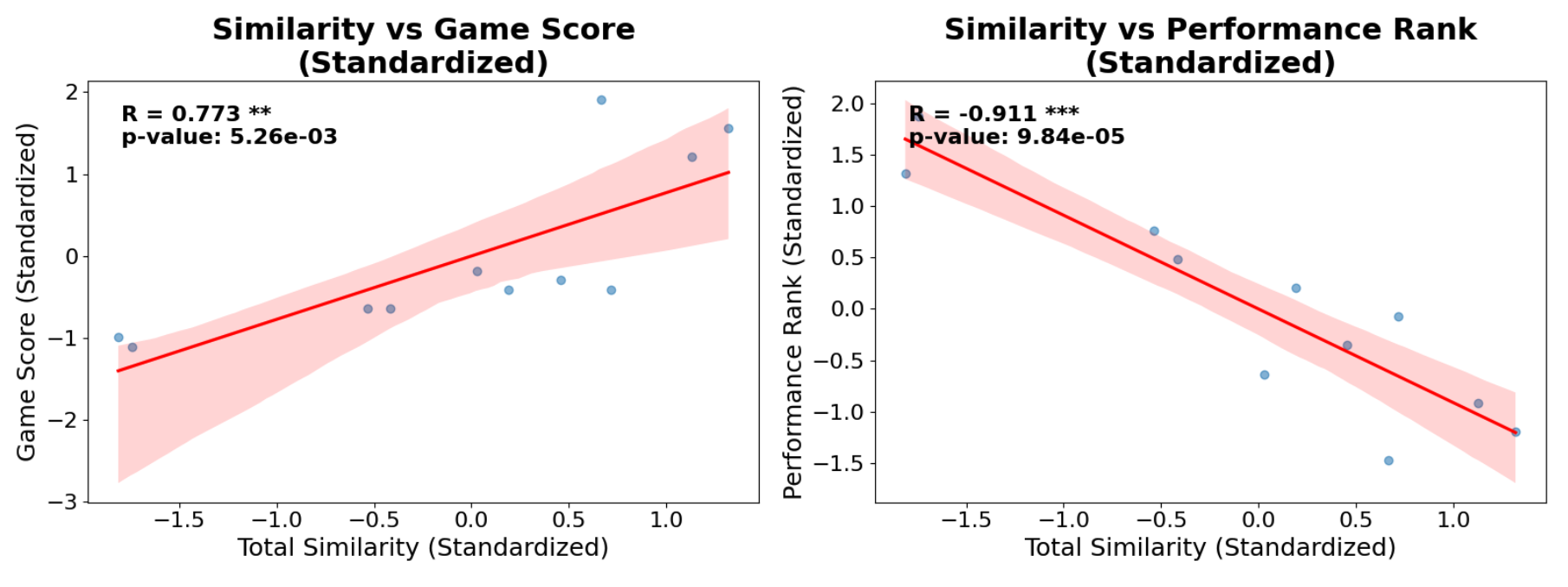}
        \caption{Full model set. Left: similarity vs. game score ($r = 0.773$, $p = 0.005$). Right: similarity vs. leaderboard rank ($r = -0.911$, $p = 9.840 \times 10^{-5}$).}
        \label{fig:similarity_full}
    \end{subfigure}
    \hfill
    \begin{subfigure}[t]{0.49\linewidth}
        \includegraphics[width=\linewidth]{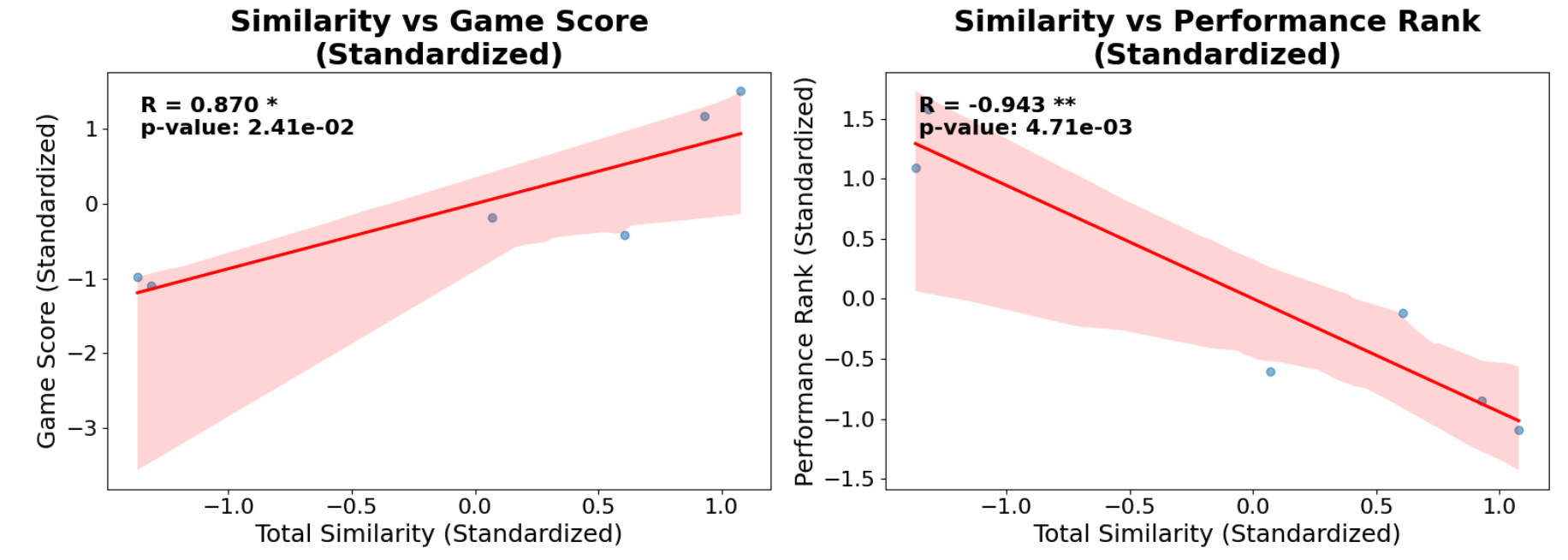}
        \caption{6-model subset. Left: similarity vs. game score ($r = 0.870$, $p = 0.024$). Right: similarity vs. leaderboard rank ($r = -0.943$, $p = 0.005$).}
        \label{fig:similarity_subset}
    \end{subfigure}
    \caption{Linear correlations between script similarity and model performance across the full model set (left) and the 6-model subset used for cross-modality comparison (right). Shaded areas indicate 95\% confidence intervals. In both settings, higher similarity is significantly associated with higher game scores and  higher leaderboard position (i.e., lower rank number)}
    \label{fig:similarity_correlations_combined}
    \vspace{-20pt}
\end{figure}

%% file: figure_text/ace_data_linear_newrank.tex
\begin{figure}[h]
    \centering
    \vspace{-10pt}
    \begin{subfigure}[t]{0.49\linewidth}
        \includegraphics[width=\linewidth]{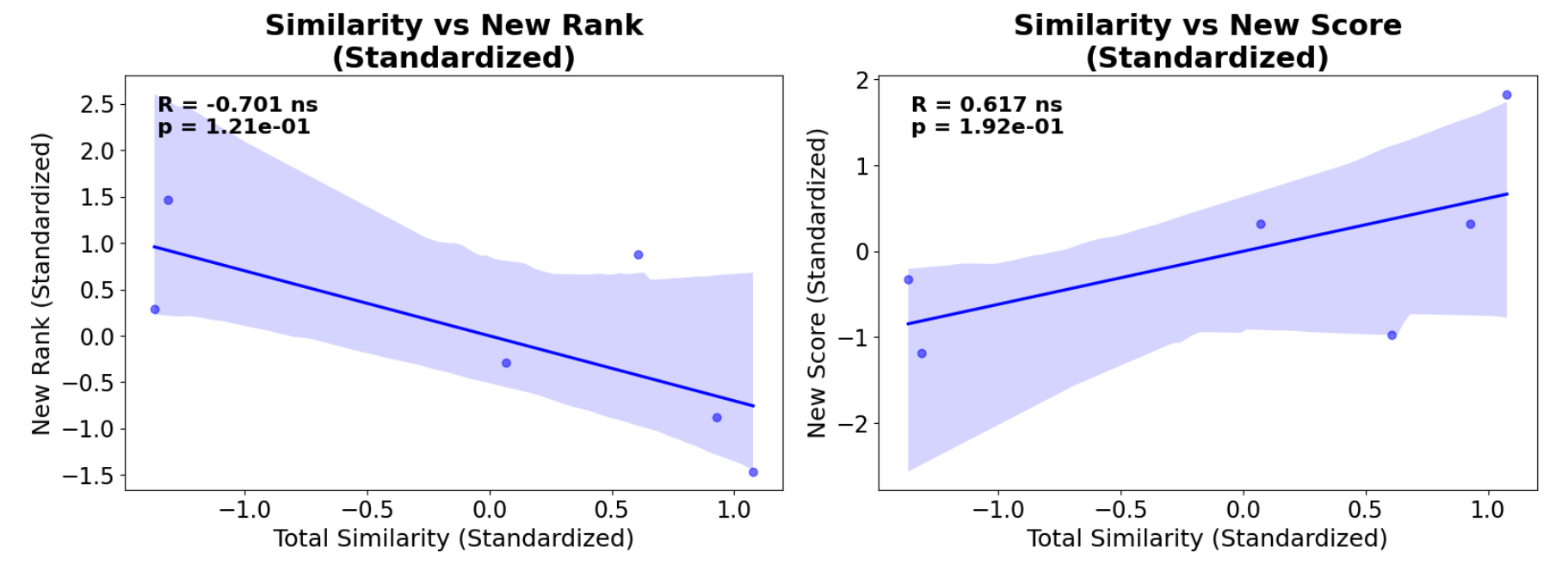}
        \caption{Similarity vs. new rank (left) and new score (right). No statistically significant correlations remain after mitigation, suggesting similarity no longer explains model success.}
        \label{fig:similarity_post_mitigation}
    \end{subfigure}
    \hfill
    \begin{subfigure}[t]{0.49\linewidth}
        \includegraphics[width=\linewidth]{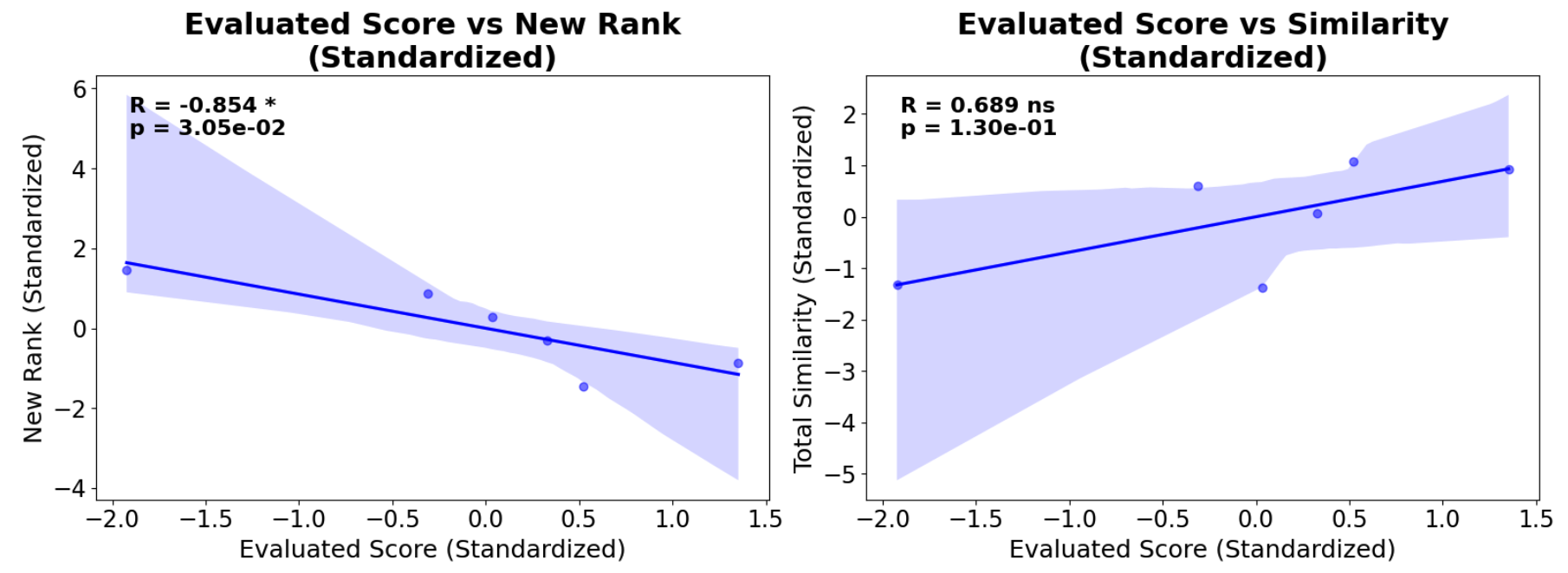} 
        \caption{Evaluator score vs. new rank (left) and similarity (right). Evaluated reasoning correlates strongly with new rank ($r = -0.854$, $p = 0.031$) but not with similarity, suggesting post-mitigation rankings reflect reasoning.}
        \label{fig:evaluator_reasoning_correlation}
    \end{subfigure}
    \caption{Post-mitigation analysis. Left: model similarity no longer predicts rank or score. Right: LLM-as-Judge (o3) evaluations suggest model ranking is now aligned with reasoning quality, not memorized content.}
    \label{fig:post_mitigation_combined}
    \vspace{-10pt}
\end{figure}

%% file: table/mario_text_datacontain_table.tex
\begin{table}[h]
    \centering
    \caption{Text-level similarity metrics for \textit{Super Mario Bros.} 1-1 layout descriptions. Similarity is measured using Sentence-BERT cosine scores. The last column shows the ground truth similarity between level 1-1 and 1-2 as a baseline.}
    \label{tab:smb_text_sim}
    \resizebox{\textwidth}{!}{
    \begin{tabular}{lcccccc}
        \toprule
        \textbf{Model} & \textbf{Gen[1-1] vs GT[1-1]} & \textbf{Gen[1-2] vs GT[1-2]} & \textbf{Gen[1-1] vs Gen[1-2]} & \textbf{Gen[1-1] vs GT[1-2]} & \textbf{Gen[1-2] vs GT[1-1]} & \textbf{GT[1-1] vs GT[1-2]} \\
        \midrule
        o3 (2025-04-16) & 0.753076 & 0.813358 & 0.882990 & 0.749037 & 0.800998 & 0.844817 \\
        o4-Mini (2025-04-16) & 0.708403 & 0.725378 & 0.852628 & 0.707597 & 0.799113 & 0.844817 \\
        Claude-3.7 Sonnet (2025-02-19) & 0.679185 & 0.627961 & 0.821135 & 0.560105 & 0.750460 & 0.844817 \\
        Claude-3.5 Sonnet (2024-10-22) & 0.686953 & 0.604426 & 0.795933 & 0.587022 & 0.701139 & 0.844817 \\
        LLaMA-4 Maverick FP8 & 0.643627 & 0.590322 & 0.846092 & 0.542652 & 0.651531 & 0.844817 \\
        
        \bottomrule
    \end{tabular}
    }
\end{table}

%% file: figure_text/mario_text_datacontain_figure.tex
\begin{figure}[h]
  \centering
  \begin{tcolorbox}[
    colback=gray!10,
    colframe=black,
    title={\bfseries Representative side‐by‐side sample comparison (T3–T8)},
    fonttitle=\bfseries,
    sharp corners=south,
    boxrule=0.5pt
  ]
    \scriptsize
    \begin{minipage}{0.475\textwidth}
      \raggedright
      \textbf{Ground Truth}\\[0.3em]
      \textnormal{[T3]} They then reach the first “?” Block of the game.\\
      \textnormal{[T4]} They encounter the six‑block triangle formation of “?” Blocks and regular bricks, with a Magic Mushroom inside the left block.\\
      \textnormal{[T5]} A set of three Pipes can be found, with Little Goombas located in between them.\\
      \textnormal{[T6]} The final pipe among the set leads to a bonus area with 19 coins and can be entered to bypass much of the level, exiting at the fifth pipe near the end.\\
      \textnormal{[T7]} If the pipe is not taken, a hidden 1‑Up Mushroom block appears just after the pipes.\\
      \textnormal{[T8]} A pit follows, then another “?” Block containing either a Magic Mushroom or Fire Flower, depending on the player’s form.
    \end{minipage}%
    \hfill
    \begin{minipage}{0.475\textwidth}
      \raggedright
      \textbf{o3 Generated}\\[0.3em]
      \textnormal{[T3]} Striking the Question Block yields a Super Mushroom that slides right along the ground past the hill.\\
      \textnormal{[T4]} \textcolor{red}{Mario next encounters a pair of spaced green pipes emerging from the ground, each occasionally spawning a snapping Piranha Plant.}\\
      \textnormal{[T5]} Between those pipes, two more Little Goombas approach in tandem over flat terrain.\\
      \textnormal{[T6]} \textcolor{red}{Three floating Question Blocks form a row above bricks, with the middle hiding a helpful Fire Flower.}\\
      \textnormal{[T7]} \textcolor{red}{A lone Koopa Troopa patrols beneath these blocks.}\\
      \textnormal{[T8]} Shortly past this, a tall staircase of ground bricks leads up then down, concealing a hidden 1‑Up Block atop the first step.
    \end{minipage}
  \end{tcolorbox}
  \caption{A representative side‐by‐side sample comparison between o3 and ground truth.}
  \label{fig:smb_text_comparison}
\end{figure}

%% file: text/7_appendix_3_prompts_optimization.tex
\section{Prompt Optimization}
\label{appendix:prompt_optimization}

In this section, we present a case study illustrating our two-stage prompt optimization approach. We design two empirically optimized baseline prompts developed by equally adequate computer science graduate students based on Figure~\ref{fig:game_and_overview}. Subsequently, we employ DSPy to bootstrap an optimized prompt for each baseline, selecting from a diverse set of five optimizer models and retaining only the best-performing prompt. We show performance variance across prompts optimized through bootstrapping is lower than that of baseline prompts. 
\subsection{Empirically Optimized Baseline Prompts}


\tcbset{
  reasoningstyle/.style={
     colback=gray!5,
     colframe=purple,          
     coltitle=green,             
     fonttitle=\bfseries,      
     boxrule=1.2pt,   
     arc=4pt,              
     left=1em,right=1em,top=1em,bottom=1em,
  }
}

\newtcolorbox{vision-prompt-1}{
  reasoningstyle,
  title={Game 2048 — Empirically Optimized Prompt Template 1}
}

\newtcolorbox{vision-prompt-2}{
  reasoningstyle,
  title={Game 2048 — Empirically Optimized Prompt Template 2}
}

\newtcolorbox{vision-prompt-dspy1}{
  reasoningstyle,
  title={Game 2048 — DSPy Optimized Prompt Template 1}
}

\newtcolorbox{vision-prompt-dspy2}{
  reasoningstyle,
  title={Game 2048 — DSPy Optimized Prompt Template 2}
}

\begin{vision-prompt-1}

\textbf{\textcolor{blue}{system\_prompt:}}\\
You are an intelligent AI player playing the 2048 game. Your goal is to make strategic moves to combine tiles and reach the highest possible tile value.\\

IMPORTANT: You MUST format your response using EXACTLY these lines:\\
thought: [Your reasoning about the game state]\\
move: [move]\\
Where [move] must be one of: "up", "down", "left", or "right".\\
Do not include \# or any other prefix. Start directly with "thought:" followed by your analysis.\\

\textbf{\textcolor{blue}{user\_prompt:}}\\
2048 Game Quick Guide:\\
Primary Goal: Combine like tiles to create tiles with higher values.\\
Ultimate Goal: Create a tile with the value 2048 or higher.\\
Game Mechanics:\\
- The game is played on a 4x4 grid.\\
- Each move (up, down, left, right) shifts all tiles in that direction.\\
- Tiles with the same value that collide during a move combine into a single tile with twice the value.\\
- After each move, a new tile (2 or 4) appears in a random empty cell.\\
- The game ends when there are no valid moves left.\\
Action Space:\\
You must select one of these 4 moves:\\
- up: Shift all tiles upward\\
- down: Shift all tiles downward\\
- left: Shift all tiles to the left\\
- right: Shift all tiles to the right\\
Key Strategies:\\
1. Build a stable structure - Keep your highest value tiles in a corner.\\
2. Maintain a clear path - Always have a direction where you can combine tiles.\\
3. Chain reactions - Set up sequences of merges that can happen in a single move.\\
4. Look ahead - Think about the consequences of your moves 2-3 steps ahead.\\
5. Building patterns - Common patterns include: (1) Snake/Zig-zag pattern: Arrange tiles in decreasing order in a zigzag; (2) Corner anchoring: Keep the highest tile in a corner and build around it.\\
Avoid:\\
- Getting high-value tiles stuck in the middle of the board\\
- Creating scattered small values that block potential merges\\
- Making moves that could lead to grid lock\\
Previous Game History:\\
\textcolor{blue}{\{Previous Game History\}}\\
Please analyze the 2048 board and determine the best move.\\
\textcolor{blue}{\{Symbolic Board Features\}}\\
Key considerations:\\
- Look for opportunities to merge similar tiles\\
- Maintain your highest tiles in a corner\\
- Keep space for new tiles to appear\\
- Avoid trapping high-value tiles in the middle\\
IMPORTANT - FORMAT YOUR RESPONSE EXACTLY LIKE THIS:\\
thought: [your analysis here]\\
move: [move]\\
Where [move] must be one of: "up", "down", "left", or "right".\\
Do NOT use \# or any other prefix. Start directly with "thought:" followed by your analysis.

\end{vision-prompt-1}

\begin{vision-prompt-2}

\textbf{\textcolor{blue}{system\_prompt:}}\\
You are an AI agent specialized in 2048 gameplay, your purpose is to analyze board states and suggest optimal moves that maximize your scores.\\

\#\# Your Available Actions\\
For each turn, you must select one command:\\
- up: Shifts the entire grid upward\\
- down: Shifts the entire grid downward\\
- left: Shifts the entire grid leftward\\
- right: Shifts the entire grid rightward\\

When you choose a direction (up, down, left, or right), all tiles shift accordingly. Matching tiles that collide during this shift combine into a single tile representing their sum. After every move, a new tile with a value of either 2 or 4 appears in a random empty cell. The game concludes when no legal moves remain.\\

\textbf{\textcolor{blue}{user\_prompt:}}\\
\#\# 2048 Gameplay Strategies

\#\#\# Principles
The most successful 2048 strategies typically involve:\\
1. Establish your highest-value tile in one corner and build a descending value structure around it.\\
2. Maintain consistent movement patterns that preserve your high-value corner configuration while allowing for regular merges.\\
3. Anticipate how each potential move affects not just the immediate board state but your options 2-3 moves ahead.\\
4. Create opportunities for chain reactions where multiple merges can occur in a single directional move.\\
5. Implement proven arrangements such as:\\
   - Decreasing value snakes that zigzag across the board.\\
   - Corner-anchored structures with decreasing values along the edges.\\
   
\#\#\# Pitfalls to Avoid\\
Certain decisions consistently lead to board deterioration:\\
- Allowing high-value tiles to become isolated in central positions.\\
- Creating scattered low-value tiles that impede potential combinations.\\
- Making moves that reduce overall board fluidity and movement options.\\

\#\# Current Game Context\\
\textcolor{blue}{\{Previous Game History\}}\\

\#\# Board Analysis\\
\textcolor{blue}{\{Symbolic Board Features\}}\\

\#\# Response Protocol\\
**YOUR ANALYSIS MUST STRICTLY ADHERE TO THIS FORMAT:**\\
thought: [Provide your detailed reasoning about the current board state, potential moves, and strategic implications]\\
move: [move]\\

Your move selection must be one of these exact terms: "up", "down", "left", or "right".\\

Begin your response directly with "thought:" followed by your strategic analysis. Do not include any prefixes, headers, or additional formatting.

\end{vision-prompt-2}

\subsection{DSPy Optimized Prompts and Comparison}

\begin{algorithm}[h]
\caption{Standardizing Gaming Prompt Optimization with \textsc{SIMBA} from \textsc{DSPy}}
\KwIn{%
  Training environments $\mathcal{E}_{\mathrm{train}}$, %
  development environments $\mathcal{E}_{\mathrm{dev}}$, %
  \emph{target} LM set $\mathcal{M}_t$ for performance evaluation,\\
  \emph{optimizer} LM set $\mathcal{M}_o=\{\texttt{o3},\texttt{gemini‑2.5‑pro},\texttt{claude‑3.7‑think},\texttt{deepseek‑R1},\texttt{grok3‑mini}\}$, %
  maximum optimisation steps $k$}
\KwOut{Best prompt module $\mathcal{P}^\star$ (highest mean dev score over all $M_t$)}
\BlankLine

$\mathcal{P}\,\leftarrow\,\texttt{ChainOfThought}(\text{``state}\!\rightarrow\!\text{action''})$
$s_{\mathrm{best}}\leftarrow -\infty$,  $\mathcal{P}^\star\leftarrow\mathcal{P}$\;

\ForEach{$M_o \in \mathcal{M}_o$}{%
  \texttt{dspy.configure(lm=$M_o$)}\;
  
  \textbf{// joint optimisation across \emph{all} target LMs}\\
  $\mathcal{O}\leftarrow\texttt{SIMBA}(\{M_t\},k)$\;
  $\widehat{\mathcal{P}}\leftarrow\mathcal{O}.\texttt{compile}(\mathcal{P},\mathcal{E}_{\mathrm{train}})$\;
  
  \BlankLine
  \textbf{// evaluate average dev score over every $M_t$}\\
  $s_{\mathrm{avg}}\leftarrow 0$\;
  \ForEach{$M_t \in \{M_t\}$}{%
    \texttt{dspy.configure(lm=$M_t$)}\;
    $s_{\mathrm{avg}}\mathrel{+}= \texttt{Evaluate}(\widehat{\mathcal{P}},\mathcal{E}_{\mathrm{dev}})$\;
  }
  $s_{\mathrm{avg}}\leftarrow s_{\mathrm{avg}} / |\{M_t\}|$\;
  
  \If{$s_{\mathrm{avg}} > s_{\mathrm{best}}$}{%
    $s_{\mathrm{best}}\leftarrow s_{\mathrm{avg}}$\;
    $\mathcal{P}^\star\leftarrow\widehat{\mathcal{P}}$\;
  }
}

\Return $\mathcal{P}^\star$
\label{alg:dspy}
\end{algorithm}

\begin{vision-prompt-dspy1}
\textbf{\textcolor{blue}{system\_prompt:}}\\

You are an AI agent specifically designed to play the game 2048. Your primary objective is to make strategic moves that effectively merge tiles to achieve the highest possible tile value.\\

\textbf{\textcolor{blue}{user\_prompt:}}\\
\#\# Game Overview\\
The game 2048 involves combining identical number tiles on a grid to create tiles with progressively higher values.\\

\#\# Game Mechanics\\
- The game is played on a **4×4 grid**\\
- Each move (up, down, left, right) shifts all tiles in the chosen direction\\
- When two identical tiles collide during a move, they merge into a single tile with twice the value\\
- After each move, a new tile (either 2 or 4) appears randomly in an empty cell\\
- The game concludes when no legal moves remain available\\

\#\# Action Space\\
- **up**: Shifts all tiles toward the top of the grid\\
- **down**: Shifts all tiles toward the bottom of the grid\\
- **left**: Shifts all tiles toward the left side of the grid\\
- **right**: Shifts all tiles toward the right side of the grid\\

\#\# Strategic Principles\\
1. **Corner Anchoring**: Position your highest-value tile in a corner and build around it\\
2. **Structural Stability**: Arrange surrounding tiles in descending order to create a stable formation\\
3. **Maintaining Merge Paths**: Always keep at least one direction available for safe combinations\\
4. **Creating Chain Reactions**: Set up moves that trigger multiple merges in a single action\\
5. **Forward Planning**: Think 2-3 moves ahead to avoid grid-lock and maintain empty spaces\\

\#\# Pitfalls to Avoid\\
- Allowing high-value tiles to drift into central positions\\
- Scattering small-value tiles that obstruct potential merges\\
- Making moves that leave the board with no follow-up merge opportunities\\

\#\# Context Variables\\
\#\#\# Previous Game History\\
\textcolor{blue}{\{Previous Game History\}}\\

\#\#\# Board Features\\
\textcolor{blue}{\{Previous Game History\}}\\

\#\# Response Format\\
Your response must follow this exact two-line format:\\
\begin{verbatim}
```
thought: [your brief analysis of the current board state]\\
move: [up|down|left|right]\\
```
\end{verbatim}
**Important**: Include nothing else beyond these two lines. No additional text, prefixes, symbols, or explanations.\\

\end{vision-prompt-dspy1}

\begin{vision-prompt-dspy2}
\textbf{\textcolor{blue}{system\_prompt:}}

You are an **AI agent** playing **2048**. Your objective is to select moves that merge tiles efficiently and achieve the highest possible tile value.\\

\textbf{\textcolor{blue}{user\_prompt:}}\\
\# 2048 Gaming Guide\\
\#\# Primary Goal\\
Combine like tiles to reach **2048** or higher.\\

---\\

\#\# Game Mechanics\\
- Played on a **4×4 grid**; each move (“up”, “down”, “left”, “right”) shifts every tile.\\
- Identical tiles that collide merge into one tile with **double the value**.\\
- After each move, a new tile (**2** or **4**) appears randomly in an empty cell.\\
- The game ends when **no legal moves** remain.\\

---\\

\#\# Action Space\\
- **up**: Shift all tiles upward.\\
- **down**: Shift all tiles downward.\\
- **left**: Shift all tiles to the left.\\
- **right**: Shift all tiles to the right.\\

---\\

\#\# Key Strategies\\
1. **Corner anchor** - Park your highest tile in one corner and build around it.\\
2. **Stable structure** - Arrange surrounding tiles in descending order to protect the corner.\\
3. **Clear merging path** - Keep at least one direction available for safe combinations.\\
4. **Chain reactions** - Set up moves that trigger multiple merges in one swipe.\\
5. **Look ahead** - Plan 2–3 moves in advance to avoid grid-lock. Preserve empty spaces for new tiles.\\

---\\

\#\# Avoid\\
- Letting high-value tiles drift into the center.\\
- Scattering small tiles that block merges.\\
- Moves that leave the board with no follow-up merges.\\

---\\

\#\# Previous Game History\\
\textcolor{blue}{\{Previous Game History\}}\\

\#\# Board Features\\
\textcolor{blue}{\{Symbolic Board Features\}}\\

**Response format (use exactly two lines):**\\
thought: your brief analysis of the current board\\
move: up | down | left | right\\

Include nothing else—no prefixes, symbols, or extra text.

\end{vision-prompt-dspy2}

\subsection{Performance Comparison}
In \sysnamenospace, we follow the SIMBA optimizer implementation in DSPy with performance metrics defined in Section~\ref{sec:game_metrics} to optimize the prompt with five optimizer models: Claude-3-7-sonnet, Gemini-2.5-Pro-Preview, o3, Deepseek-r1, Grok-3-Mini-Beta, to search the best performing prompt yielding highest average reward cross all target models in 20 optimization steps ($k=20$).

\input{table/prompt_optimization}

Among three target models: Gemini-2.5-Flash-Preview, Claude-3-5-Sonnet, o4-mini, experiments results from \sysname show evidence that our prompt optimization pipeline can reduce performance discrepancy between two candidate prompts by 33.8\% to 63.5\% on the three models across 3 runs. Details are reported in Table~\ref{tab:prompt_optimization}.

%% file: table/prompt_optimization.tex
\begin{table}[h]
    \centering
    \scriptsize
    \caption{Model performance across various prompt types in Game 2048 with harness, where $\Delta_e$ and $\Delta_p$ stand for performance difference between empirically deisgn prompt pairs and DSPy optimzied prompt pairs. P1 and P2 denotes to different prompt templates.}
     \resizebox{\linewidth}{!}{%
    \begin{tabular}{l|ccc|ccc}
        \toprule
        \textbf{Model} & \textbf{Empirical P1} & \textbf{Empirical P2} & $ {\lvert}\Delta_e {\rvert} \left(\downarrow \right)$ & \textbf{DSPy P1} & \textbf{DSPy P2} &${\rvert}\Delta_p {\lvert} \left(\downarrow \right)$\\
        \midrule
        gemini-2.5-flash-preview-04-17       &  1697.3{\scriptsize$\pm$548}                   &        1478.7{\scriptsize$\pm$440}   &  218.6 &      1746.0{\scriptsize$\pm$518}   &         1601.3{\scriptsize$\pm$174}       &   \textbf{144.7}  \\
        claude-3-5-sonnet-20241022        &            2624.0{\scriptsize$\pm$466}             &            2235.3{\scriptsize$\pm$862}   &  388.7  &      2786.0{\scriptsize$\pm$290}          &    2928.0{\scriptsize$\pm$318}   &       \textbf{142.0}       \\
        o4-mini-2025-04-16        &           4432.0{\scriptsize$\pm$1096}              &               3680.0{\scriptsize$\pm$963}   &     752.0
        &      3851.3{\scriptsize$\pm$864.4}    &   4320.0{\scriptsize$\pm$700}  &       \textbf{468.7}         \\
        \bottomrule
    \end{tabular}%
    }
    \vspace{0.5em}
    \label{tab:prompt_optimization}
\end{table}


%% file: text/7_appendix_4_additional_correlation_analysis.tex
\section{Additional Correlation Study}

\subsection{Benchmark List for Correlation Study}
\label{appendix:correlation}

We use 20 publicly available benchmarks spanning seven capability categories, including factual knowledge, physics, mathematics, coding, visual reasoning, language understanding, and puzzle solving. These benchmarks are chosen to provide a comprehensive view of general-purpose model abilities and to support the correlation and decomposition analyses.

Table~\ref{tab:benchmark_ranks_only} summarizes the per-model rankings across all benchmarks, grouped by category. The rankings are used to compute Spearman correlations and to uncover latent capability axes through low-rank decomposition.

\input{table/benchmarks}

\subsection{Correlation and Latent Feature Analysis with Super Mario Bros.}
\label{appendix:Mario_inlclude}

To better understand the impact of high-variance games, we conducted a supplementary analysis that includes \textit{Super Mario Bros. (SMB)} in both the Spearman correlation matrix and latent ability decomposition.

\textbf{Spearman Correlation with SMB.}
Figure~\ref{fig:corr_mario} shows the extended Spearman correlation matrix including SMB. Although most benchmarks retain positive mutual correlations, SMB displays a notably weaker with language-heavy and code-generation benchmarks, where correlations drop to near zero or slightly negative values. This suggests that SMB performance is less stable between models and less aligned with other benchmarked capabilities, likely due to high variance or vision-specific difficulties. These observations support our decision to exclude SMB from the main correlation and decomposition analyses in Section~\ref{sec:exp_analysis}.

\textbf{Latent Ability Decomposition with SMB.}
We also repeated the low-rank factorization analysis with SMB included. Figure~\ref{fig:low_rank_mario} shows the contribution of each benchmark (including SMB) to the discovered latent features. SMB contributes moderately across most features, particularly in Feature 1 (long-context language reasoning) and Feature 3 (puzzle solving \& coding capabilities). This aligns with SMB’s demand for multimodal reasoning—visual perception, spatial planning, and action timing. However, its weights are less concentrated, likely due to variance in model rankings across runs.

\input{figure_text/corr_feature_mario}

\subsection{Visualizing Benchmark Relationships.} 

To better understand how \sysname{} compares with established benchmarks, we visualize benchmark similarity using t-SNE. We embed each benchmark and game as a high-dimensional vector based on either model performance scores (ranging from 0–100) or model rankings, and project them into 2D using t-SNE. Benchmarks with NaN values (e.g., missing model scores) were excluded to ensure reliable embeddings.

We show two versions:

- The first version (Figure~\ref{fig:tsne_wo_smb}) excludes Super Mario Bros. (SMB), reflecting the setup used in the main paper.

- The second version (Figure~\ref{fig:tsne_w_smb}) includes SMB to explore its positioning relative to other benchmarks.

In the score-based plots, \textit{Ace Attorney}, \textit{Sokoban}, and \textit{Tetris} cluster closely with reasoning-heavy benchmarks like EnigmaEval and HLE, reflecting shared demands in long-horizon planning and symbolic reasoning. With SMB included, it appears adjacent to 2048 and ChatbotArena-Vision, consistent with its reliance on visual perception and spatial coordination.

In the rank-based projections, \textit{Candy Crush} and \textit{2048} are near Math 500 and AIME 2025, while \textit{Sokoban} remains isolated, likely due to its unforgiving action space. SMB's placement shifts closer to vision benchmarks, but does not cluster tightly, likely due to its high inter-model variance..

Putting together, the two tSNE graphs support the conclusion that different \sysname{} games probe distinct capabilities and align with well-established benchmark clusters in meaningful ways.
\input{figure_text/tsne}

\subsection{Fitting Polynomial Models Technical Details.}
\label{appendix:polynomial_modeling}

\input{table/polynomial_model_linear_ablation}

Let $m$ be the number of models to be ranked, $n$ be the total number of existing benchmark categories, $R_{i,j} \in \mathcal{Z}_{\geq 1}$ be an ordinal rank of model $i$ on a benchmark from category $j$, $G_{i, g} \in \mathcal{Z}_{\geq 1}$ be the rank for model $i$ on game $g$. Correspondingly, we define $\mathbf R_j$ and $\mathbf G_g$ be the rank vector for category $j$ and a game $g$ across all models. Since we can collect ranking data from all models across combinations of benchmarks in different categories, for each benchmark combination, we can then define a polynomial feature map for polynomial expansion as shown in Eq.~\ref{eq:feature-map}. 

\begin{equation}
\phi : \mathbb{R}^n \to \mathbb{R}^{p(d)}, \text{ where }
p(d)=\sum_{k=0}^{d}\binom{n+k-1}{k},\text{ denote }
\Phi=
\begin{pmatrix}
\phi(\mathbf{L}_1)^\top\\
\vdots\\
\phi(\mathbf{L}_m)^\top
\end{pmatrix}
\in\mathbb{R}^{m\times p}
\label{eq:feature-map}
\end{equation}

As a result, let $\mathbf w \in \mathcal R^p$ be the parameters to be learned, we can predict the gaming ranking of a model $i$ given its ranking on a set of benchmarks from all categories of interest. $\hat G_{i,g} = \mathbf w \phi \left( \mathbf L_i\right), \hat {\mathbf G_g} = \Phi \mathbf w$. With the non-negative least-square-fit objective, $\displaystyle \min_{\mathbf w}\; \bigl\lVert \Phi \mathbf w - \mathbf g \bigr\rVert_{2}^2$ s.t. $w_k \geq 0$ for all linear terms $k$, where the closed-form solution can be expressed as $\mathbf w^*=\left(\Phi^{\top} \Phi\right)^{-1} \Phi^{\top} \mathbf g$. Non‑negativity ensures every feature contributes additively, otherwise being worse in category$j$ makes gaming performance better is counter-intuitive. each term in $\mathbf w$ quantifies how much and in what direction the polynomial term of the $n$ categories drives the game ranking. We show linear and quadratic models trained on benchmarks presented in Appx.~\ref{appendix:correlation}. Comparisons of polynomial fitting using linear models and quadratic models of different categorical combinations are presented in Table~\ref{tab:polynomial_model}.

%% file: table/benchmarks.tex
\label{appendix:benchmark_ranks_only}

\begin{table}[h]
\centering
\small
\caption{
Model rankings (1 = best) across 20 benchmarks, grouped by capability category. Abbreviated model names: C3.5 = claude-3.5-Sonnet-20241022, C3.7 = claude-3.7-Sonnet-20250219-thinking, G4O = gpt-4o-2024-11-20, O1 = o1-2024-12-17, O3 = o3-2025-04-16, Gem = gemini-2.5-pro-preview-05-06(thinking), L4 = llama-4-maverick-17b-128e-instruct-fp8, O4m = o4-mini.}
\begin{tabular}{llcccccccc}
\toprule
\textbf{Category} & \textbf{Benchmark} & C3.5 & C3.7 & G4O & O1 & O3 & Gem & L4 & O4m \\
\midrule
\multirow{4}{*}{Factual}
  & MMLU-Pro       & 7 & 4 & 2 & 6 & 8 & 3 & 1 & 5 \\
  & GPQA           & 7 & 3 & 2 & 6 & 8 & 5 & 1 & 4 \\
  & HLE            & 7 & 4 & 3 & 6 & 8 & 5 & 1 & 2 \\
  & HLE (Text)     & 7 & 4 & 3 & 6 & 8 & 5 & 1 & 2 \\
\midrule
\multirow{2}{*}{Physics}
  & EMMA-Physics   & 7 & 4 & 1 & 8 & 6 & 5 & 3 & 2 \\
  & PHYBench       & 7 & 5 & 1 & 8 & 6 & 4 & 2 & 3 \\
\midrule
\multirow{4}{*}{Math}
  & Math 500       & 8 & 4 & 1 & 6 & 7 & 5 & 2 & 3 \\
  & AIME 2025      & 8 & 5 & 1 & 6 & 7 & 4 & 2 & 3 \\
  & EMMA-Math      & 6 & 4 & 1 & 8 & 6 & 5 & 3 & 2 \\
  & LiveBench-Math & 7 & 5 & 1 & 6 & 8 & 4 & 2 & 3 \\
\midrule
\multirow{3}{*}{Code}
  & BigCodeBench   & 5 & 1 & 3 & 6 & 4 & 2 & 7 & 8 \\
  & Aider Coding   & 6 & 4 & 2 & 8 & 7 & 5 & 1 & 3 \\
  & LiveBench-Code & 3 & 4 & 5 & 8 & 6 & 7 & 2 & 1 \\
\midrule
\multirow{3}{*}{Vision}
  & VISTA          & 6 & 4 & 1 & 7 & 8 & 5 & 3 & 2 \\
  & MMMU           & 7 & 5 & 1 & 6 & 8 & 4 & 2 & 3 \\
  & Chatbot Arena (Vision) & 7 & 5 & 1 & 6 & 8 & 4 & 2 & 3 \\
\midrule
\multirow{2}{*}{Language}
  & MultiChallenge & 5 & 2 & 3 & 7 & 8 & 4 & 1 & 6 \\
  & LiveBench-Lang & 6 & 4 & 3 & 7 & 8 & 2 & 1 & 5 \\
\midrule
\multirow{2}{*}{Puzzle}
  & EnigmaEval     & 6 & 4 & 5 & 8 & 7 & 3 & 1 & 2 \\
  & NYT Connections& 8 & 5 & 4 & 7 & 6 & 2 & 1 & 3 \\
\bottomrule
\end{tabular}
\vspace{0.5em}
\label{tab:benchmark_ranks_only}
\end{table}

%% file: figure_text/corr_feature_mario.tex
\begin{figure}[t]
    \centering
    \begin{minipage}[c]{0.49\linewidth}
        \centering
        \includegraphics[width=\linewidth]{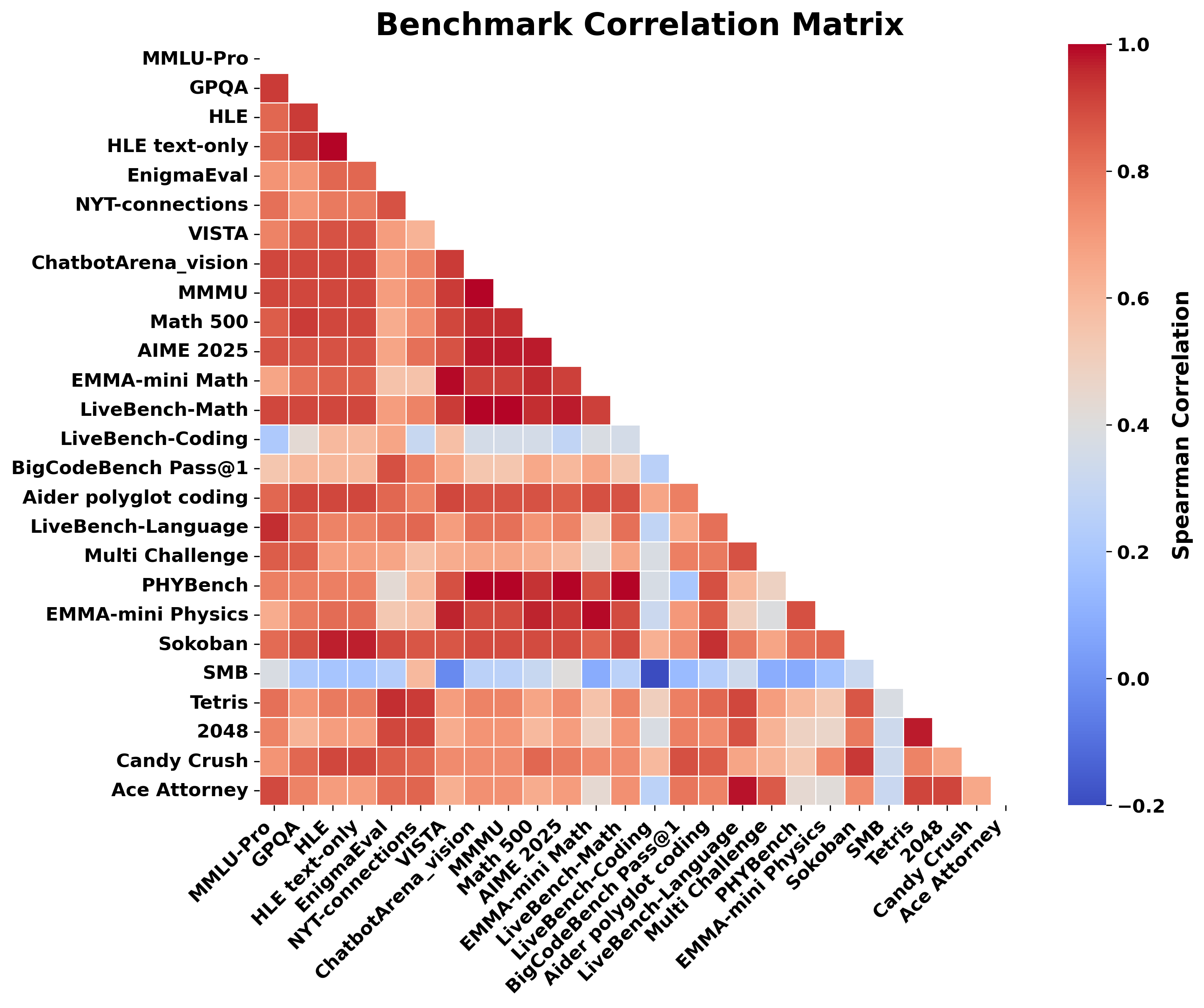}
        \subcaption{Spearman correlation matrix including \textit{Super Mario Bros. (SMB)}.}
        \label{fig:corr_mario}
    \end{minipage}
    \hfill
    \begin{minipage}[c]{0.49\linewidth}
        \centering
        \includegraphics[width=\linewidth]{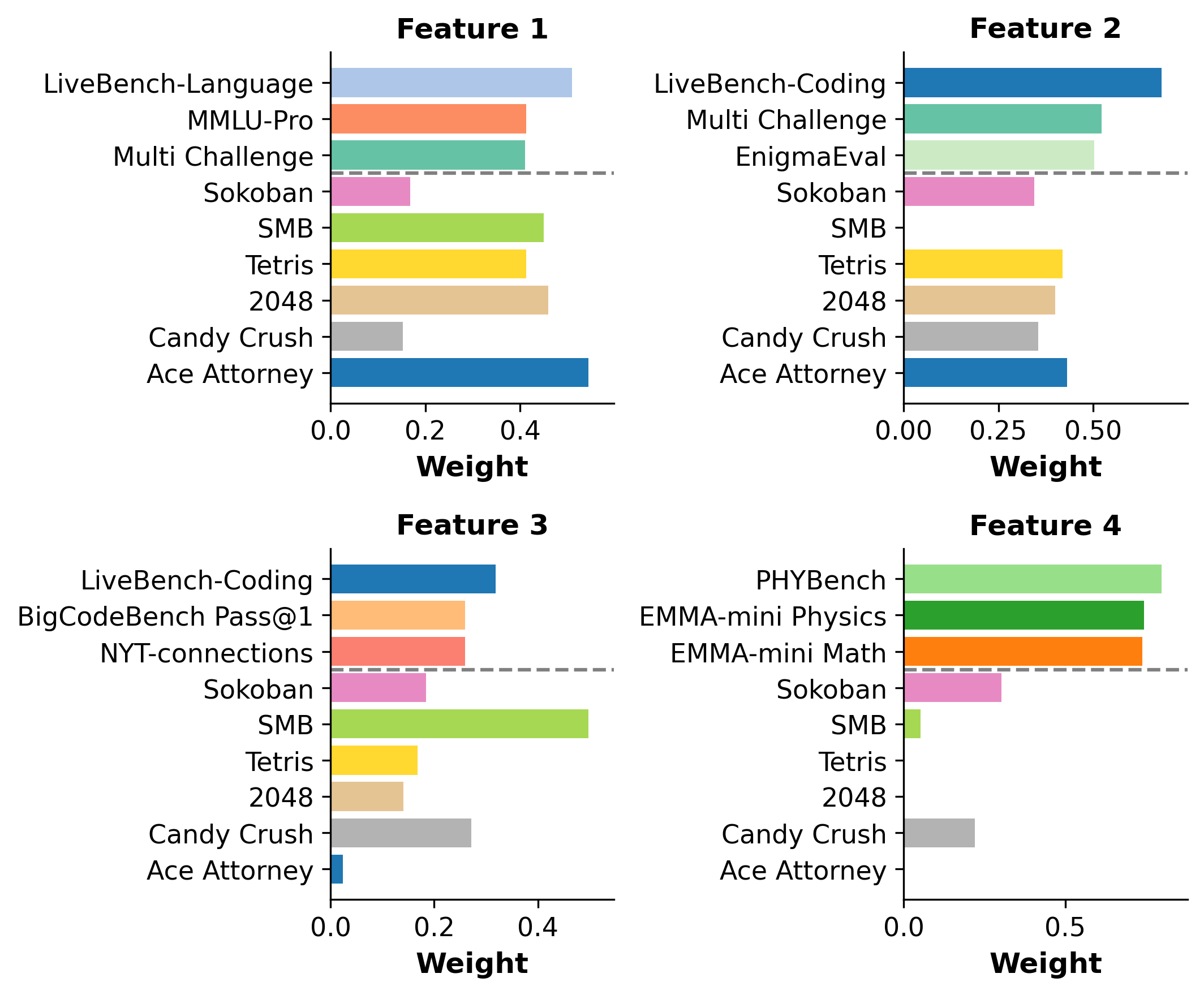}
        \subcaption{Top-weight benchmarks for each latent feature when SMB is included. }
        \label{fig:low_rank_mario}
    \end{minipage}
    \caption{Benchmark relationships overview.}
    \label{fig:corr_and_feature_mario}
\end{figure}

%% file: figure_text/tsne.tex
\begin{figure}[t]
    \centering
    \begin{minipage}[c]{0.49\linewidth}
        \centering
        \includegraphics[width=\linewidth]{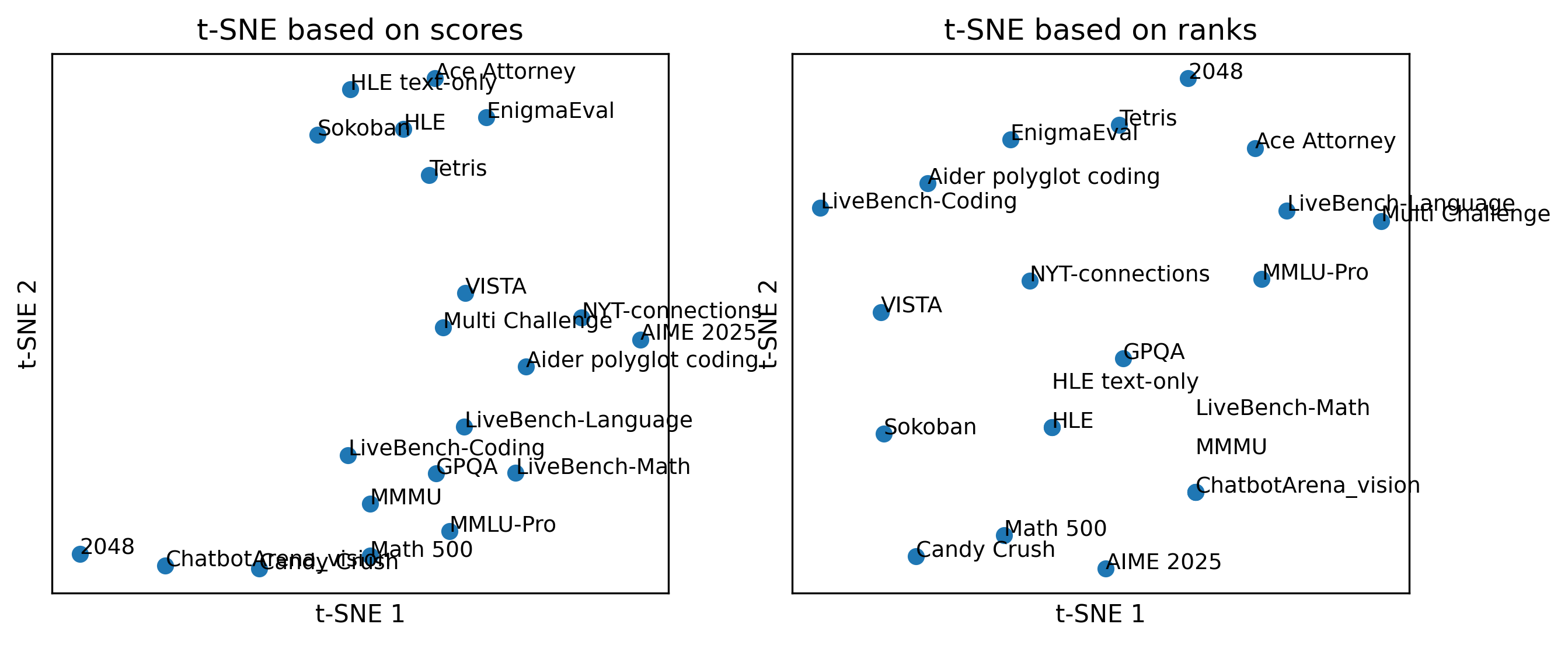}
        \subcaption{t-SNE analysis. Without Super Mario Bros.}
        \label{fig:tsne_w_smb}
    \end{minipage}
    \hfill
    \begin{minipage}[c]{0.49\linewidth}
        \centering
        \includegraphics[width=\linewidth]{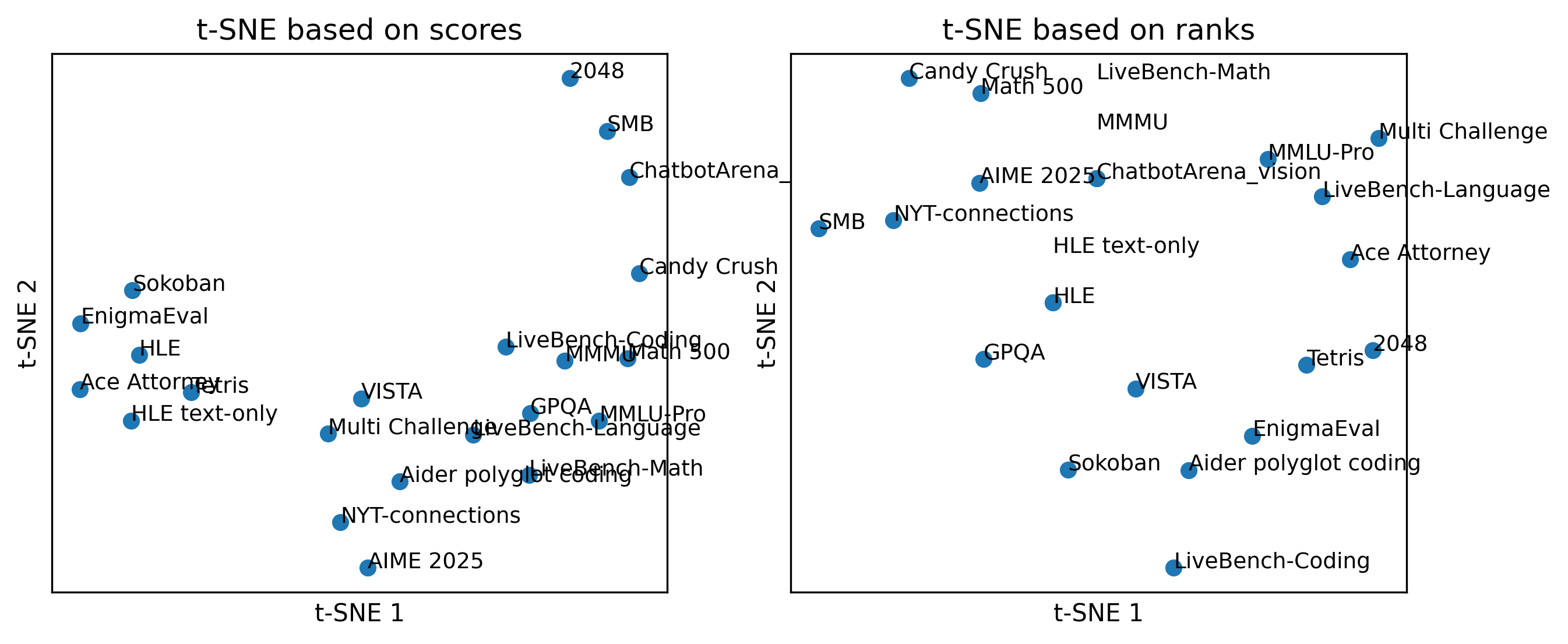}
        \subcaption{t-SNE analysis. With Super Mario Bros.}
        \label{fig:tsne_wo_smb}
    \end{minipage}
    \caption{Benchmark relationships overview.}
    \label{fig:tsne}
\end{figure}

%% file: table/polynomial_model_linear_ablation.tex
\begin{table}[t]
  \centering
  \caption{Additional ablation on polynomial (linear \& quadratic) modeling with different combinations of core capablities.}
  \scriptsize

  \begin{tabular}{l|ccccc|cc}
    \toprule
    \multicolumn{8}{c}{\textbf{Linear Model $(n=5)$}}\\
    \textbf{Game} & \textbf{Knowledge} & \textbf{Puzzle} & \textbf{Visual} &
    \textbf{Math} & \textbf{Coding} & Offset & $r$ \\
    \midrule
    Sokoban            & 1.299 & 2.426 & 0.009 & 1.731 & 1.482 & 0.106 & 0.9754\\
    Tetris             & 0.000    & 6.559 & 0.964 & 0.005 & 0.455 & 0.009 & 0.9370\\
    Ace~Attorney       & 1.579 & 4.850 & 0.003 & 0.680 & 0.000      & 0.245 & 0.8519\\
    Super~Mario~Bros   & 0.653   & 0.000      & 0.000      & 1.304 & 0.737 & 2.970 & 0.2215 \\
    2048               & 0.000     & 4.958 & 0.000     & 0.000    & 0.000   & 1.588 & 0.7338\\
    Candy~Crush        & 2.188 & 3.326 & 0.000      & 0.916 & 1.583 & 0.000      & 0.9086\\
    \bottomrule
  \end{tabular}

  \vspace{.8em}

  \begin{tabular}{l|cccc|cc}
    \toprule
    \multicolumn{7}{c}{\textbf{Linear Model $(n=4)$}}\\
    \textbf{Game} & \textbf{Knowledge} & \textbf{Visual} &
    \textbf{Math} & \textbf{Coding} & Offset & $r$ \\
    \midrule
    Sokoban            & 2.581 & 0.011 & 1.954 & 2.029 & 0.308 & 0.9471\\
    Tetris             & 3.432 & 0.405 & 1.063 & 2.116 & 0.558 & 0.8128\\
    Ace~Attorney       & 4.493 & 0.014 & 1.582 & 0.000   & 0.820 & 0.7481\\
    Super~Mario~Bros   & 0.942 & 0.715 & 1.102 & 0.974   & 2.289 & 0.2535\\
    2048               & 0.004 & 0.000 & 2.249 & 0.000   & 3.111 & 0.3610\\
    Candy~Crush        & 3.646 & 0.000 & 1.600 & 2.393  & 0.198 & 0.8388\\
    \bottomrule
  \end{tabular}

  \vspace{.8em} 


  \resizebox{\linewidth}{!}{%
  \begin{tabular}{l|ccccccccc|cc}
    \toprule
    \multicolumn{12}{c}{\textbf{Quadratic Model $(n=3)$}}\\
    \textbf{Game} & \textbf{Klg} & \textbf{Math} & \textbf{Coding} & \textbf{Klg$^2$} & \textbf{Math$^2$} & \textbf{Coding$^2$}  & \textbf{Klg $\times$ Math} & \textbf{Klg $\times$ Coding} & \textbf{Math $\times$ Coding}  & Offset & RE \\
    \midrule
    Sokoban  & 1.284  & 0.439  & 0.758 & 0.499 & 0.934 & 0.963 & 0.673 & 0.616 & 0.586 &   1.497  &  0.801         \\
    Tetris   &  1.266 & 0.423  & 0.745  & 0.481 & 0.921 & 0.946  & 0.657 &  0.623 &  0.600 &   1.489  & 0.797  \\
    Ace~Attorney    & 1.546 &  0.629  &  0.0111  & 1.184  & 0.725  & 0.025 & 0.910 & 0.428 & 0.330 & 1.626  &   1.037    \\
    Candy~Crush     & 1.456  & 0.428  &  1.410  & 0.343  &  1.627  & 1.021 & 0.951 &  0.000 & 0.000 &  1.098 &    0.7503  \\
    \bottomrule
  \end{tabular}%
  }
  \label{tab:polynomial_model}
  \vspace{5pt}
\end{table}

%% file: text/7_appendix_5_harness_effectiveness.tex
\section{Harness Effectiveness: Quatitative Analysis}
\label{appendix:harness_hypothesis_test}
In this section, we employ statistical methods to explore the effectiveness of applying all harness combined in bringing improvements to model performance. Given the tiny sample due to the cost of running latest models, results should be considered preliminary.

Given that gameplay inherently involves random noise, we aim for our harnessed model performance to be both noise-resistant and consistent, enabling clearer assessment of the model’s true ability. Accordingly, we make two key claims: (1) harnessed evaluations better isolate model ability from game randomness; and (2) harnessed performance is more consistent and robust to random variation.

\subsection{Separation from Random Baseline: Glass’s $\delta$ Effect Sizes}

To quantify how far harnessed and unharnessed model evaluations depart from random play, we simulated 30 random runs per game to estimate the baseline mean $\bar X_{\mathrm{rand}}$ and standard deviation $s_{\mathrm{rand}}$.  Glass’s $\delta$ for each model–game–condition is then \cite{glass1976primary}:
\begin{equation}
  \delta \;=\; \frac{\bar X_{\mathrm{model}} - \bar X_{\mathrm{rand}}}{s_{\mathrm{rand}}}\
\label{eq:glass_delta}
\end{equation}

Because Sokoban and Ace Attorney exhibit zero variance under random play, we exclude them, focusing on the four remaining games.  Importantly, harnessed runs yield positive \(\delta\) in 38 out of 40 model–game pairs, compared to only 26 out of 40 for unharnessed runs—demonstrating that harnessed evaluations are far more consistently pulled away from the random baseline.  Across those 40 pairs, harnessed runs outperform unharnessed in 29 cases (72.5\%), with overall averages

\begin{equation}
  \bar\delta_{\mathrm{harness}} = 3.334,\quad
  \bar\delta_{\mathrm{no}} = 0.750,\quad
  \Delta^* = \bar\delta_{\mathrm{harness}} - \bar\delta_{\mathrm{no}} = 2.585
\label{eq:glass_delta_value}
\end{equation}

This demonstrates that the harness pulls model scores substantially farther from randomness than unharnessed evaluations.

\input{table/glass_delta}

\subsection{Direct Comparison of Harnessed vs.\ Unharnessed: Paired‑Sample t‑Test}

Beyond Glass’s \(\delta\), we directly compare harnessed and unharnessed mean scores via paired‑sample t‑tests \cite{student1908probable} across our ten models for each game. All six games exhibit positive mean improvements under harnessing; for five of them—Candy Crush (+217.50 points, \(t(9)=4.22,\,p=0.0022\)), Sokoban (+1.97 points, \(t(9)=3.02,\,p=0.0144\)), 2048 (+17.81 points, \(t(9)=2.36,\,p=0.0424\)), Ace Attorney (+3.20 points, \(t(9)=2.36,\,p=0.0427\)), and Tetris (+5.60 points, \(t(9)=2.27,\,p=0.0490\))—the increase is statistically significant at \(p<0.05\). Super Mario Bros.\ shows a smaller, non‑significant gain (+289.10 points, \(t(9)=1.45,\,p=0.1806\)).

Figure~\ref{fig:paired_diffs_boxplot} displays the full distribution of per‑model score differences (Harness – No Harness) for each game, with boxes indicating the interquartile range and whiskers covering 1.5× IQR. Candy Crush and Sokoban show the largest median gains, while Super Mario Bros.\ exhibits the greatest spread, underscoring its high inherent stochasticity.

\input{table/t_test}
\input{figure_text/paired-t-test-boxplots}

\subsection{Consistency of Performance: Coefficient of Variation Across Conditions}

Because our per‑model samples are small (\(n\approx3\)), raw variance comparisons can be misleading.  We therefore compute the coefficient of variation (expressed as a percentage)

\begin{equation}
  \mathrm{CV} = \frac{s}{\bar X}\times 100\%\,
\label{eq:coeff_var}
\end{equation}

for each model–game–condition (Random, With Harness, Without Harness) to measure relative dispersion around the mean.  Sokoban and Ace Attorney are excluded (zero random variance), and models with only a single run (o1, o3) are omitted.  Table~\ref{tab:cv_full} lists the rounded CV values (in \%) for the remaining ten models across four games.

Across the four games, harnessed runs yield lower CV than random play in 8/8 cases for 2048 (100.0\%), 6/8 for Candy Crush (75.0\%), 6/8 for Super Mario Bros. (75.0\%), and 5/8 for Tetris (62.5\%).  Comparing harnessed to unharnessed, CV is lower under harness in 6/8 for 2048 (75.0\%), 8/8 for Candy Crush (100.0\%), 4/8 for Super Mario Bros. (50.0\%), and 4/8 for Tetris (50.0\%).  Overall, out of 32 valid model–game pairs, 25 (78.1\%) have smaller CV under harness versus random, and 22 (68.8\%) have smaller CV under harness versus unharnessed.  These results indicate that the harness not only elevates mean performance but also lowers the coefficient of variation—i.e. reduces relative score dispersion—which yields more stable, reliable assessments of model ability.

\input{table/coefficients_of_variance}


\tcbset{
  reasoningstyle/.style={
     colback=gray!5,
     colframe=purple,          
     coltitle=green,             
     fonttitle=\bfseries,      
     boxrule=1.2pt,   
     arc=4pt,              
     left=1em,right=1em,top=1em,bottom=1em,
  }
}
\section{Harness Effectiveness: Qualitative Analysis}
\label{appendix:memory_module_analysis}

In this section, we present case studies that demonstrate the reasoning traces of the latest models during their 2048 gameplay.

\subsection{Memory Module}

\textbf{Memory Module Reflection Comparison: GPT-4o vs. o3} 

In our 2048 case study, GPT-4o's reflections tend to focus on immediate effect of each move ("up increased empty spaces but produced no merges"), whereas o3's reflections unpack exactly which tiles merged, how merge ladders are forming, and even quantify changes in empty-cell counts("two 2's became a 4 and two 16's became 32's, opening an extra cell"). Importantly, an interactive game environment isn't a one-off Q\&A but a multi-turn dialogue with a constantly shifting game board - each action affects next decision. o3 better links its fine-grained observations to concrete next steps, integrating short-term move outcomes into its long-term corner-dominance strategy. The success of local decisions is crucial to achieving long‑term goals in a multi‑turn interactive game environment.

\noindent\textbf{Highlight Legend:}

\hlred{Red text}: the model’s \emph{immediate} game‐state reflection  
    
\hlyellow{Yellow text}: o3’s \emph{immediate} game‐state reflection  
    
\hlgreen{Green text}: the model’s \emph{future move} suggestion

\newtcolorbox{reasoningbox-4o}{
  reasoningstyle,
  title={GPT-4o Game 2048 Memory Module Reflections}
}

\newtcolorbox{reasoningbox-o3}{
  reasoningstyle,
  title={o3 Game 2048 Memory Module Reflections}
}
\tcbset{%
  sampleboxstyle/.style={%
    reasoningstyle,
    boxrule=0.8pt,
    coltitle=white,
    fonttitle=\bfseries,
    leftrule=0pt, rightrule=0pt, bottomrule=0pt, toprule=0pt,
    arc=2pt,
    left=1em, right=1em, top=1em, bottom=1em,
    boxsep=2pt,
    colback=white
  }
}

\newtcolorbox{samplebox}[2][]{%
  sampleboxstyle,
  title={Sample~#2},
  #1
}

\begin{reasoningbox-4o}
  \begin{samplebox}{1}
    \begin{minipage}[c]{0.48\textwidth}
      \textbf{Reflection:} \hlred{The move "up" increased empty spaces from 2 to 4, improving board flexibility while maintaining the highest tile (256) in position. However, no significant merges occurred, suggesting the move was minimally effective for progress.} The key pattern is the consistent maintenance of the highest tile in a corner, aligning with the strategy. \hlgreen{Future moves should focus on creating merge opportunities, particularly around the 256 tile, while avoiding disruptions to its placement. Consider prioritizing "down" or "left" next to consolidate lower tiles toward the main corner, optimizing for larger merges.}
    \end{minipage}\hfill
    \begin{minipage}[c]{0.48\textwidth}
      \centering
      \includegraphics[width=0.9\linewidth]{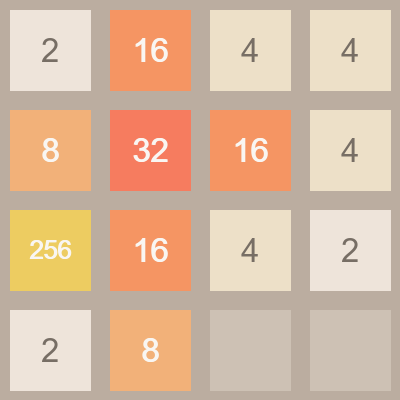}
    \end{minipage}
  \end{samplebox}

  \begin{samplebox}{2}
    \begin{minipage}[c]{0.48\textwidth}
      \textbf{Reflection:} \hlred{The "left" action was suboptimal as it failed to create merges or improve board position, leaving only two empty spaces while slightly disrupting alignment.} The highest tile (256) remains safe in a corner, but the addition of a 2 at (0,3) weakens the merging path potential. Merges were not prioritized, and the action did not capitalize on available opportunities to consolidate tiles. \hlgreen{Future moves should focus on maximizing merges and preserving space, particularly by aligning higher tiles on one edge to maintain control and avoid corner disruption. Aim to optimize empty space usage for better tile positioning.}
    \end{minipage}\hfill
    \begin{minipage}[c]{0.48\textwidth}
      \centering
      \includegraphics[width=0.9\linewidth]{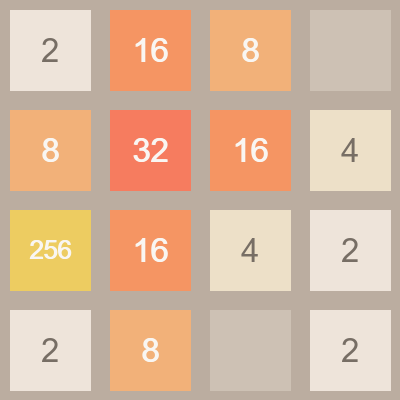}
    \end{minipage}
  \end{samplebox}

  \begin{samplebox}{3}
    \begin{minipage}[c]{0.48\textwidth}
      \textbf{Reflection:} \hlred{The "down" move consolidated smaller tiles, creating merges (e.g., 2s to 4s), but did not impact the highest tile (256) or significantly improve positioning.} Empty spaces shifted, but the same two remain, limiting opportunities. The action maintained balance but lacked strategic progression toward higher merges. The key issue is the scattered mid-tier tiles, which could restrict future merges. \hlgreen{Focus on better clustering tiles toward one corner and maintain merge paths for the 256 tile. Prioritizing "up" or "left" next may help consolidate the board and preserve the highest tile’s strategic position.}
    \end{minipage}\hfill
    \begin{minipage}[c]{0.48\textwidth}
      \centering
      \includegraphics[width=0.9\linewidth]{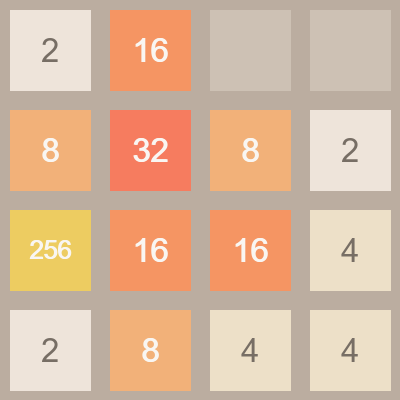}
    \end{minipage}
  \end{samplebox}
\end{reasoningbox-4o}

\begin{reasoningbox-o3}
  \begin{samplebox}{1}
    \begin{minipage}[c]{0.48\textwidth}
      \textbf{Reflection:} \hlyellow{The right swipe merged the two 8‑tiles in row 2 (3 → 4), created a 16 and opened an extra gap; the new 2 spawned harmlessly at (1,0).} 256 stayed anchored in the bottom‑right, and the right column is now 256‑128‑64‑16, a strong merge ladder. Empty count stayed at four, so the board is still breathable.Good move: it scored, preserved the corner strategy and improved board order.Watch for: isolated 2’s creeping into the left half; they can freeze rows if ignored. \hlgreen{Next: keep sliding Right/Down to compress toward the bottom‑right; avoid Up unless desperate to prevent dislodging the 256.}

    \end{minipage}\hfill
    \begin{minipage}[c]{0.48\textwidth}
      \centering
      \includegraphics[width=0.9\linewidth]{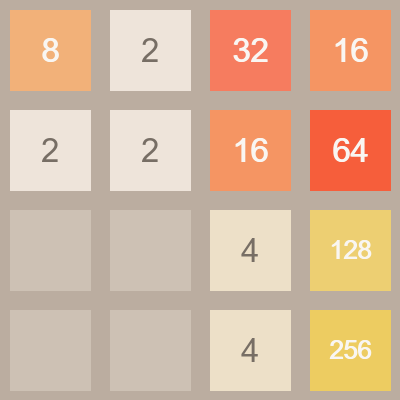}
    \end{minipage}
  \end{samplebox}

  \vspace{1em}

  \begin{samplebox}{2}
    \begin{minipage}[c]{0.48\textwidth}
      \textbf{Reflection:} \hlyellow{Up combined the two 2‑tiles in column 2 into an 8 and merged the pair of 2‑tiles in column 1 into a 4, netting +2 empty squares; after the new spawn you still gained one extra space (4 → 5).} 256 stayed anchored in the bottom‑right, so the corner strategy is intact. Good move: it advanced tile values, improved board looseness and preserved ordering. Watch the freshly spawned 2 at (3,2); if the bottom row clogs you’ll lose the 256 corner. \hlgreen{Prefer more up/left sequences to clear column 2 and funnel new merges toward the right column.}
    \end{minipage}\hfill
    \begin{minipage}[c]{0.48\textwidth}
      \centering
      \includegraphics[width=0.9\linewidth]{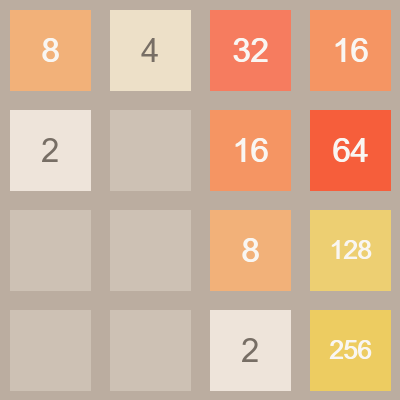}
    \end{minipage}
  \end{samplebox}

  \vspace{1em}

  \begin{samplebox}{3}
    \begin{minipage}[c]{0.48\textwidth}
      \textbf{Reflection:} \hlyellow{Down kept the 256 anchored and slid every column smoothly, but it yielded zero merges and cost one empty slot (5 → 4).} Column 0 merely shuffled 8‑2 downward and spawned a new 2 on top, so no pairing potential was created. Repeated space‑losing moves like this can clog the board; always look for turns that both respect the corner strategy and create or set up merges (e.g., left could have joined the two 2‑tiles on the bottom row next). \hlgreen{Prioritize moves that grow space or prepare obvious combines before defaulting to safe ‘corner‑holding’ slides.}
    \end{minipage}\hfill
    \begin{minipage}[c]{0.48\textwidth}
      \centering
      \includegraphics[width=0.9\linewidth]{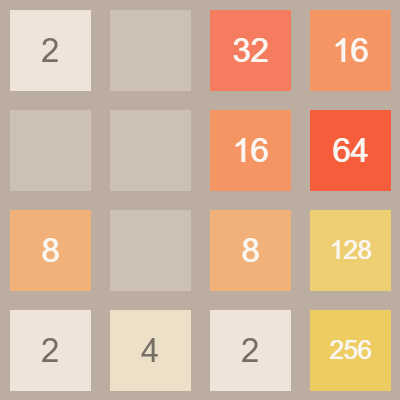}
    \end{minipage}
  \end{samplebox}
\end{reasoningbox-o3}

\textbf{Memory Module Prompt Template} 
\newtcolorbox{memory-prompt}{
  reasoningstyle,
  title={Game 2048 Memory Module}
}
\begin{memory-prompt}
\textbf{\textcolor{blue}{system\_prompt:}}

You are an analytical assistant for a 2048 AI agent. \\
Your task is to generate a brief, insightful reflection on the game state changes and the effectiveness of recent actions.Focus on strategic insights and patterns that would help the agent make better decisions.\\
Keep your reflections short, precise, and actionable.\\

\textbf{\textcolor{blue}{user\_prompt:}}

Please analyze the following 2048 game states and actions to generate a brief reflection:\\
Previous Game States and Actions:\\
\textcolor{blue}{\{Previous Game History\}}\\
Focus your reflection on:\\
1. How the game state changed after the last action\\
2. Whether the action was effective for the situation\\
3. Patterns or issues to be aware of\\
4. Any strategic insights for future actions\\
Keep your reflection under 100 words and focus only on the most important insights.
\end{memory-prompt}

%% file: table/glass_delta.tex
\begin{table}[h!]
  \centering
  \small
  \caption{Glass’s \(\delta\) per Model, Condition, and Game (rounded to 3 decimals)}
  \begin{tabular}{llrrrr}
    \toprule
    \textbf{Model} & \textbf{Cond.} & \textbf{2048} & \textbf{Candy Crush} & \textbf{SMB} & \textbf{Tetris} \\
    \midrule
    \multirow{2}{*}{claude-3-5-sonnet-20241022}  
      & With    &  0.992 & –0.204 & 1.763 & 2.524 \\
      & Without & –5.446 & –1.933 & 2.593 & 1.215 \\
    \multirow{2}{*}{claude-3-7-sonnet-20250219 (thinking)}  
      & With    &  1.648 &  7.140 & 2.223 & 3.459 \\
      & Without &  1.752 &  0.191 & 2.258 & 1.589 \\
    \multirow{2}{*}{gemini-2.5-flash-preview-04-17 (thinking)}  
      & With    &  0.787 &  4.238 & 2.151 & 3.459 \\
      & Without &  0.883 & –0.366 & 2.595 & 4.955 \\
    \multirow{2}{*}{gemini-2.5-pro-preview-05-06 (thinking)}  
      & With    &  2.148 &  5.825 & 2.466 & 7.386 \\
      & Without &  2.558 &  1.182 & 1.024 & 1.215 \\
    \multirow{2}{*}{gpt-4.1-2025-04-14}  
      & With    &  0.675 &  1.273 & 4.382 & 1.963 \\
      & Without & –0.762 & –0.301 & 3.970 & 1.589 \\
    \multirow{2}{*}{gpt-4o-2024-11-20}  
      & With    &  0.793 &  0.599 & 4.141 & 2.150 \\
      & Without & –3.833 & –1.117 & 1.033 & 2.524 \\
    \multirow{2}{*}{llama-4-maverick-17b-128e-instruct-fp8}  
      & With    &  0.707 &  0.236 & 2.376 & 0.093 \\
      & Without & –7.124 & –1.635 & 0.293 & 0.841 \\
    \multirow{2}{*}{o1-2024-12-17}  
      & With    &  3.631 &  0.826 & 0.504 & 13.930 \\
      & Without &  3.530 & –0.515 & 2.270 & 1.589 \\
    \multirow{2}{*}{o3-2025-04-16}  
      & With    &  3.516 & 10.306 & 8.404 & 17.856 \\
      & Without &  3.541 & –0.204 & 3.859 & 11.686 \\
    \multirow{2}{*}{o4-mini-2025-04-16}  
      & With    &  2.577 &  7.204 & 2.313 & 8.508 \\
      & Without & –0.368 & –0.113 & 2.009 & 2.711 \\
    \bottomrule
  \end{tabular}
  \vspace{0.5em}
  \label{tab:glass_delta_full}
\end{table}

%% file: table/t_test.tex
\begin{table}[!ht]
  \centering
  \vspace{-10pt}
  \caption{Paired‑Sample t‑Test Results for Harnessed vs.\ Unharnessed Mean Scores}
  \setlength{\tabcolsep}{12pt} 
  \renewcommand{\arraystretch}{1.2} 
  \begin{tabular}{lrrrr}
    \toprule
    \textbf{Game}           & \(\Delta\) Mean & \(\%\Delta\)   & \(t\,(df=9)\) & \(p\)      \\
    \midrule
    Candy Crush             & +217.50      & +224.8\%  & 4.22    & 0.0022 ** \\
    Sokoban                 & +1.97        & +537.5\%  & 3.02    & 0.0144 *  \\
    2048                    & +17.81       & +22.4\%   & 2.36    & 0.0424 *  \\
    Ace Attorney            & +3.20        & +123.1\%  & 2.36    & 0.0427 *  \\
    Tetris                  & +5.60        & +27.1\%   & 2.27    & 0.0490 *  \\
    Super Mario Bros.\      & +289.10      & +19.3\%   & 1.45    & 0.1806    \\
    \midrule
    \multicolumn{5}{l}{\footnotesize{* \(p<0.05\), ** \(p<0.01\)}} \\
  \bottomrule
  \end{tabular}
  \vspace{0.5em}
  \label{tab:paired_t}
  \vspace{-10pt}
\end{table}

%% file: figure_text/paired-t-test-boxplots.tex
\begin{figure}[h!]
  \centering
  \includegraphics[width=1\linewidth]{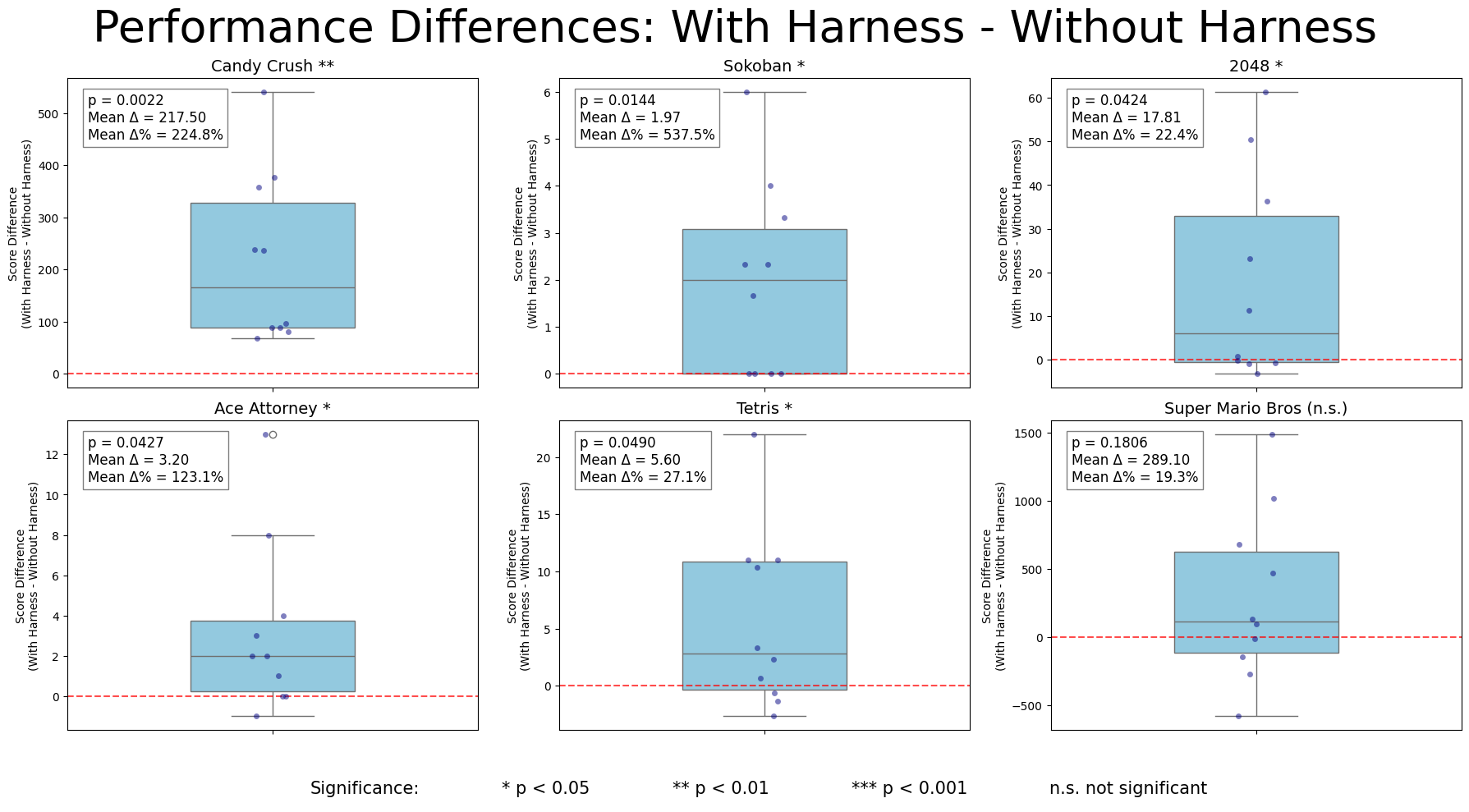}
  \caption{Distribution of paired score improvements (Harness – No Harness) across ten models for each game.}
  \label{fig:paired_diffs_boxplot}
  \vspace{-5pt}
\end{figure}

%% file: table/coefficients_of_variance.tex
\begin{table}[h!]
  \centering
  \vspace{-1pt}
  \small
  \caption{Coefficient of Variation (CV \%) by Model, Condition, and Game}
  \label{tab:cv_full}
  \resizebox{\linewidth}{!}{%
  \begin{tabular}{llrrrr}
    \toprule
    \textbf{Model} & \textbf{Condition} 
      & \textbf{2048} & \textbf{Candy Crush} 
      & \textbf{Super Mario Bros} & \textbf{Tetris} \\
    \midrule
    Random                        
      & Random         &   7.798 &  44.183 &  41.995 &  17.535 \\
    \addlinespace
    claude-3-5-sonnet-20241022    
      & With Harness   &   5.351 &  50.372 &  38.184 &   7.873 \\
      & Without Harness&  28.377 & 106.371 &   1.409 &  20.405 \\
    \addlinespace
    claude-3-7-sonnet-20250219 (thinking)    
      & With Harness   &   2.766 &  11.090 &  46.546 &  14.139 \\
      & Without Harness&   6.285 &  54.694 &  11.341 &   0.000 \\
    \addlinespace
    gemini-2.5-flash-preview-04-17 (thinking)
      & With Harness   &   4.962 &  19.581 &  17.210 &  19.681 \\
      & Without Harness&   3.172 &  36.978 &  17.034 &  24.119 \\
    \addlinespace
    gemini-2.5-pro-preview-05-06 (thinking)
      & With Harness   &   4.993 &   1.635 &  13.573 &   2.474 \\
      & Without Harness&   3.245 &  36.599 &  43.228 &  24.771 \\
    \addlinespace
    gpt-4.1-2025-04-14           
      & With Harness   &   6.605 &  15.763 &  83.636 &   4.225 \\
      & Without Harness&  17.945 & 119.047 &  51.144 &  13.323 \\
    \addlinespace
    gpt-4o-2024-11-20            
      & With Harness   &   3.266 &  36.249 &  25.797 &  25.754 \\
      & Without Harness&  21.545 &  92.509 &  63.796 &  14.193 \\
    \addlinespace
    llama-4-maverick-17b-128e-instruct-fp8 
      & With Harness   &   3.556 &  44.459 &  37.835 &  14.783 \\
      & Without Harness&  26.470 & 128.055 &  58.857 &   9.897 \\
    \addlinespace
    o4-mini-2025-04-16           
      & With Harness   &   4.030 &  40.634 &  11.116 &  33.572 \\
      & Without Harness&  29.880 &  44.866 &  13.210 &  24.037 \\
    \bottomrule
  \end{tabular}
  }
  \vspace{-1pt}
\end{table}

%% file: text/7_appendix_6_games.tex
\section{Details of Evaluation Metrics}
\label{appendix:game_details}
Because games are well-designed, we utilize their built-in metrics to quantify models' proficiency. For each game, we choose the single score that most faithfully reflects a model’s capability, then transform and normalize it onto a continuous, linear scale. This curation ensures that our evaluation can sensitively capture performance differences and supports consistent statistical analysis.
\begin{itemize}
  \item \textbf{Sokoban:} Total number of boxes pushed onto targets, summed over all levels, until the first deadlock.
  \item \textbf{Super Mario Bros.:} Cumulative horizontal distance traveled by Mario (in game units) across all levels, until all three lives are lost or the final level is completed.
  \item \textbf{Tetris:} Total reward equals pieces placed plus ten times the number of lines cleared, measured until game over. Each placed piece yields +1; each cleared line yields +10.
  \item \textbf{2048:} Sum of all merged tile values (e.g.\ merging two 2’s yields +4), recorded until the board stagnates (no merges or moves that change the board for ten consecutive turns). We then report
    \[
      \text{Score}_{2048} \;=\; 10 \times \log_{2}\!\bigl(\text{total merged sum}\bigr).
    \]
  \item \textbf{Candy Crush:} Total number of candies eliminated over a fixed 50–move session.
  \item \textbf{Ace Attorney:} Total count of correct actions (evidence submissions, dialogue choices, etc.) across all case levels, measured until five incorrect decisions (lives) have been used.
\end{itemize}
